\documentclass[11pt]{article}

\usepackage{acl}
\usepackage{times}
\usepackage{latexsym}

\usepackage{booktabs}
\usepackage{multirow}
\usepackage{graphicx}
\usepackage{tcolorbox}
\usepackage{xcolor}
\tcbuselibrary{skins}
\usepackage{pifont}
\usepackage{xurl}
\usepackage{amsmath}
\usepackage{tabularx}
\usepackage{hyperref}
\usepackage{fontawesome5}
\usepackage{array}
\usepackage{amsfonts}
\usepackage{placeins}

\usepackage{booktabs}
\usepackage{tcolorbox}
\usepackage{xcolor}
\usepackage{array}
\usepackage{multirow}
\tcbuselibrary{skins, breakable}

\newcommand{\hfrepo}[2]{%
\href{https://huggingface.co/#1}{%
\raisebox{-0.2em}{\includegraphics[height=1em]{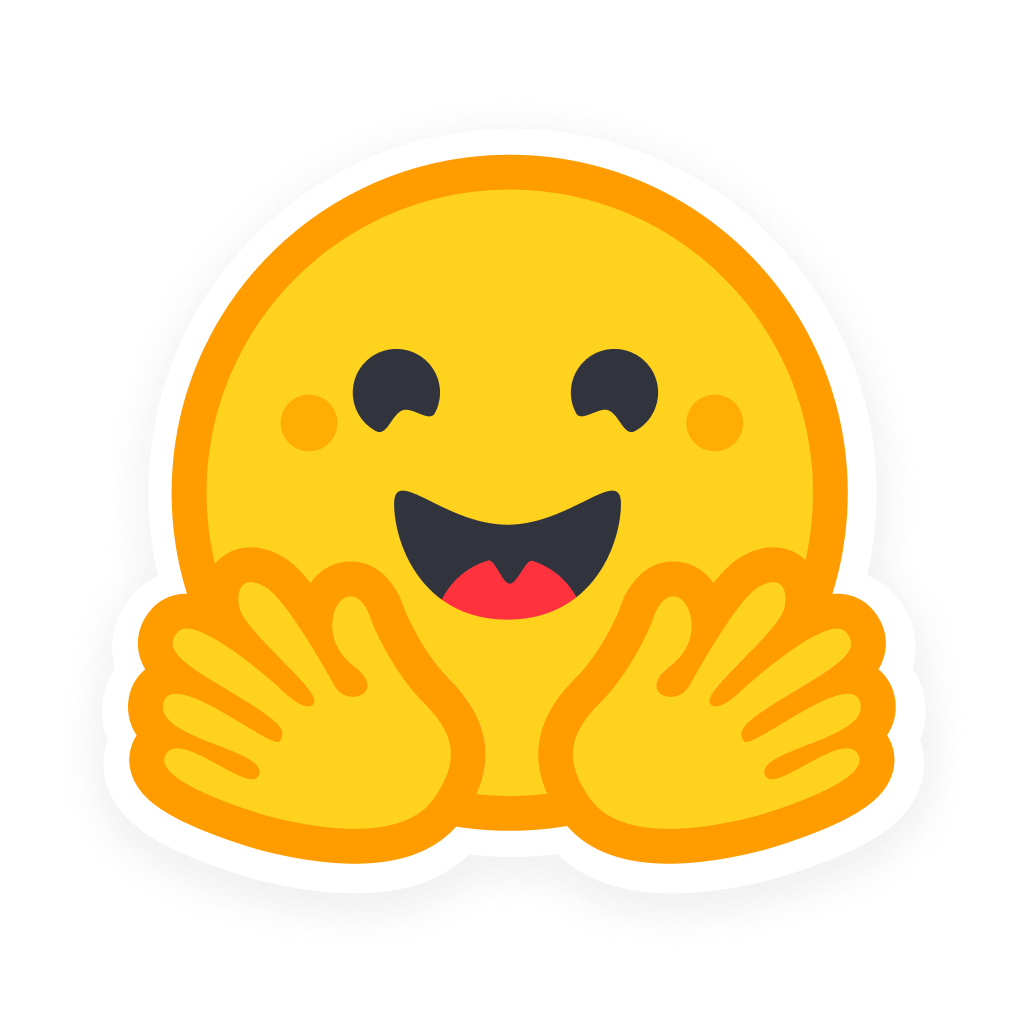}}\,\texttt{#2}%
}}

\usepackage[T1]{fontenc}
\usepackage[utf8]{inputenc}
\usepackage{microtype}
\usepackage{inconsolata}

\newcounter{promptbox}
\usepackage{colortbl}
\usepackage{makecell}

\definecolor{indomain}{RGB}{198,224,180}
\definecolor{notable}{RGB}{255,230,153}
\definecolor{locked}{RGB}{255,199,206}
\definecolor{generalist}{RGB}{221,235,247}

\newcommand{\cmark}{\textcolor{green!60!black}{\ding{51}}}
\newcommand{\xmark}{\textcolor{red!70!black}{\ding{55}}}
\newcommand{\pmark}{\textcolor{orange!80!black}{\ding{108}}}

\title{
\raisebox{-0.4\height}{\includegraphics[width=1.5cm]{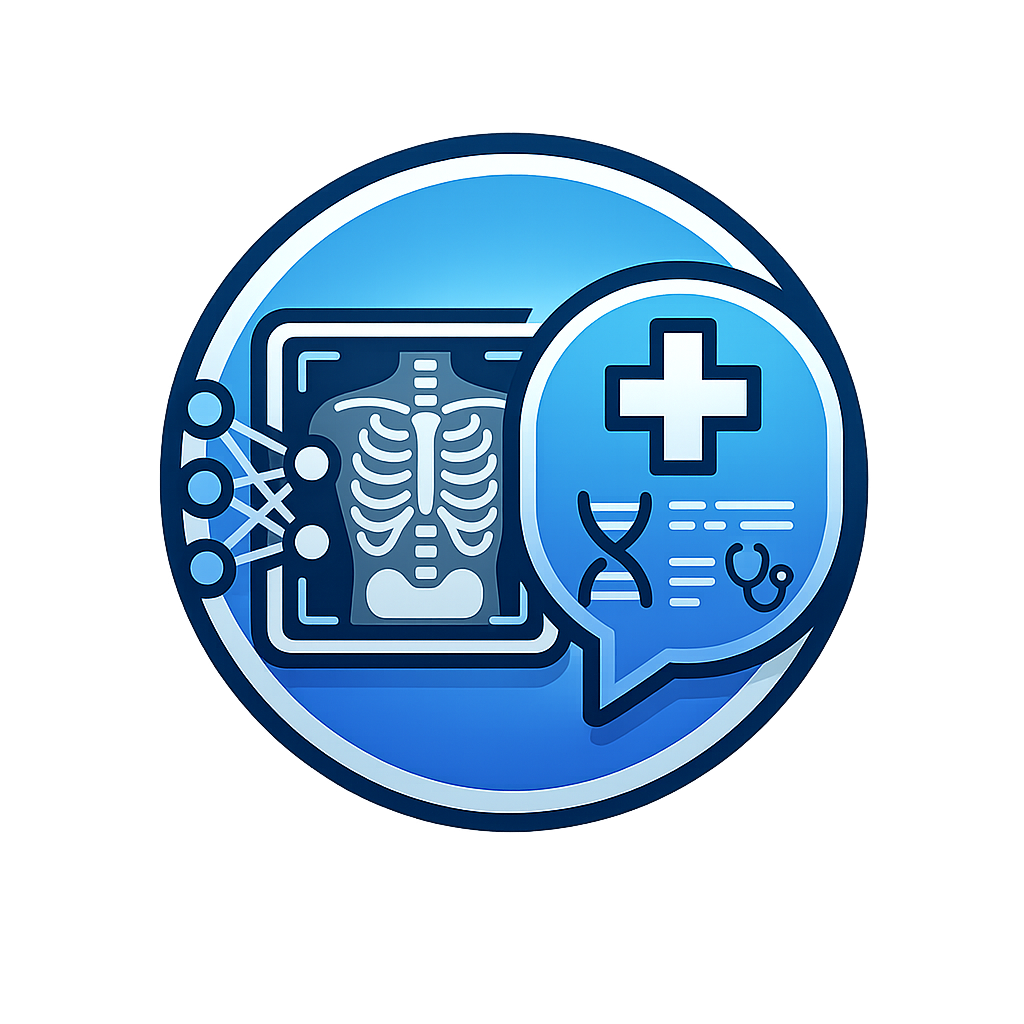}}
MedProb: Probing Internal Representations of Vision-Language Models for Medical Question Answering
}

\author{Erfan Nourbakhsh, Ke Yang, \and Anthony Rios \\
  The University of Texas at San Antonio \\
  \texttt{\{erfan.nourbakhsh, ke.yang, anthony.rios\}@utsa.edu} \\}

\begin{document}
\maketitle

\begin{abstract}
Medical visual question answering (Med-VQA) is often assumed to require medical fine-tuning, large models, or complex multi-agent pipelines. We revisit this assumption with \textbf{MedProb}, a lightweight probing framework that predicts multiple-choice Med-VQA answers from frozen VLM representations without free-text generation. Across PATH-VQA, SLAKE, and VQA-RAD, MedProb recovers substantially more answer-relevant signal than prompting and performs stronger than medical VLMs and agentic systems. Probing also reduces the apparent gap between small and large models compared to prompting, suggesting that smaller VLMs contain more recoverable Med-VQA signal than generation-based evaluation reveals. Across 14 matched general-purpose and medical VLM pairs, medical adaptation does not consistently improve this linear decodability. 
Finally, free-text generation exhibits an answer-position bias of up to 10 percentage points, {whereas MedProb also has positional bias, however, it is impacted differently than prompting}. {Our main results target the multiple-choice/multiclass Med-VQA setting; we additionally show the probe can be extended to open-ended generation via a rejection-sampling scoring procedure.} Code is available here: \url{https://github.com/erfan-nourbakhsh/MedProb}

\end{abstract}

\section{Introduction}

\begin{figure}[t]
    \centering
    \includegraphics[width=\linewidth]{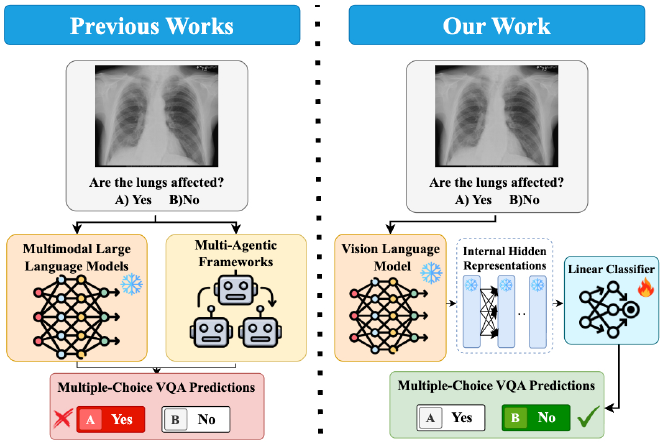}
    \caption{Prior methods (left) give a medical image and question to a VLM or multi-agent system, which then generates a free-text answer. This can lead to unsupported clinical reasoning. MedProb (right) uses the same input, but keeps the VLM frozen, extracts hidden representations from each layer, and predicts the multiple-choice answer with a linear classifier.}
    \label{fig:motivation} 
\end{figure}

Medical visual question answering (Med-VQA) asks models to answer clinical questions about medical images. Med-VQA is widely used to evaluate vision-language models (VLMs) in medical settings \cite{achiam2023gpt, lu2024deepseek, chen2024internvl} because clinical AI systems must connect visual evidence with medical language \cite{fries2022foundation, moor2023foundation}. Recent work often improves Med-VQA by adding more resources to the model or inference pipeline. This includes domain-adaptive pretraining on biomedical image-text data \cite{gururangan2020don, chen2024towards, xie2024medtrinity, wu2025towards}, multi-agent systems \cite{kim2024mdagents, li2024mmedagent}, tool-augmented pipelines \cite{fallahpour2025medrax}, retrieval-augmented generation \cite{nourbakhsh2026retrieval} and large closed-source models that are costly and difficult to inspect in privacy-sensitive clinical settings \cite{marks2023ai, sellergren2025medgemma}. These approaches differ in architecture and training, but they often share a common assumption that better Med-VQA requires larger models, more medical adaptation, or more complex inference.

We ask the question, when a VLM gives the wrong answer, is the answer-relevant ``signal'' absent from its internal representations, or is it present but not reflected in the generated answer? Current Med-VQA benchmarks usually evaluate models through free-text generation, which mixes several factors, including what the model represents in their internal features, how the prompt is written, how answer options are ordered, and how the model expresses its answer in language. A model may represent useful information about a medical image but still choose the wrong answer because the prompt is brittle, the answer options appear in a different order, or the generated explanation introduces unsupported reasoning.

This distinction matters for how we interpret model scale and medical adaptation. Under prompting, large VLMs often appear much stronger than small VLMs, and medically adapted models \textit{may} appear stronger than their general-purpose bases. However, part of these gaps may reflect differences in generation behavior rather than differences in the answer-relevant information available in the hidden states. Prior work supports this concern. Medically adapted models do not always outperform general-domain models under controlled prompting~\cite{Jeong2024MedicalAO}, clinical fine-tuning does not always make knowledge more accessible~\cite{zhao2025comparing}, and generalist VLMs can match or outperform specialist medical models on clinical tasks \cite{zhong2025can}. Prompting itself is also unstable, with semantically equivalent prompts changing accuracy by up to 76 points~\cite{sclar2023quantifying}.

We address this gap with \textbf{MedProb} (Figure~\ref{fig:motivation}), a lightweight probing framework for Med-VQA. MedProb freezes the VLM, extracts hidden representations from each transformer layer, and trains a linear classifier to predict the correct multiple-choice answer. MedProb is not intended to replace clinical reasoning systems or provide a fully supervised Med-VQA solution. Instead, it tests whether answer-relevant information is linearly recoverable from frozen VLM representations, separate from the difficulties introduced by free-text generation. {Our contribution is empirical and diagnostic rather than algorithmic: MedProb applies standard linear probing to frozen medical and general-purpose VLMs to measure how much of the closed-ended Med-VQA answer signal is linearly recoverable from their internal representations.}

Across PATH-VQA, SLAKE, and VQA-RAD, MedProb recovers substantially more answer-relevant signal than prompting. It also achieves competitive or higher closed-ended accuracy than several generation-based systems, including medically fine-tuned VLMs and multi-agent methods. We further find that the gap between small and large models is much smaller under probing than under prompting, suggesting that generation-based evaluation can underestimate the answer-relevant signal present in smaller models. Across 14 matched general-purpose and medically adapted VLM pairs, medical adaptation does not consistently improve the linear decodability of Med-VQA answer signal. Finally, free-text generation exhibits an answer-position bias of up to 10 percentage points, whereas MedProb is more stable under option reordering.

Our main contributions are as follows:
\begin{itemize}
\item We propose \textbf{MedProb}, a lightweight linear-probing framework that predicts multiple-choice Med-VQA answers from frozen VLM representations, separating linearly decodable answer signal from free-text generation.

\item We show that generation-based evaluation can underestimate the answer-relevant signal in VLM representations. Across three Med-VQA benchmarks, MedProb recovers substantially more task-relevant signal than prompting, and the small-versus-large model gap is much smaller under probing.

\item We analyze two evaluation confounds, domain adaptation and answer-option position. Across 14 matched model pairs, medical adaptation does not consistently improve the linear decodability of Med-VQA answer signal, and prompting accuracy changes by up to 10 percentage points depending on answer placement.
\end{itemize}

\section{Related Work}

\noindent \textbf{Medical Vision-Language Models.}
Medical VLMs have developed along two trajectories: adapting general-purpose VLMs to the biomedical domain and building purpose-built systems.
Early efforts include Med-Flamingo~\cite{moor2023med}, which extended OpenFlamingo~\cite{awadalla2023openflamingo} with few-shot medical reasoning, and LLaVA-Med~\cite{li2023llava}, which applied visual instruction tuning~\cite{liu2023visual} to PubMed image-text pairs.
Subsequent domain-adaptive pretraining (DAPT)~\cite{gururangan2020don} approaches include HuatuoGPT-Vision~\cite{chen2024towards} and MedTrinity-25M~\cite{xie2024medtrinity}, while BioMed variants of InternVL3~\cite{zhu2025internvl3}, Qwen2-VL~\cite{wang2024qwen2}, and Llama 3.2~\cite{grattafiori2024llama3herdmodels} follow the domain post-training recipe of \citet{cheng-etal-2025-domain}.
On the closed-source frontier, Med-Gemini~\cite{yang2024advancing} and GPT-4 leverage proprietary scale, while reasoning-augmented variants such as MedVLThinker~\cite{huang2025medvlthinker} and MediX-R1~\cite{mullappilly2026medix} demand extensive training resources.
Unlike these works, MedProb does not build a new medical VLM; instead, it probes frozen models and shows that medical adaptation does not consistently improve the linear decodability of answer-relevant signal in internal representations.

\vspace{1mm}
\noindent \textbf{Prompt Sensitivity in Medical VLM Evaluation.}
A growing body of work questions the reliability of prompting-based evaluation. \citet{sclar2023quantifying} showed that LLMs are highly sensitive to prompt formatting, with accuracy swings up to 76 points across semantically equivalent formats that persist even under scaling or instruction tuning.
In the medical domain, \citet{Jeong2024MedicalAO} found that domain-adaptive pretraining fails to consistently outperform general-domain baselines under controlled prompting. \citet{pezeshkpour2024large} further demonstrated that LLM performance varies substantially with the ordering of answer options in multiple-choice settings, a positional bias orthogonal to model knowledge.
Rather than documenting these instabilities, MedProb sidesteps them entirely by reading directly from frozen internal activations, yielding an evaluation paradigm that is inherently robust to both prompt formatting and option ordering.

\vspace{1mm}
\noindent \textbf{Representation Probing in Language Models.}
Probing has emerged as a principled alternative to generation-based evaluation for diagnosing model knowledge. \citet{maiya2025improving} showed that linear probes on frozen activations consistently outperform generation-based judgment across model families and datasets, generalizing under domain shifts. In the medical setting, \citet{berkowitz2025probing} demonstrated that linear classifiers on drug-name embeddings achieve AUC-ROC above 0.95 for adverse drug reaction classification without fine-tuning. Most directly relevant, the PING framework~\cite{berkowitz2025probing2} recovers up to 87\% of performance lost to safety alignment by probing frozen transformers on clinical QA, indicating that the primary bottleneck is generative rather than representational.
MedProb extends this paradigm to the multimodal Med-VQA setting, providing the systematic comparison of probing across 14 matched general-purpose and medically adapted VLM pairs on three clinical benchmarks.

\section{Methodology}
\label{sec:methodology}

We study Med-VQA as a multimodal multiple-choice classification task. Our goal is to compare what a VLM can produce through prompting with what can be read from its internal representations using simple linear probes. Figure~\ref{fig:overview} shows the full setup.

\noindent \textbf{Problem Setup.}
\label{sec:pf:data}
Each Med-VQA example contains a medical image, a question, a set of answer options, and the correct answer. Let $\mathcal{D}_{\mathrm{train}} = \{(I_i, q_i, \mathcal{O}_i, y_i)\}_{i=1}^{N_{\mathrm{train}}}$ and $\mathcal{D}_{\mathrm{test}} = \{(I_j, q_j, \mathcal{O}_j, y_j)\}_{j=1}^{N_{\mathrm{test}}}$ denote the training and test splits. Here, $I_i$ is the image, $q_i$ is the question, $\mathcal{O}_i = \{o_1,\dots,o_K\}$ is the set of candidate answers, and $y_i \in \mathcal{Y} = \{1,\dots,K\}$ is the gold label. We use $K=2$ for yes/no questions and $K=4$ for non-binary multiple-choice questions.

Intuitively, prompting asks the VLM to read the image and question, consider the answer choices, and then generate the option it thinks is correct. Given a prompt construction strategy $c$, we build the multimodal prompt as $p_i = \Phi(I_i, q_i, \mathcal{O}_i; c)$, where $\Phi$ combines the image, question, and answer options using template $c$. The VLM $f$ then produces a free-text response, and we extract the predicted option letter as $\hat{y}_i^{\mathrm{gen}} = \mathrm{Dec}(f, p_i)$, where $\mathrm{Dec}(\cdot)$ maps the generated text to one of the answer choices.

\noindent \textbf{MedProb: Layer-wise Representation Probing.}
\label{sec:pf:probe}
MedProb tests whether the correct answer is already present in the frozen VLM representations, even when the model does not generate the correct answer. Rather than changing the VLM or adding external knowledge, we freeze the model and train simple classifiers on hidden states from each layer.

Intuitively, if a linear probe can predict the correct answer from a layer's hidden state, then that layer contains answer-relevant information in a form that is easy to read out. For each transformer layer $\ell \in \{0,\dots,L\}$, we extract the hidden state of the last input token as a summary of the full multimodal context~\cite{zhao2025comparing}. We denote this representation as $h_i^{(\ell)} = \psi_\ell(f, p_i) \in \mathbb{R}^{d_\ell}$.

For each layer, we train a separate multinomial logistic regression probe. The probe maps the representation to a distribution over answer choices as $g_\ell(h_i^{(\ell)}) = \mathrm{softmax}(W^{(\ell)}h_i^{(\ell)} + b^{(\ell)})$, and predicts the answer with highest probability as $\hat{y}_i^{(\ell)} = \arg\max_{y \in \mathcal{Y}} [g_\ell(h_i^{(\ell)})]_y$. The parameters $W^{(\ell)} \in \mathbb{R}^{4 \times d_\ell}$ and $b^{(\ell)} \in \mathbb{R}^{4}$ are learned from the training representations $\{(h_i^{(\ell)}, y_i)\}_{i=1}^{N_{\mathrm{train}}}$.

We standardize features using training-set statistics. We choose the regularization coefficient $C$ with 5-fold cross-validation over a logarithmic grid. The VLM parameters are never updated. At inference time, we report the layer that performs best on a held-out validation split, which is separate from the test set. This layer-wise setup lets us ask two questions. First, does the frozen VLM encode enough information to identify the correct answer? Second, where in the network does that information become linearly decodable?

\section{Experiments}
\label{sec:experiments}

\begin{figure*}[t]
    \centering
    \includegraphics[width=\linewidth]{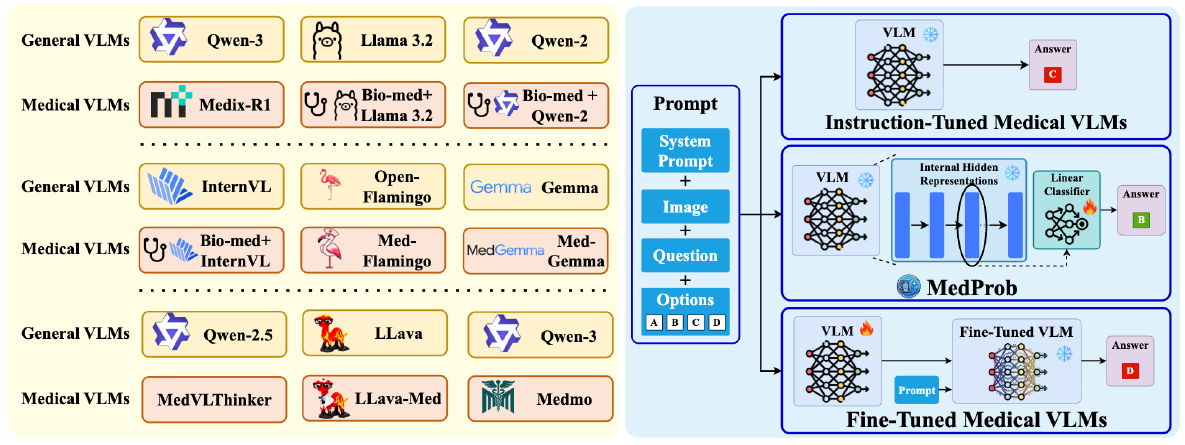}
    \caption{Overview of the MedProb evaluation framework. \textbf{Left}: general-purpose and medical VLMs evaluated, grouped by backbone architecture. \textbf{Right}: prompts consist of a system prompt, medical image, clinical question, and multiple-choice options (A--D), evaluated under three settings. \textit{Top}: instruction-tuned VLMs are prompted directly (frozen). \textit{Middle}: MedProb extracts hidden representations from each transformer layer and decodes answers via a linear classifier (frozen). \textit{Bottom}: fine-tuned VLMs are trained on medical data, then prompted (frozen).}
    \label{fig:overview}
    \vspace{0em}
\end{figure*}

\paragraph{Models.}
To the best of our knowledge, we include all publicly available VLMs, both general-purpose and medically specialized, released up to March 2026 (Figure~\ref{fig:overview}, left). General-purpose models span: LLaVA~\citep{liu2023visual}, OpenFlamingo~\citep{awadalla2023openflamingo}, InternVL3-1B~\citep{zhu2025internvl3}, Llama-3.2-11B-Vision~\citep{grattafiori2024llama3herdmodels}, Qwen2-VL-2B~\citep{wang2024qwen2}, Qwen2.5-VL (3B, 7B, 32B)~\citep{bai2025qwen25vl}, Qwen3-VL (2B, 4B, 8B, 30B)~\citep{bai2025qwen3}, and Gemma (4B, 27B)~\citep{gemma2025gemma3}. Their domain-adapted counterparts include: LLaVA-Med-7B~\citep{li2023llava}, Med-Flamingo-9B~\citep{moor2023med}, BioMed-InternVL3-1B, BioMed-Llama-3.2-11B, BioMed-Qwen2-VL-2B~\citep{cheng-etal-2025-domain}, MedVLThinker-RL (3B, 7B, 32B)~\citep{huang2025medvlthinker}, MedMO (4B, 8B)~\citep{deria2026medmo}, MediX-R1 (2B, 30B)~\citep{mullappilly2026medix}, and MedGemma (4B, 27B)~\citep{sellergren2025medgemma}. A complete listing with medical training datasets is in Table~\ref{tab:model_pairs_datasets}.
We compare only matched medical and general-purpose models with the same backbone and vision encoder, excluding unmatched cases (e.g., LLaVA-Med v1.5 vs. Mistral-7B) to isolate domain adaptation.

\paragraph{Datasets.}
We evaluate on VQA-RAD~\citep{Lau2018ADO}, PathVQA~\citep{He2020PathVQA3Q}, and SLAKE~\citep{Liu2021SlakeAS}. Following \citet{Jeong2024MedicalAO}, we restrict to closed-ended questions, convert binary yes/no questions to multiple-choice format, and construct plausible distractors for non-binary questions; samples with missing images or training-set leakage are removed. PathVQA is subsampled to 500 examples via stratified sampling; VQA-RAD and SLAKE use full test sets. Full dataset statistics and preprocessing details are in Appendix~\ref{sec:dataset_details}. For broader cross-modal generalization, we additionally evaluate on OmniMedVQA~\cite{Hu_2024_CVPR} across eight imaging modalities (Section~\ref{sec:cross_dataset}).

\begin{table*}[t]
\centering
\small
\setlength{\tabcolsep}{4pt}
\renewcommand{\arraystretch}{1.3}
\resizebox{\textwidth}{!}{%
\begin{tabular}{
  >{\raggedright\arraybackslash}m{6.11cm}
  >{\centering\arraybackslash}m{2.2cm}
  *{8}{>{\centering\arraybackslash}m{1.05cm}}
}
\toprule
\multirow{2}{*}{\textbf{Model}} & \multirow{2}{*}{\textbf{Method}}
  & \multicolumn{2}{c}{\textbf{PATH-VQA}}
  & \multicolumn{2}{c}{\textbf{SLAKE}}
  & \multicolumn{2}{c}{\textbf{VQA-RAD}}
  & \multicolumn{2}{c}{\textbf{Average}} \\
\cmidrule(lr){3-4}\cmidrule(lr){5-6}\cmidrule(lr){7-8}\cmidrule(lr){9-10}
 & & F1 & Acc & F1 & Acc & F1 & Acc & F1 & Acc \\
\midrule
MAM~\cite{zhou2025mam}                   & Agentic & 51.85 & 51.80 & 39.92 & 59.51 & 39.31 & 59.31 & 39.62 & 56.87 \\
MMedAgent~\cite{li2024mmedagent}         & Agentic & 36.11 & 37.00 & 28.30 & 43.13 & 59.80 & 58.17 & 41.40 & 46.10 \\
UCAgents~\cite{feng2025ucagents}         & Agentic & 60.55 & 61.19 & 78.09 & 77.10 & 64.17 & 71.10 & 62.36 & 69.80 \\
\midrule\midrule
HuatuoGPT-Vision-34B~\cite{chen2024towards} & Medical VLM        & 60.61 & 61.80 & 77.89 & 76.86 & 83.57 & 75.66 & 74.02 & 71.44 \\
Aloe-Vision-72B-AR~\cite{guaschaloe}        & Medical VLM        & 67.70 & 69.20 & 86.80 & 80.24 & 75.73 & 79.46 & 76.74 & 76.30 \\
UniMedVL-14B~\cite{ning2025unimedvl}        & Medical VLM        & 52.84 & 48.00 & 66.46 & 58.07 & 66.24 & 58.94 & 61.85 & 55.00 \\
\midrule
MedVLM-R1-2B~\cite{medvlm_r1_pan}              & Medical VLM$^\dagger$ & 46.73 & 58.00 & 49.29 & 61.20 & 58.71 & 51.71 & 51.58 & 56.97 \\
InfiMed-RL-3B~\cite{liu2025infimed}         & Medical VLM$^\dagger$ & 79.79 & 79.60 & 88.16 & 81.92 & 83.43 & 74.90 & 83.79 & 78.81 \\
MedMo-8B~\cite{deria2026medmo}              & Medical VLM$^\dagger$ & 65.99 & 67.80 & 80.64 & 72.29 & 73.39 & 58.94 & 73.34 & 66.34 \\
Medix-R1-30B~\cite{mullappilly2026medix}    & Medical VLM$^\dagger$ & 73.00 & 73.00 & 82.46 & 83.61 & 70.32 & 72.24 & 75.26 & 76.28 \\
\midrule\midrule
InternVL3-1B~\cite{zhu2025internvl3}
  & Prompting & 62.01 & 62.80 & 83.68 & 75.66 & 76.12 & 65.02 & 73.94 & 67.83 \\
\rowcolor{blue!10}
InternVL3-1B~\cite{zhu2025internvl3}
  & MedProb   & 82.09 & 82.20 & 88.70 & 83.13 & 81.51 & 72.62 & 84.10 & 79.32 \\
\cmidrule{2-10}
Qwen3-VL-2B~\cite{bai2025qwen3}
  & Prompting & 55.78 & 56.80 & 68.73 & 77.83 & 64.72 & 64.64 & 63.08 & 66.42 \\
\rowcolor{blue!10}
Qwen3-VL-2B~\cite{bai2025qwen3}
  & MedProb   & 83.67 & 83.80 & 89.02 & 84.10 & 80.79 & 73.00 & 84.49 & 80.30 \\
\cmidrule{2-10}
Gemma-4B~\cite{gemma2025gemma3}
  & Prompting & 53.94 & 57.00 & 68.93 & 61.69 & 57.96 & 52.47 & 60.28 & 57.05 \\
\rowcolor{blue!10}
Gemma-4B~\cite{gemma2025gemma3}
  & MedProb   & 86.33 & 86.40 & 86.43 & 79.76 & 80.33 & 72.24 & 84.36 & 79.47 \\
\cmidrule{2-10}
Qwen3-VL-4B~\cite{bai2025qwen3}
  & Prompting & 59.88 & 54.60 & 86.03 & 77.35 & 46.24 & 67.30 & 64.05 & 66.42 \\
\rowcolor{blue!10}
Qwen3-VL-4B~\cite{bai2025qwen3}
  & MedProb   & 85.69 & 85.80 & 90.63 & 86.02 & 83.14 & 74.90 & 86.49 & 82.24 \\
\cmidrule{2-10}
Qwen3-VL-8B~\cite{bai2025qwen3}
  & Prompting & 58.98 & 58.60 & 70.36 & 76.63 & 46.25 & 69.58 & 58.53 & 68.27 \\
\rowcolor{blue!10}
Qwen3-VL-8B~\cite{bai2025qwen3}
  & MedProb   & 86.12 & 86.20 & 89.35 & 84.09 & 82.34 & 75.28 & 85.93 & 81.86 \\
\cmidrule{2-10}
Llama-3.2-11B-Vision~\cite{grattafiori2024llama3herdmodels}
  & Prompting & 48.94 & 59.80 & 68.43 & 69.88 & 67.60 & 68.06 & 61.66 & 65.91 \\
\rowcolor{blue!10}
Llama-3.2-11B-Vision~\cite{grattafiori2024llama3herdmodels}
  & MedProb   & 86.12 & 86.20 & 88.21 & 82.40 & 83.89 & 76.04 & 86.07 & 81.55 \\
\cmidrule{2-10}
Gemma-27B~\cite{gemma2025gemma3}
  & Prompting & 61.18 & 62.00 & 73.12 & 75.42 & 69.56 & 69.58 & 67.95 & 69.00 \\
\rowcolor{blue!10}
Gemma-27B~\cite{gemma2025gemma3}
  & MedProb   & \textbf{86.86} & \textbf{87.00} & 89.17 & 82.89 & 83.14 & 76.04 & 86.39 & 81.98 \\
\cmidrule{2-10}
Qwen3-VL-30B~\cite{bai2025qwen3}
  & Prompting & 63.23 & 61.20 & 88.04 & 80.96 & 74.53 & 77.95 & 75.27 & 73.37 \\
\rowcolor{blue!10}
Qwen3-VL-30B~\cite{bai2025qwen3}
  & MedProb   & 85.70 & 85.80 & \textbf{91.12} & \textbf{86.75} & \textbf{86.46} & \textbf{81.37} & \textbf{87.76} & \textbf{84.64} \\
\bottomrule
\end{tabular}%
}\caption{Performance comparison on medical VQA benchmarks.
  \textbf{Bold}: best per column;
  {\colorbox{blue!10}{\strut MedProb}} rows are highlighted in blue;
  $^\dagger$Trained on PATH-VQA, VQA-RAD, and SLAKE.}
\label{tab:main_results}
\vspace{0em}
\end{table*}

\paragraph{Baselines.}
We compare MedProb against three categories of baselines, corresponding to the three settings in Figure~\ref{fig:overview} (right).
\textbf{Agentic frameworks}: MAM~\cite{zhou2025mam}, MMedAgent~\cite{li2024mmedagent}, and UCAgents~\cite{feng2025ucagents} orchestrate multiple models or tools through multi-step reasoning pipelines.
\textbf{Instruction-tuned medical VLMs} (frozen, directly prompted): HuatuoGPT-Vision-34B~\cite{chen2024towards}, Aloe-Vision-72B-AR~\cite{guaschaloe}, and UniMedVL-14B~\cite{ning2025unimedvl}, instruction-tuned on large-scale biomedical data to generalize to unseen tasks.
\textbf{Benchmark-trained medical VLMs} (instruction-tuned on benchmark data, then prompted frozen): InfiMed-RL-3B~\cite{liu2025infimed}, MedVLM-R1~\cite{medvlm_r1_pan}, MedMO (4B, 8B)~\cite{deria2026medmo}, and MediX-R1 (2B, 30B)~\cite{mullappilly2026medix}.
These models are instruction-tuned with access to training splits of our evaluation benchmarks; they are included as upper-bound references for supervised specialization.
The effect of fine-tuning on the probing gap is analyzed separately in Section~\ref{sec:finetune}.
None of the baseline medical VLMs share backbone architecture with our probed models; they are included to situate MedProb within the broader landscape of specialized systems. All baselines are re-evaluated from scratch on our identical test splits; inference configurations are in Appendix~\ref{sec:baseline_details}.

MedProb uses up to 1,000 labeled training examples to fit the linear probe, while prompting baselines are zero-shot. This is intentional: our central claim is that even this minimal supervised elicitation, a single logistic regression layer with no VLM parameter updates, consistently outperforms zero-shot prompting and extensively fine-tuned medical VLMs, cleanly separating \emph{knowledge representation} from \emph{knowledge elicitation}.

\paragraph{Evaluation Metrics.}
Following \citet{Jeong2024MedicalAO}, we report macro-averaged F1 and accuracy. F1 is computed over two classes for binary questions and macro-averaged across all answer categories for non-binary questions. Prompting baselines use zero-shot evaluation; MedProb trains a linear probe on the training split without updating any VLM parameters.

\paragraph{Implementation Details.}
We extract the hidden state of the \emph{last input token} at every transformer layer of the frozen VLM. Features are standardized using training-set mean and variance. An independent multinomial logistic regression probe is trained per layer, with regularization coefficient $C$ selected via 5-fold cross-validation over $\{10^{-4}, \ldots, 10^{4}\}$ (max 1,000 iterations per fold). The best-performing layer is selected on a held-out validation split strictly disjoint from the test set. For the prompting baseline, we follow each model's official evaluation protocol; full details are in Appendix~\ref{sec:prompting_details}.

\section{Results and Analysis}
\label{sec:results}

\paragraph{Main Results.}
\label{sec:main_results}

Table~\ref{tab:main_results} presents a comprehensive comparison of MedProb against agentic frameworks and medical VLM baselines across all three benchmarks.
MedProb consistently and substantially outperforms both categories.
Among agentic systems, UCAgents achieves the strongest average accuracy at 69.80\%, while MMedAgent trails significantly at 46.10\%.
Medical VLMs fare better: InfiMed-RL-3B reaches an average F1 of 83.79\% and Aloe-Vision-72B-AR achieves 76.74\%; yet even these large-scale, instruction-tuned systems fall short of what linear probing extracts from frozen general-purpose VLMs.
Notably, the smallest probing model, InternVL3-1B, already surpasses most agentic and several medical VLM baselines with an average accuracy of 79.32\%, despite having far fewer parameters and no domain-specific training.
To rule out generation artifacts as the explanation for the prompting gap, we additionally evaluate option-likelihood scoring and conduct paired McNemar significance tests; full results and analysis are in Appendix~\ref{sec:option_likelihood_and_significance}.
Few-shot (3-shot) prompting results are provided in Appendix~\ref{sec:fewshot}.

{The probing results also reveal a clear scaling trend: larger models often yield higher linear-probe accuracy on these benchmarks, with dataset-dependent gains.}
Within the Qwen3-VL family, average accuracy grows from 80.30\% at 2B to 82.24\% at 4B and 84.64\% at 30B.
The best overall results are achieved by Qwen2.5-VL-32B, which sets the top scores across SLAKE (F1: 91.60, Acc: 87.46), VQA-RAD (F1: 92.33, Acc: 88.59), and the overall average (F1: 89.94, Acc: 87.35).
Similarly, within the Gemma family, scaling from 4B to 27B improves PATH-VQA F1 from 86.33\% to 86.86\%.

\begin{figure}[t]
    \centering
    \includegraphics[width=\linewidth]{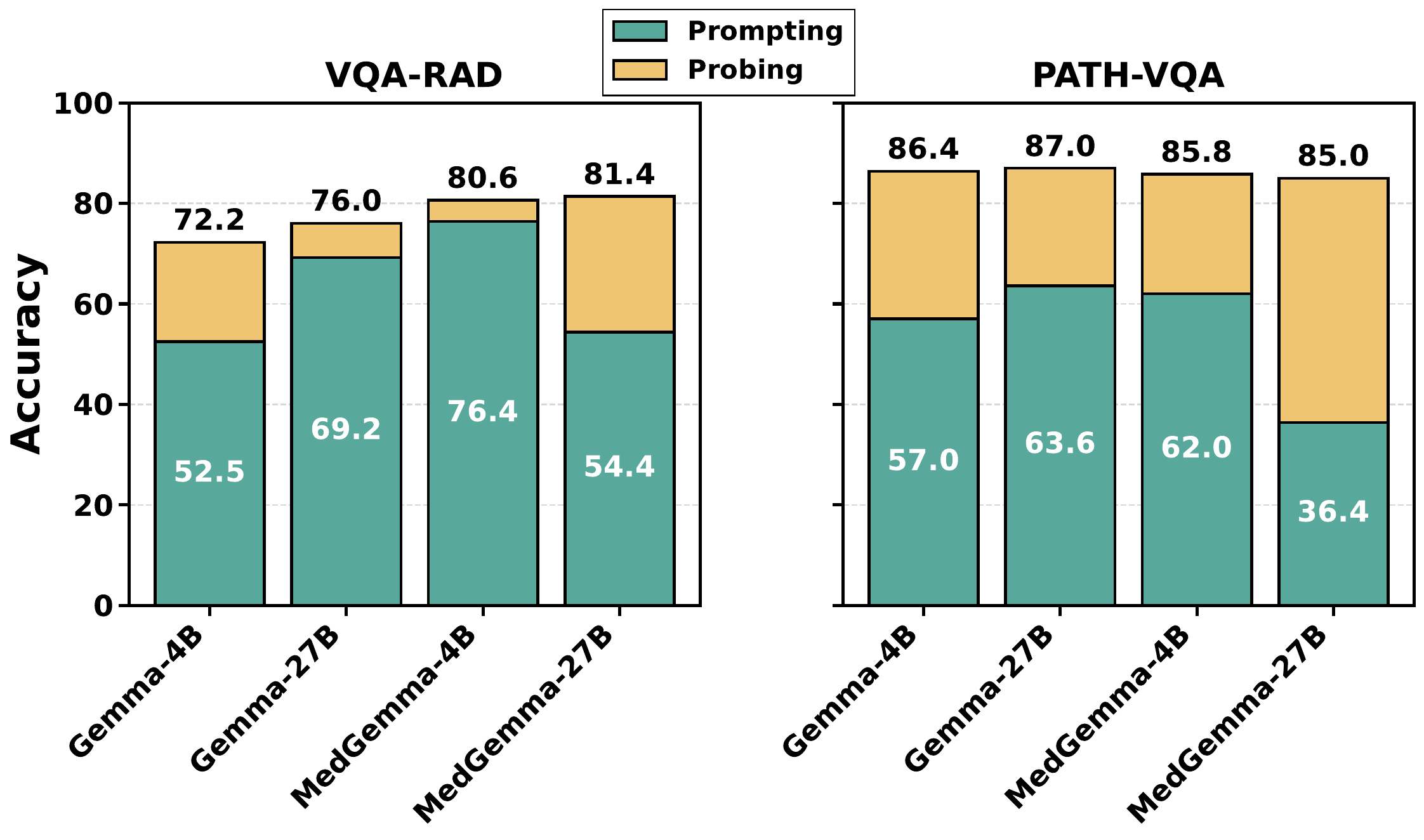}
    \caption{Prompting vs.\ MedProb accuracy for general (Gemma-4B/27B) and medical (MedGemma-4B/27B) VLMs on VQA-RAD and PATH-VQA.}
    \label{fig:stacked_accuracy}
   
\end{figure}

Figure~\ref{fig:stacked_accuracy} directly contrasts probing and prompting for four representative general-purpose models on VQA-RAD and PATH-VQA.
On VQA-RAD, the probing advantage ranges from a clear gain for Gemma-27B (76.0\% vs.\ 69.2\%) to a pronounced improvement for Gemma-4B (72.2\% vs.\ 52.5\%).
The gap is far more dramatic on PATH-VQA, where prompting accuracy collapses to between 57.2\% and 63.6\% for all four models, while MedProb consistently achieves 82.8\%--87.0\% for the same models, improvements exceeding 20 percentage points in several cases.
This stark contrast on a pathology-focused benchmark strongly indicates that the clinical knowledge needed to interpret pathological images is richly encoded in these models' internal representations, yet systematically suppressed by the generation process.

{
\paragraph{Binary vs.\ Multiclass Evaluation.}
\label{sec:binary_multiclass}
We separate binary yes/no items from true multiclass items (three or more options). Most non-yes/no items on SLAKE and VQA-RAD are still effectively 2-way, so we do not treat them as multiclass. We report a dedicated true-multiclass evaluation on OmniMedVQA (4-way; chance 25\%), where probing outperforms prompting overall; binary gains remain large on PATH-VQA, SLAKE, and VQA-RAD. Full per-model binary and multiclass breakdowns are in Appendix~\ref{app:binary_multiclass}.
}

\paragraph{Effect of Fine-tuning on the Probing Gap.}
\label{sec:finetune}

\begin{figure}[t]
    \centering
    \includegraphics[width=\linewidth]{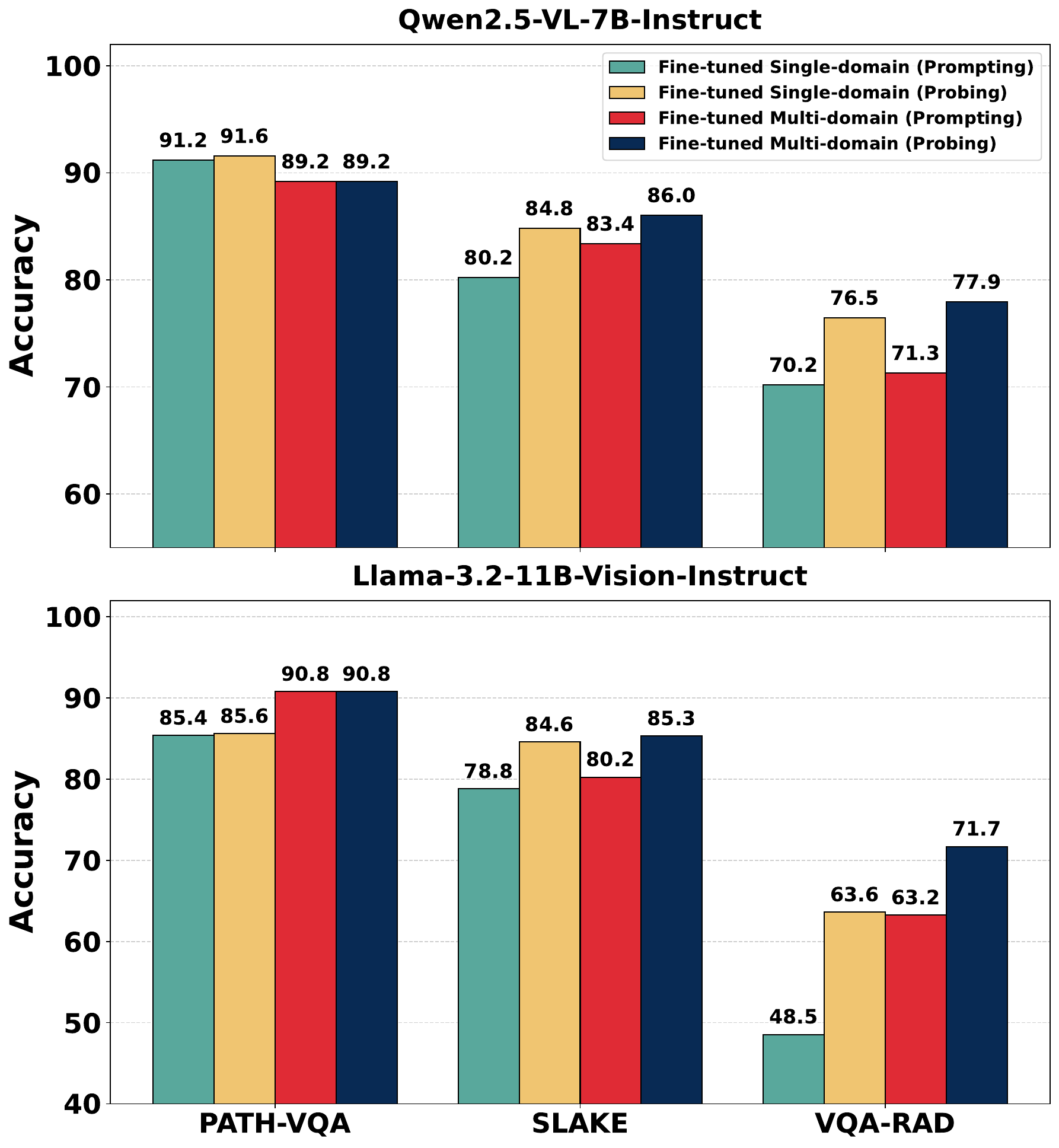}
    \caption{Probing vs.\ prompting accuracy after single-domain and multi-domain fine-tuning.}
    \label{fig:finetune_ablation}
    \vspace{0em}
\end{figure}

Figure~\ref{fig:finetune_ablation} examines whether fine-tuning the underlying VLM narrows the gap between prompting and probing. {We compare both the zero-shot and fine-tuned settings: the fine-tuned VLM in this figure is trained on the entire training set for each dataset, and the probe is then trained on only a 1{,}000-example subset of that same data, so the probe never sees additional labeled examples beyond what the fine-tuned generation model already used.}
Fine-tuning the base VLM does not close the probing gap.
After in-domain fine-tuning, the probe still outperforms prompting on average across the two evaluated models on SLAKE (84.7\% vs.\ 79.5\%) and VQA-RAD (70.3\% vs.\ 59.5\%). PATH-VQA is the only exception, where the binary yes/no structure allows prompted generation to match probing.
Under multi-domain fine-tuning the same pattern holds, and fine-tuning yields only marginal gains for prompting across most settings, confirming that the elicitation mechanism, not the encoded knowledge, remains the bottleneck.
{Stronger supervision for prompting, few-shot in-context examples or fine-tuning on the same labeled data used by the probe, improves generation in some settings but does not eliminate the probe-prompt gap. We interpret the main zero-shot comparison as a representation diagnostic, with Figure~\ref{fig:finetune_ablation} as the primary apples-to-apples supervised control.}

\begin{figure}[t]
    \centering
    \includegraphics[width=\linewidth]{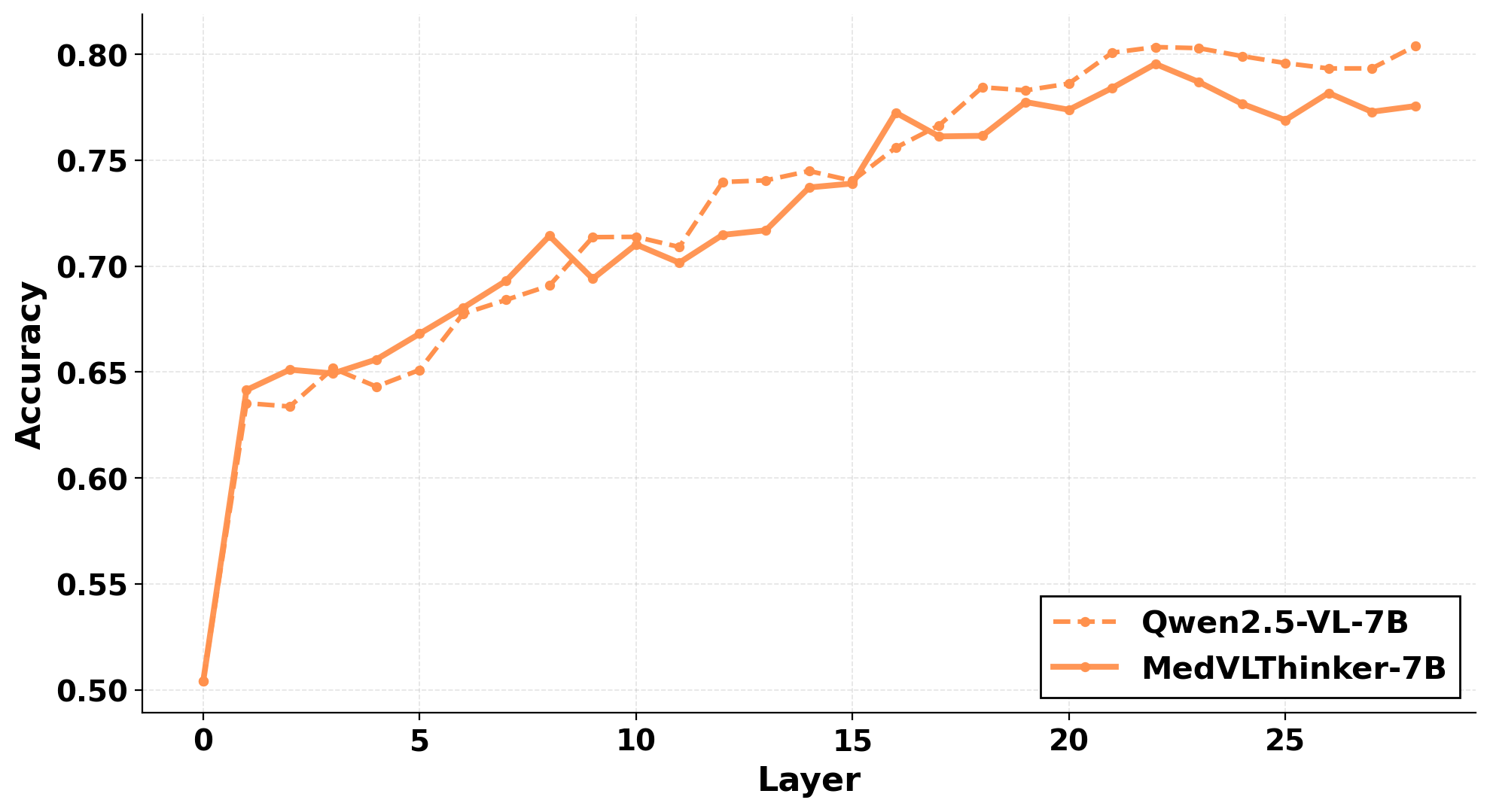}
    \caption{Layer-wise probing accuracy for Qwen2.5-VL-7B and MedVLThinker-7B.}
    \label{fig:layerwise}
    \vspace{0em}
\end{figure}

\paragraph{Layer-wise Probing Analysis.}
\label{sec:layerwise}
Figure~\ref{fig:layerwise} shows that probing accuracy rises monotonically with layer depth.
Both models start near chance at layer~0 ($\approx$50.4\%) and improve steadily, with the most clinically informative signal concentrated in the deeper layers: Qwen2.5-VL-7B peaks at 80.4\% at layer~28 and MedVLThinker-7B at 79.5\% at layer~22, after which the medical model's accuracy declines.
MedVLThinker-7B holds a slight edge in early-to-middle layers (71.4\% vs.\ 69.1\% at layer~8), but the general model overtakes it from layer~9 onward, indicating that medical adaptation accelerates early-layer encoding but fails to produce richer representations in the deeper layers that matter most.

\paragraph{Robustness to Answer Option Ordering.}
\label{sec:order_robustness}

\begin{figure}[t]
    \centering
    \includegraphics[width=\linewidth]{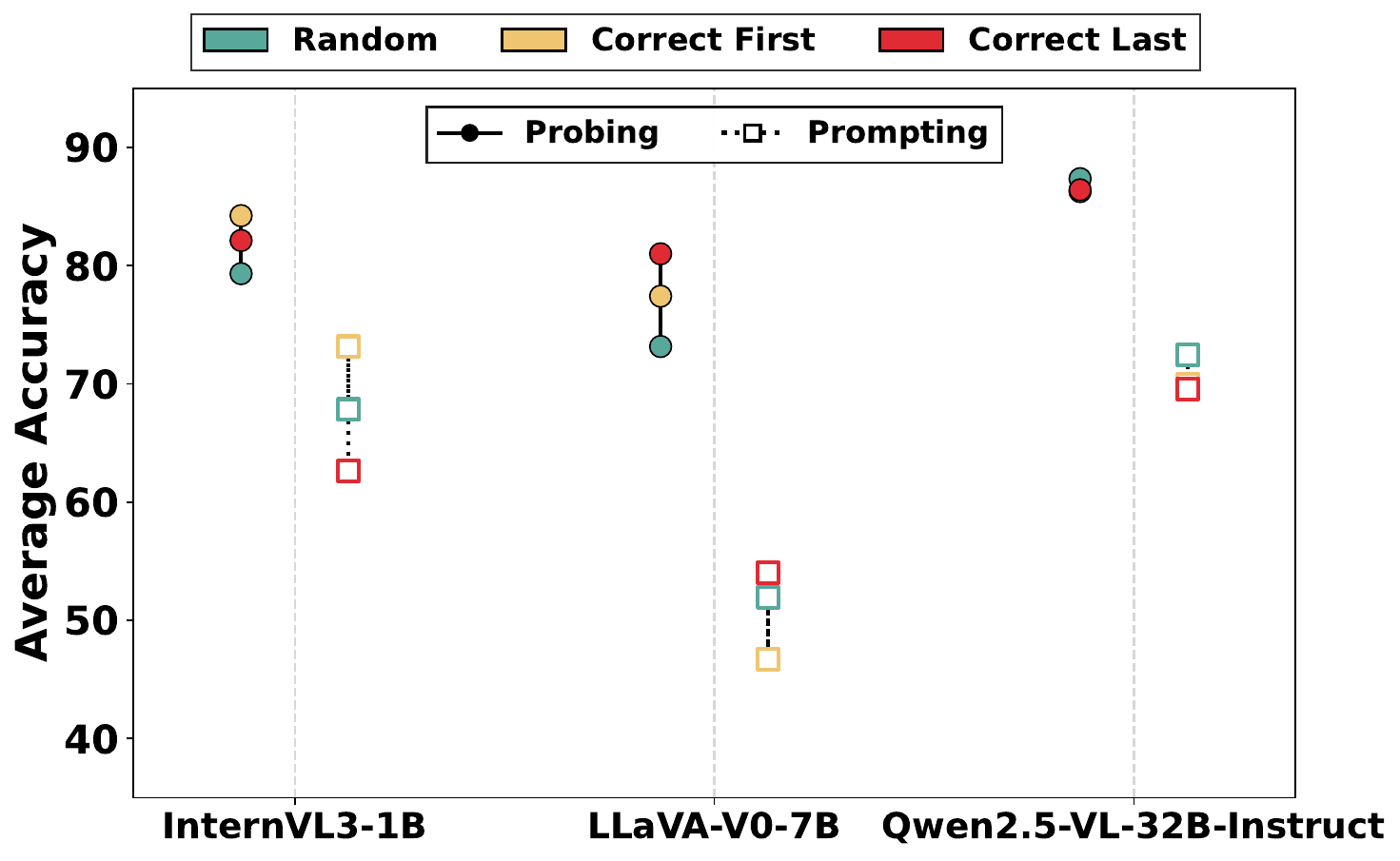}
    \caption{Probing vs.\ prompting accuracy under random, correct-first, and correct-last answer orderings.}
    \label{fig:order_robustness}
    \vspace{0em}
\end{figure}

A critical concern with prompting-based evaluation is positional bias~\cite{pezeshkpour2024large}.
Figure~\ref{fig:order_robustness} reports accuracy for InternVL3-1B, LLaVA-V0-7B, and Qwen2.5-VL-32B under three conditions: random (standard balanced evaluation), correct-answer first, and correct-answer last.

MedProb is far less sensitive to answer position than prompting.
The first-vs-last accuracy gap for probing is 2.10 pp (InternVL3-1B), 3.58 pp (LLaVA-V0-7B), and 0.14 pp (Qwen2.5-VL-32B), compared to 10.49 pp, 7.33 pp, and 0.44 pp respectively under prompting.
Prompting further exhibits inconsistent, model-dependent biases: LLaVA-V0-7B favors correct-last (recency effect, $+$7.33 pp), while InternVL3-1B favors correct-first (primacy effect, $+$10.49 pp).
The random condition, which averages over balanced orderings and reflects standard reported results, falls within the prompting range but does not resolve these instabilities.
{MedProb also has positional bias, however, it is impacted differently than prompting: because it reads directly from internal activations without generating text, it is far less sensitive to option order, but it is not free of positional effects. We also found that the lower accuracy of the uniform-random ordering condition partly reflects that the original training data is not itself uniformly randomized with respect to option order, so both probing and prompting can learn some ordering-dependent preference.}

\paragraph{General vs.\ Medical Model Comparison.}
\label{sec:gen_vs_med}

Table~\ref{tab:gen_vs_med_names} reports the probing accuracy delta ($\Delta$) between each general-purpose model and its medically adapted counterpart.

\begin{table}[t]
\centering
\scriptsize
\setlength{\tabcolsep}{3pt}
\renewcommand{\arraystretch}{1.08}
\resizebox{\columnwidth}{!}{%
\begin{tabular}{llc}
\toprule
\textbf{General Model} & \textbf{Medical Model} & \textbf{$\Delta$} \\
\midrule
Gemma-4B            & MedGemma-4B         & \textcolor{red}{$-$4.92} \\
Gemma-27B           & MedGemma-27B        & \textcolor{red}{$-$0.95} \\
Open-Flamingo-9B    & Med-Flamingo-9B     & \textcolor{red}{$-$1.93} \\
InternVL3-1B        & BioMed-InternVL3-1B & \textcolor{green!60!black}{$+$2.36} \\
LLaVA-7B            & LLaVA-Med-7B        & \textcolor{green!60!black}{$+$0.17} \\
LLaMA3.2-11B-Vision & BioMed-LLaMA3.2-11B & \textcolor{red}{$-$2.30} \\
Qwen2-VL-2B         & BioMed-Qwen2-VL-2B  & \textcolor{green!60!black}{$+$0.05} \\
Qwen2.5-VL-3B       & MedVLThinker-RL-3B  & \textcolor{green!60!black}{$+$2.16} \\
Qwen2.5-VL-7B       & MedVLThinker-7B     & \textcolor{green!60!black}{$+$1.15} \\
Qwen2.5-VL-32B      & MedVLThinker-RL-32B & \textcolor{green!60!black}{$+$1.65} \\
Qwen3-VL-2B         & MediX-R1-2B         & \textcolor{green!60!black}{$+$11.31} \\
Qwen3-VL-4B         & MedMo-4B            & \textcolor{red}{$-$1.65} \\
Qwen3-VL-8B         & MedMo-8B            & \textcolor{red}{$-$0.19} \\
Qwen3-VL-30B        & MediX-R1-30B        & \textcolor{green!60!black}{$+$3.72} \\
\bottomrule
\end{tabular}%
}
\caption{Probing accuracy delta ($\Delta$) between matched general/medical pairs (averaged across benchmarks). Positive: medical better; negative: general better.}
\label{tab:gen_vs_med_names}

\end{table}

There is no consistent probing improvement from general to medical models.
Out of 14 matched pairs, 6 show the general model outperforming its medical counterpart, and medical instruction-tuning can actively degrade representations: MedGemma-4B trails Gemma-4B by 4.92 pp, BioMed-LLaMA3.2-11B trails its base by 2.30 pp, and where adaptation does help, gains are often negligible (LLaVA-Med: $+$0.17 pp; BioMed-Qwen2-VL-2B: $+$0.05 pp).
The exception is MediX-R1, which uses RL-based adaptation rather than standard SFT and outperforms its base by 11.31 pp (2B) and 3.72 pp (30B), suggesting RL training may more effectively shape internal clinical representations.

\paragraph{Ablation and Shuffle Controls.}
\label{sec:image_ablation}
A natural concern is that the probe could rely on textual regularities alone rather than genuinely integrating visual information. For two representative general-purpose VLMs, Llama-3.2-11B-Vision and Qwen3-VL-8B, we retrained the same linear probe under two image-ablation settings: (1) a text-only / image-ablated control that replaces every image with a blank black image, and (2) an image-shuffled control that pairs each question with a mismatched image from the same split. Table~\ref{tab:image_ablation} reports best-layer accuracy for the original probe and both controls.

\begin{table}[t]
\centering
\resizebox{\columnwidth}{!}{%
\begin{tabular}{llcc}
\toprule
\textbf{Model} & \textbf{Method} & \textbf{SLAKE} & \textbf{VQA-RAD} \\
\midrule
\multirow{3}{*}{Llama-3.2-11B-Vision}
 & MedProb & \textbf{82.4} & \textbf{76.0} \\
 & Text-only (image-ablated) & 67.2 & 68.4 \\
 & Image-shuffled & 66.5 & 65.8 \\
\midrule
\multirow{3}{*}{Qwen3-VL-8B}
 & MedProb & \textbf{84.1} & \textbf{75.3} \\
 & Text-only (image-ablated) & 67.7 & 68.4 \\
 & Image-shuffled & 65.8 & 69.1 \\
\bottomrule
\end{tabular}%
}
\caption{Best-layer probe accuracy (\%) with the original image, an image-ablated (blank) control, and an image-shuffled (mismatched) control.}
\label{tab:image_ablation}
\vspace{0em}
\end{table}

Accuracy decreases by 6.2 to 18.3 percentage points when images are shuffled relative to the original probe, indicating that the probe depends on both the image and the question rather than on textual regularities alone.

\paragraph{Error Analysis.}
\label{sec:error_analysis}

To understand where MedProb succeeds and where it falls short relative to prompting,
we conducted a targeted qualitative analysis over a stratified sample of model outputs.
We selected 10 instances from each of the three evaluation benchmarks, yielding 30 samples per model,
across all 10 evaluated models, for a total of 300 manually examined responses.
For each sample, we inspected the model's full generated output alongside the probe's prediction
and the ground-truth label, classifying each case into one of three behavioral patterns.
A detailed presentation of representative examples, including verbatim model outputs, is provided in
Appendix~\ref{app:error_analysis}.

\begin{figure}[t]
\centering
\begin{tcolorbox}[
    title=Prompt model output (idx 440),
    colback=gray!5,
    colframe=blue!50!black,
    boxrule=0.6pt,
    arc=2mm
]
\small
``Based on the image, the answer is \textbf{A: yes}.
The image shows a lobulated mass with a bluish cartilaginous hue infiltrating
the soft tissues.
This is a characteristic appearance of a cartilaginous tumor.''

\par\smallskip

\textit{
Prompt prediction: \textcolor{red}{\textbf{A (Yes)} $\times$}
\quad
Probe prediction: \textcolor{teal!70!black}{\textbf{B (No)} \checkmark}
\quad
GT: \textcolor{teal!70!black}{\textbf{B (No)}}
}
\end{tcolorbox}

\caption{Semantic hallucination failure: prompted model identifies real visual features but draws a false clinical conclusion; probe recovers the correct answer from latent activations.}
\label{fig:e1_main}

\end{figure}

\paragraph{Where does MedProb improve over prompting?}
MedProb improves performance primarily on visually grounded binary classification tasks
where free-form generation introduces hallucinated medical reasoning.
In these cases, internal representations encode the correct answer signal,
but language generation leads the model to over-interpret image features
and reason itself into the wrong answer.
This failure pattern recurs consistently across modalities:
in radiology, the prompted model applies textbook logic correctly but misattributes it to the wrong scan;
in histopathology, it generates plausible-sounding but incorrect tissue-level narratives.
{Because the linear probe reads directly from latent activations without generating a reasoning chain, it is not subject to this ``hallucinated chain-of-thought'' failure mode, though it remains susceptible to other sources of error discussed below.}
MedProb helps most on questions requiring straightforward visual feature matching, organ presence,
gross pathological category, imaging modality, where intermediate-layer activations already encode
the correct signal but the output layer is misled by overconfident medical language generation.
Figure~\ref{fig:e1_main} illustrates the most common failure mode corrected by the probe.

As shown in Figure~\ref{fig:e1_main}, the prompted model correctly identifies real visual features
(lobulated shape, cartilaginous hue) but constructs a false pathology narrative from them and
answers \emph{yes}; the probe at layer~16 reads directly from latent activations and is
unaffected by this hallucinated reasoning chain.

When both methods fail simultaneously, the deficit lies in the model's perceptual encoding rather
than the elicitation strategy; when prompting outperforms probing, the task demands compositional
clinical reasoning or grounded semantic labels that exceed what a linear probe can decode from a
single layer (see Appendix~\ref{app:error_analysis} for detailed examples of both patterns).

\paragraph{Sample Efficiency.}

\begin{figure}[t]
    \centering
    \includegraphics[width=\linewidth]{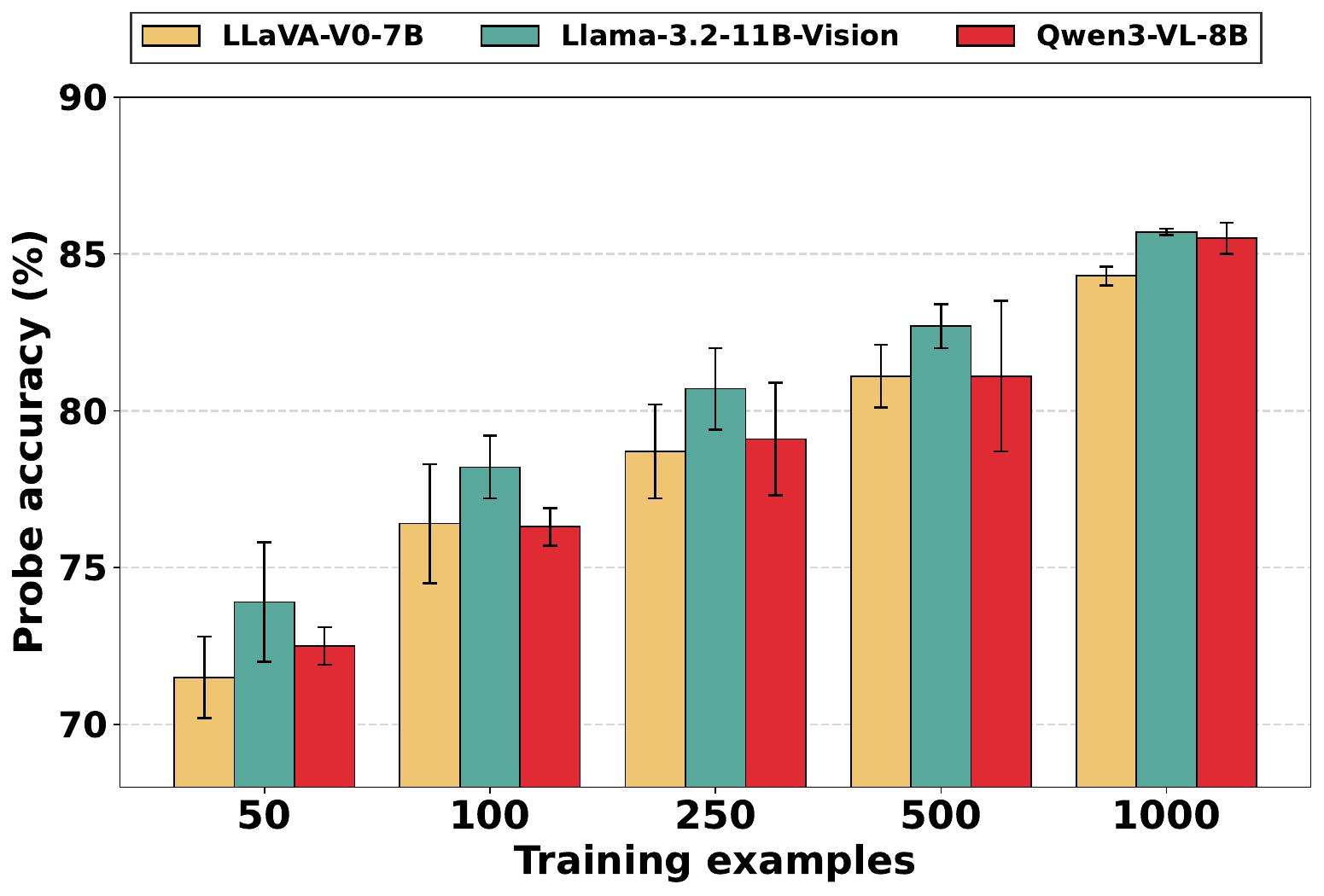}
    \caption{PATH-VQA probe accuracy (\%) vs.\ training-set size (mean\,$\pm$\,std over different random seeds).}
    \label{tab:learning_curve}
    \vspace{0em}
\end{figure}

We measured how probe accuracy scales with labeled training data by retraining on progressively larger subsets (50, 100, 250, 500, 1{,}000 examples), averaging over different random seeds.
Figure~\ref{tab:learning_curve} shows that all three models improve monotonically, gaining approximately 12 percentage points from 50 to 1{,}000 training examples.
Crucially, even at 50 examples all three models exceed 71\% accuracy, establishing that strong clinical decoding is achievable with limited supervision, a practically important property for low-resource medical settings.
Gains diminish above 500 examples, suggesting that 500--1{,}000 labeled samples is a sufficient operating range for MedProb.

\section{Conclusion}
\label{sec:conclusion}

We introduced MedProb, a lightweight linear probing framework that decodes clinical answers directly from the frozen internal representations of general-purpose VLMs, outperforming medical VLMs and multi-agent pipelines across PATH-VQA, SLAKE, and VQA-RAD with only a logistic regression classifier trained on a small labeled set. Across 14 matched architecture pairs, domain-adaptive pretraining does not reliably enrich internal clinical representations, with general models outperforming their medical counterparts in 6 of 14 cases and RL-based adaptation (MediX-R1) the notable exception. MedProb is also substantially more robust to answer option ordering than prompting, which exhibits inconsistent positional biases across models. Together, these findings reframe the Med-VQA challenge: the bottleneck is not what VLMs know, but how that knowledge is elicited. Future work should explore multi-layer aggregation, nonlinear probes, and open-ended VQA settings to further unlock the clinical knowledge latent in general-purpose representations. {As an initial step toward the open-ended setting, we show in Appendix~\ref{app:rejection_sampling} that the probe can rerank open-ended VLM generations in a rejection-sampling framework, though future work should explore this in more depth.}

\section*{Acknowledgments}
This material is based upon work supported by the National Science Foundation (NSF) under Grant~No. 2145357.


\section*{Limitations}

This work has several limitations. First, MedProb requires a small labeled training set, while prompted VLMs can be evaluated zero-shot. This means MedProb should not be interpreted as a purely zero-shot method. However, our goal is not to compare zero-shot prompting with supervised learning in general, but to test whether answer-relevant information is already present in frozen VLM representations and can be recovered with minimal supervision. The probe is intentionally simple, updates no VLM parameters, and uses only a lightweight linear classifier.

Second, our experiments focus primarily on closed-ended Med-VQA benchmarks{, and most non-yes/no items on SLAKE and VQA-RAD are still effectively 2-way rather than true multiclass; we report a dedicated true-multiclass evaluation on OmniMedVQA (4-way) in Appendix~\ref{app:binary_multiclass}}. This setting allows controlled comparison across models and avoids ambiguity in free-text answer matching, but it does not capture the full complexity of open-ended clinical reasoning. We therefore view MedProb as a diagnostic tool for measuring recoverable answer signal, not as a complete clinical question-answering system. Extending the approach to open-ended VQA, report generation, and more complex clinical workflows is an important direction for future work{; as an initial step in this direction, Appendix~\ref{app:rejection_sampling} shows the probe can score open-ended candidate generations in a rejection-sampling framework}.

Third, linear probing measures what is linearly decodable from a chosen representation, not everything a model may know. A failure of the probe does not necessarily mean that the information is absent from the model; it may be encoded nonlinearly, distributed across layers, or dependent on richer aggregation strategies. This limitation makes our results conservative: despite using a simple probe, MedProb still recovers substantially more task-relevant signal than prompting across benchmarks. {Following \citet{belinkov2022probing}, we treat MedProb as a diagnostic tool for assessing representational content rather than model behavior. High probing accuracy means that closed-ended answer labels are linearly recoverable from frozen VLM hidden states on these benchmarks. It does not by itself show that the decoded directions are causally responsible for the model's predictions, nor that they correspond to clinically meaningful knowledge in a causal sense.}

Finally, MedProb improves answer prediction but does not itself provide faithful clinical explanations or evidence grounding. This is by design. Our central claim is that generation-based evaluation can underestimate the medical VQA signal present in VLM representations. For deployment-facing systems, probing would need to be combined with calibrated uncertainty estimates, evidence grounding, and clinical validation. These limitations do not change the main finding: much of the answer-relevant signal needed for Med-VQA is already present in frozen VLMs, and the main bottleneck is often how that signal is elicited.

\bibliography{custom}

\appendix

\section{Dataset Details}
\label{sec:dataset_details}
\begin{table}[ht]
\centering
\resizebox{\columnwidth}{!}{%
\begin{tabular}{lcccc}
\toprule
\textbf{Dataset} & \textbf{Split} & \textbf{Size} & \textbf{Yes/No} & \textbf{Multiple Choice} \\
\midrule
\multirow{4}{*}{VQA-RAD}
  & Train    & 820  & 753  & 67  \\
  & Validation & 205  & 187  & 18  \\
  & Test     & 272  & 251  & 21  \\
\cmidrule(lr){2-5}
  & \textit{Total} & \textit{1297} & \textit{1191} & \textit{106} \\
\midrule
\multirow{4}{*}{PATH-VQA}
  & Train    & 9804 & 9804 & 0 \\
  & Validation & 3135 & 3135 & 0 \\
  & Test     & 3389 & 3389 & 0 \\
\cmidrule(lr){2-5}
  & \textit{Total} & \textit{16328} & \textit{16328} & \textit{0} \\
\midrule
\multirow{4}{*}{SLAKE}
  & Train    & 1943 & 1681 & 262 \\
  & Validation & 422  & 358  & 64  \\
  & Test     & 415  & 355  & 60  \\
\cmidrule(lr){2-5}
  & \textit{Total} & \textit{2780} & \textit{2394} & \textit{386} \\
\midrule
OmniMedVQA & \textit{Total} & \textit{88,996} & \textit{402} & \textit{88,594} \\
\bottomrule
\end{tabular}%
}
\caption{Summary of the number of close-ended examples in the train, validation, and test sets of all visual QA datasets.}
\label{tab:dataset_stats}
\end{table}

In this section, we provide additional details on the visual QA datasets. In Table~\ref{tab:dataset_stats}, we summarize 
the number of QA examples included in the train, validation, and test sets of each 
dataset, following the preprocessing steps described in 
Section~\ref{sec:experiments}. For VQA-RAD \cite{Lau2018ADO}, PathVQA 
\cite{He2020PathVQA3Q}, and SLAKE \cite{Liu2021SlakeAS}, we report only the 
closed-ended visual QA examples, as our evaluations are restricted to the closed-ended 
setting. Following \citet{Jeong2024MedicalAO} and prior work, we convert all
closed-ended questions into a consistent multiple-choice format by constructing answer
options directly from the question content. No external distractor generation or
cross-question negative sampling is performed; all candidate options are either
explicitly stated in the question text or follow trivially from the binary answer type.

Concretely, the option construction procedure operates as follows.
For \textbf{binary yes/no questions}, the two options \textit{yes} and \textit{no} are
used as candidates. This covers all closed-ended questions in PATH-VQA (which is
exclusively binary) and the majority of questions in VQA-RAD and SLAKE.
For \textbf{non-binary closed questions} in VQA-RAD and SLAKE, the candidate options
are the explicit answer alternatives already named in the question text. For example,
a question such as \textit{``Which is bigger in this image, left lung or left kidney?''}
is assigned the options \textit{left lung} and \textit{left kidney}; a question such as
\textit{``Is this an MRI or a CT scan?''} is assigned \textit{MRI} and \textit{CT}.
These options are extracted via a curated lookup table that enumerates all
non-binary closed questions occurring in the official VQA-RAD and SLAKE datasets, with
the corresponding answer alternatives read verbatim from the question phrasing,
following the same approach as \citet{Jeong2024MedicalAO}. For SLAKE, a text-parsing
fallback additionally covers residual comparison questions by extracting the two or
three alternatives separated by ``or'' after the first comma in the question string.
Records for which options cannot be unambiguously resolved, or for which the ground-truth
answer does not appear in the extracted option set, are excluded from evaluation.
The full preprocessing pipeline is available in our project repository. For OmniMedVQA \cite{Hu_2024_CVPR}, which is released in two 
portions, open-access and restricted-access, we restrict our use to the open-access 
subset, as the restricted-access portion is not publicly available. Furthermore, since 
OmniMedVQA does not provide an official train/validation/test split, we report only 
the total number of examples for this dataset in Table~\ref{tab:dataset_stats}.

To ensure tractable and comparable evaluations across datasets, we cap each dataset at a maximum of 500 test examples and 1,000 training examples for the probing experiments. Subsets are constructed using stratified sampling to preserve the distribution of question types and difficulty levels across splits, following prior work \cite{Liu2025MedMMVAC, Mishra2025TeamMedAgentsEM}. Concretely, for the test set we use 500 examples from PATH-VQA, 415 from SLAKE (its full test set), and 263 from VQA-RAD, where the original 272 test examples are reduced after excluding questions whose associated images are unavailable. For the training set used in probing, we use 1,000 examples from PATH-VQA, 1,000 from SLAKE, and the full 820 examples from VQA-RAD.

For PATH-VQA specifically, since all closed-ended questions are binary (yes/no), stratified sampling is performed over the answer label to ensure that the 500-sample test subset maintains a balanced yes/no class distribution matching that of the full 3,389-sample test set. The resulting subset preserves an approximately equal proportion of affirmative and negative examples, ensuring that accuracy and macro-F1 scores computed on the subsample are representative of those on the full test set. For OmniMedVQA, the open-access subset spans eight imaging modalities: CT (Computed Tomography), Dermoscopy, Fundus Photography, MRI (Magnetic Resonance Imaging), Microscopy, OCT (Optical Coherence Tomography), X-Ray, and Ultrasound. To ensure balanced representation across all modalities, we construct our subsets via stratified sampling over modalities, selecting 100 examples per modality for training (800 examples in total) and 50 examples per modality for evaluation (400 examples in total).

\vspace{2mm} \noindent \textbf{Dataset Links}
\begin{itemize}
    \setlength\itemsep{0pt}
    \setlength\parsep{0pt}
    \item \textbf{VQA-RAD}: \url{https://osf.io/89kps/overview}
    \item \textbf{PathVQA}: \url{https://huggingface.co/datasets/flaviagiammarino/path-vqa}
    \item \textbf{SLAKE}: \url{https://www.med-vqa.com/slake/}
    \item \textbf{OmniMedVQA}: \url{https://openxlab.org.cn/datasets/GMAI/OmniMedVQA}
\end{itemize}

\begin{table}[t]
\centering
\small
\begin{tabular}{ll}
\toprule
\textbf{Model Name} & \textbf{HF Link} \\
\midrule

HuatuoGPT-Vision-34B & \hfrepo{FreedomIntelligence/HuatuoGPT-Vision-34B}{Link} \\
Aloe-Vision-72B-AR & \hfrepo{HPAI-BSC/Aloe-Vision-72B-AR}{Link} \\
InfiMed-RL-3B & \hfrepo{InfiX-ai/InfiMed-RL-3B}{Link} \\
UniMedVL-14B & \hfrepo{General-Medical-AI/UniMedVL}{Link} \\

Gemma-3-4B-IT & \hfrepo{google/gemma-3-4b-it}{Link} \\
Gemma-3-27B-IT & \hfrepo{google/gemma-3-27b-it}{Link} \\
MedGemma-4B-IT & \hfrepo{google/medgemma-4b-it}{Link} \\
MedGemma-27B-IT & \hfrepo{google/medgemma-27b-it}{Link} \\

LLaVA-7B-Delta-v0 & \hfrepo{liuhaotian/LLaVA-7b-delta-v0}{Link} \\
LLaVA-Med-7B-Delta & \hfrepo{microsoft/llava-med-7b-delta}{Link} \\
OpenFlamingo-9B & \hfrepo{openflamingo/OpenFlamingo-9B-deprecated}{Link} \\
Med-Flamingo & \hfrepo{med-flamingo/med-flamingo}{Link} \\

MedVLThinker-3B & \hfrepo{UCSC-VLAA/MedVLThinker-3B-RL\_m23k}{Link} \\
MedVLThinker-7B & \hfrepo{UCSC-VLAA/MedVLThinker-7B-RL\_m23k}{Link} \\
MedVLThinker-32B & \hfrepo{UCSC-VLAA/MedVLThinker-32B-RL\_m23k}{Link} \\

MedMO-4B-Next & \hfrepo{MBZUAI/MedMO-4B-Next}{Link} \\
MedMO-8B-Next & \hfrepo{MBZUAI/MedMO-8B-Next}{Link} \\
MediX-R1-2B & \hfrepo{MBZUAI/MediX-R1-2B}{Link} \\
MediX-R1-30B & \hfrepo{MBZUAI/MediX-R1-30B}{Link} \\

Qwen2.5-VL-3B-Instruct & \hfrepo{Qwen/Qwen2.5-VL-3B-Instruct}{Link} \\
Qwen2.5-VL-7B-Instruct & \hfrepo{Qwen/Qwen2.5-VL-7B-Instruct}{Link} \\
Qwen2.5-VL-32B-Instruct & \hfrepo{Qwen/Qwen2.5-VL-32B-Instruct}{Link} \\
Qwen2-VL-2B-Instruct & \hfrepo{Qwen/Qwen2-VL-2B-Instruct}{Link} \\
Qwen3-VL-2B-Instruct & \hfrepo{Qwen/Qwen3-VL-2B-Instruct}{Link} \\
Qwen3-VL-4B-Instruct & \hfrepo{Qwen/Qwen3-VL-4B-Instruct}{Link} \\
Qwen3-VL-8B-Instruct & \hfrepo{Qwen/Qwen3-VL-8B-Instruct}{Link} \\
Qwen3-VL-30B-A3B-Instruct & \hfrepo{Qwen/Qwen3-VL-30B-A3B-Instruct}{Link} \\

BioMed-Qwen2-VL-2B & \hfrepo{AdaptLLM/biomed-Qwen2-VL-2B-Instruct}{Link} \\
BioMed-Llama-3.2-11B & \hfrepo{AdaptLLM/biomed-Llama-3.2-11B-Vision-Instruct}{Link} \\
BioMed-InternVL3-1B & \hfrepo{AdaptLLM/biomed-InternVL3-1B}{Link} \\
Llama-3.2-11B-Vision-Instruct & \hfrepo{meta-llama/Llama-3.2-11B-Vision-Instruct}{Link} \\

\bottomrule
\end{tabular}
\caption{Vision-language models and their Hugging Face repositories.}
\label{tab:model_hf_repos}
\end{table}

\section{Training process of Med-VLMs}

\begin{table*}[t]
\centering
\scriptsize
\renewcommand{\arraystretch}{1.3}
\begin{tabular}{>{\centering\arraybackslash}p{3.2cm} >{\centering\arraybackslash}p{3.5cm} >{\centering\arraybackslash}p{7.5cm}}
\toprule
\textbf{General Model} & \textbf{Medical Model} & \textbf{Medical Training Datasets} \\
\midrule

Gemma (4B, 27B)~\cite{gemma2025gemma3}
    & MedGemma (4B, 27B)~\cite{sellergren2025medgemma}
    & MIMIC-CXR, CT-US1, MRI-US1, PMC-OA, EyePACS, PAD-UFES-20, Digital Knee X-ray, internal histopathology (${\sim}$32.55M patches), internal dermatology (51K images), SLAKE, VQA-RAD. \\[6pt]

LLaVA-V0-7B~\cite{liu2023visual}
    & LLaVA-Med-7B~\cite{li2023llava}
    & PMC-15M (600K caption pairs; 60K GPT-4 instruction pairs across CXR, CT, MRI, histopathology, and gross pathology). \\[6pt]

OpenFlamingo-9B~\cite{awadalla2023openflamingo}
    & Med-Flamingo-9B~\cite{moor2023med}
    & Medical Textbook dataset (MTB; ${\sim}$0.8M images), PMC-OA (1.3M pairs). \\[6pt]

InternVL3-1B~\cite{zhu2025internvl3}
    & BioMed-InternVL3-1B~\cite{cheng-etal-2025-domain}
    & \multirow{3}{7.5cm}{\centering PubMedVision/PMCRefined (500K caption pairs), 144K synthetic visual instruction tasks.} \\
Llama-3.2-11B Vision~\cite{grattafiori2024llama3herdmodels}
    & BioMed-Llama-3.2-11B~\cite{cheng-etal-2025-domain}
    & \multirow{3}{7.5cm}{\centering PubMedVision/PMCRefined (500K caption pairs), 144K synthetic visual instruction tasks.} \\
    [6pt]

Qwen2-VL-2B~\cite{wang2024qwen2}
    & BioMed-Qwen2-VL-2B~\cite{cheng-etal-2025-domain}
    & \multirow{3}{7.5cm}{\centering PubMedVision/PMCRefined (500K caption pairs), 144K synthetic visual instruction tasks.} \\
[6pt]

Qwen2.5-VL (3B, 7B, 32B)~\cite{bai2025qwen25vl}
    & MedVLThinker-RL (3B, 7B, 32B)~\cite{huang2025medvlthinker}
    & PMC-VQA (115K filtered samples), m23K text-only medical QA (16.5K). \\[6pt]

Qwen3-VL (4B, 8B)~\cite{bai2025qwen3}
    & MedMO (4B, 8B)-Next~\cite{deria2026medmo}
    & MedTrinity (18.5M), MIMIC-CXR, IU-Xray, CheXpert, PMC-VQA, PubMedVision, ROCO-V2, NIH Chest X-ray, DeepLesion, PMC-OA, SLAKE, VQA-RAD, PathVQA. \\[6pt]

Qwen3-VL (2B, 30B)~\cite{bai2025qwen3}
    & MediX-R1 (2B, 30B)~\cite{mullappilly2026medix}
    & PMC-VQA (25K), PathVQA (19.6K), SLAKE (4.9K), VQA-RAD (1.8K). \\

\bottomrule
\end{tabular}
\caption{
    General-to-medical VLMs correspondence and training datasets.
}
\label{tab:model_pairs_datasets}
\end{table*}

Med-VLMs are developed by adapting general-purpose vision-language models through continued pretraining and instruction tuning on curated medical datasets, spanning radiology, pathology, ophthalmology, and dermatology. As summarized in Table~\ref{tab:model_pairs_datasets}, training corpora vary widely in scale and modality, from large-scale image–caption pairs harvested from PubMed Central to GPT-4-generated instruction-following examples and reinforcement learning signals aimed at improving clinical reasoning.

\section{Additional Details on Prompting}
\label{sec:prompting_details}

In this section, we summarize the prompting details used for the general-domain and medical
vision--language models evaluated in our work (Appendix~\ref{subsec:repro}), and the default
prompt formats used for each model family (Appendix~\ref{subsec:formats}).

\subsection{Reproducibility of Prompting Details}
\label{subsec:repro}

In Table~\ref{tab:prompt_summary}, we provide a summary of the prompting details available for
each model family. We share these details to document our best efforts at reproducing the
prompting setups used in our evaluations; in cases where official documentation is absent, we
describe the choices we made in designing these setups. In particular, we focus on three
components: (i)~a \textbf{system prompt}; (ii)~a \textbf{prompt format} for
closed-ended multiple-choice QA tasks; and (iii)~the \textbf{sampling details} governing text
generation (e.g., \texttt{temperature}, \texttt{top\_p}).

\paragraph{Gemma 3 (4B and 27B).}
For both Gemma~3 variants (4B and 27B), we adopt greedy decoding to ensure fully deterministic and reproducible outputs, consistent with standard practice in medical VQA benchmarking and with the generation settings specified in the official Gemma~3 model card and model paper \citep{gemma2025gemma3}. Specifically, greedy decoding means the model always selects the most probable next token at each step rather than sampling stochastically, which eliminates any randomness in the output. We also cap the number of generated tokens at 200, which is sufficient to produce any multiple-choice answer while avoiding unnecessary over-generation. Since generation is deterministic, parameters that control stochastic sampling such as nucleus sampling and top-$k$ filtering are left at their defaults and have no effect. As discussed in Appendix~\ref{subsec:formats}, the Gemma~3 instruction-tuned models are designed to operate with only two dialogue roles, namely the user and the model, and do not support a dedicated system role.\footnote{\url{https://ai.google.dev/gemma/docs/core/prompt-structure}} Accordingly, no system prompt is used, and the system prompt entry in Table~\ref{tab:prompt_summary} is marked as not applicable. The prompt format is marked as not available because it was derived from the MedGemma fine-tuning notebook rather than from a dedicated closed-ended evaluation protocol (see Appendix~\ref{subsec:formats}).

\paragraph{MedGemma (4B and 27B).}
For both MedGemma variants (4B and 27B), we apply the same generation configuration as for Gemma~3, in line with the official MedGemma model card and model paper \citep{sellergren2025medgemma}. Specifically, as with Gemma~3, we use greedy decoding, meaning the model always selects the most probable next token at each step rather than sampling stochastically. This eliminates any randomness in the output, ensuring that results are fully deterministic and reproducible across runs. We also cap the number of generated tokens at 200, which is sufficient to produce any multiple-choice answer while avoiding unnecessary over-generation. Since generation is deterministic, parameters that control stochastic sampling such as nucleus sampling and top-$k$ filtering are left at their defaults and have no effect. Crucially, MedGemma is built on the Gemma~3 architecture rather than PaliGemma, and therefore shares the same two-role chat template with no native support for a system turn.\footnote{\url{https://developers.google.com/health-ai-developer-foundations/medgemma/model-card}} As with Gemma~3, we omit any system prompt, keeping the input minimal and consistent with the architecture's design. This deterministic setup is essential for fair comparison across model families in medical imaging evaluation.

\paragraph{LLaVA-V0 (7B).}
Following \citet{Jeong2024MedicalAO}, we adopt the \texttt{simple\_conv} conversational
template\footnote{\url{https://github.com/microsoft/LLaVA-Med/blob/b9a98a736d2ef05bcf5ff345be6403fb3a664eaf/llava/conversation.py\#L243}}
from the official LLaVA-Med repository~\citep{liu2023visual} as the default prompt format.
We note the following caveats. First, the exact system prompt and conversational format used
in the original evaluation are not documented in the paper or the repository; we therefore
default to the \texttt{simple\_conv} template, which reflects the VICUNA-V0 backbone
\citep{vicuna2023} on which LLaVA-V0 is built. Second, the answer-choice formatting
convention for closed-ended QA is inferred solely from the raw VQA-RAD results
file\footnote{\url{https://github.com/microsoft/LLaVA-Med/blob/b9a98a736d2ef05bcf5ff345be6403fb3a664eaf/llava/eval/eval_metrics/answer-file-llava-zeorshot.jsonl}}
rather than from a dedicated evaluation protocol. Consistent with both our greedy decoding approach for the other model families and the generation settings of \citet{Jeong2024MedicalAO}, we set the temperature to zero and restrict generation to a single beam, ensuring fully deterministic and reproducible outputs;
accordingly, the sampling entry in Table~\ref{tab:prompt_summary} is marked as partially
available.

\begin{table*}[t]
\centering
\small
\setlength{\tabcolsep}{4pt}

\begin{tabularx}{\textwidth}{lXccc}
\toprule
\textbf{Category} & \textbf{Model} & \textbf{System Prompt} & \textbf{Prompt Format} & \textbf{Sampling} \\
\midrule

\multirow{8}{*}{General Models}
 & Gemma (4B, 27B)~\cite{gemma2025gemma3} & \xmark & \xmark & \pmark \\
 & LLaVA-V0-7B~\cite{liu2023visual} & \pmark & \pmark & \pmark \\
 & OpenFlamingo-9B~\cite{awadalla2023openflamingo} & \xmark & \xmark & \xmark \\
 & InternVL3-1B~\cite{zhu2025internvl3} & \xmark & \cmark & \pmark \\
 & Llama-3.2-11B-Vision~\cite{grattafiori2024llama3herdmodels} & \xmark & \pmark & \xmark \\
 & Qwen3-VL (2B, 4B, 8B, 30B)~\cite{bai2025qwen3} & \xmark & \cmark & \cmark \\
 & Qwen2.5-VL (3B, 7B, 32B)~\cite{bai2025qwen25vl} & \xmark & \xmark & \xmark \\
 & Qwen2-VL-2B~\cite{wang2024qwen2} & \xmark & \xmark & \xmark \\
\midrule

\multirow{9}{*}{Medical Models}
 & MedGemma (4B, 27B)~\cite{sellergren2025medgemma} & \xmark & \pmark & \pmark \\
 & LLaVA-Med-7B~\cite{li2023llava} & \pmark & \pmark & \pmark \\
 & Med-Flamingo-9B~\cite{moor2023med} & \pmark & \pmark & \xmark \\
 & BioMed-Llama-3.2-11B~\cite{cheng-etal-2025-domain} & \xmark & \cmark & \pmark \\
 & BioMed-Qwen2-VL-2B~\cite{cheng-etal-2025-domain} & \xmark & \cmark & \pmark \\
 & BioMed-InternVL3-1B~\cite{cheng-etal-2025-domain} & \xmark & \cmark & \pmark \\
 & MedVLThinker-RL (3B, 7B, 32B)~\cite{huang2025medvlthinker} & \cmark & \cmark & \cmark \\
 & MedMO (4B, 8B)-Next~\cite{deria2026medmo} & \xmark & \cmark & \xmark \\
 & MediX-R1 (2B, 30B)~\cite{mullappilly2026medix} & \cmark & \cmark & \cmark \\
\bottomrule
\end{tabularx}

\caption{
Summary of prompting details available for each evaluated model.
\cmark\ = fully provided;
\pmark\ = partially provided;
\xmark\ = not provided or not applicable.
}
\label{tab:prompt_summary}
\end{table*}

\paragraph{LLaVA-Med (7B).}
Following \citet{Jeong2024MedicalAO}, we adopt the \texttt{simple\_conv\_med}
template\footnote{\url{https://github.com/microsoft/LLaVA-Med/blob/b9a98a736d2ef05bcf5ff345be6403fb3a664eaf/llava/conversation.py\#L257}}
from the official repository~\citep{li2023llava} as the default prompt format. The caveats
regarding undocumented evaluation choices and inferred answer-choice formatting described
for LLaVA-V0 above apply here as well; given this additional uncertainty, we also search
over prompt formats for model-specific prompt selection. We select the
\texttt{simple\_conv\_med} template specifically because it includes a system prompt
tailored to LLaVA-Med (beginning \texttt{``You are LLaVA-Med, a large language and vision
assistant trained by a group of researchers at Microsoft\ldots''}) and follows the same
VICUNA-V0 conversational structure~\citep{vicuna2023}. We apply the identical generation
configuration as for LLaVA-V0, setting the temperature to zero and restricting generation
to a single beam. Because LLaVA-Med was not pre-trained on multi-image inputs nor evaluated
in a few-shot setting, details on few-shot example construction are not applicable.

\paragraph{OpenFlamingo (9B).}
For OpenFlamingo~\citep{awadalla2023openflamingo}, no official documentation is available
for the system prompt, prompt format, or sampling configuration in the context of
closed-ended medical VQA evaluation; all three entries are therefore marked as not
applicable in Table~\ref{tab:prompt_summary}. We instead follow \citet{Jeong2024MedicalAO}
exactly, adopting their prompt format and generation settings without modification. Consistent
with their official configuration
file,\footnote{\url{https://github.com/mlfoundations/open_flamingo/blob/main/open_flamingo/src/flamingo.py\#L124}}
we set the temperature to zero and restrict generation to a single beam, ensuring fully deterministic and reproducible outputs.

\paragraph{Med-Flamingo (9B).}
For Med-Flamingo~\citep{moor2023med}, we adopt by default the system prompt and prompt
format provided in the demo code of the official GitHub
repository;\footnote{\url{https://github.com/snap-stanford/med-flamingo/blob/7bcbb6c3932c814a6f7ae4b838ae4fada39b42a4/scripts/demo.py\#L54}}
accordingly, the system prompt entry in Table~\ref{tab:prompt_summary} is marked as
partially available. However, since the demo does not specify how answer choices should be
formatted in a closed-ended QA context, the prompt format entry is likewise marked as
partially available, and we additionally search over prompt formats when performing
model-specific prompt selection. As with OpenFlamingo above, we apply the identical
generation configuration from \citet{Jeong2024MedicalAO} and their official configuration
file,\footnote{\url{https://github.com/taekb/eval-medical-dapt/blob/main/configs/vlm/eval/model/med-flamingo-9b.yaml}}
setting the temperature to zero and restricting generation to a single beam; the sampling entry is therefore marked as not applicable.

\paragraph{BioMed-Llama-3.2-11B-Vision (11B).}
BioMed-Llama-3.2-11B-Vision is a medical vision--language model from
\citet{cheng-etal-2025-domain}, fine-tuned on the general-domain
Llama-3.2-11B-Vision-Instruct backbone~\citep{meta_llama3_2_2024}. No system prompt is documented in either the official paper
or the corresponding HuggingFace model repositories; we therefore do not apply any system
prompt, and that entry is marked as not applicable in Table~\ref{tab:prompt_summary}. The prompt format is adopted directly from the
multiple-choice VQA protocol documented in the official paper
appendix~\cite[page 18]{cheng-etal-2025-domain} and
is therefore marked as fully available. The sampling configuration is partially available:
we set the temperature to zero following the official inference
script,\footnote{\url{https://github.com/bigai-ai/QA-Synthesizer/blob/main/vllm_inference/inference.py\#L33}}
while other generation hyperparameters are left at their defaults; the sampling entry is
accordingly marked as partially available.

\paragraph{BioMed-Qwen2-VL-2B (2B).}
BioMed-Qwen2-VL-2B is another model released by \citet{cheng-etal-2025-domain}, obtained
by fine-tuning the Qwen2-VL-2B-Instruct backbone~\cite{wang2024qwen2} on the same
domain-adapted medical data. As with BioMed-Llama-3.2-11B-Vision above, no system prompt
is used, the prompt format is taken directly from the official paper appendix, and the
temperature is set to zero following the same inference script; all three entries in
Table~\ref{tab:prompt_summary} are therefore assigned the same marks as for
BioMed-Llama-3.2-11B-Vision.

\paragraph{BioMed-InternVL3-1B (1B).}
BioMed-InternVL3-1B is another model from \citet{cheng-etal-2025-domain}, fine-tuned on
the InternVL3-1B backbone~\citep{zhu2025internvl3}. The prompting and generation setup is
identical to that described for BioMed-Llama-3.2-11B-Vision and BioMed-Qwen2-VL-2B above:
no system prompt is applied, the prompt format follows the multiple-choice VQA protocol
from the official paper appendix, and the temperature is fixed at zero per the same
inference script, resulting in the same set of marks in Table~\ref{tab:prompt_summary}.

\paragraph{InternVL3-1B (1B).}
For InternVL3-1B~\citep{zhu2025internvl3}, no system prompt is documented in either the
official paper or the HuggingFace model repository; we therefore do not apply any system
prompt, and that entry is marked as not applicable in Table~\ref{tab:prompt_summary}. For
the prompt format, the official paper reports that evaluations are conducted using
VLMEvalKit~\cite[page 7]{zhu2025internvl3} for consistency, we
adopt the VLMEvalKit multiple-choice question prompt format
directly,\footnote{\url{https://github.com/open-compass/VLMEvalKit/blob/43af13e052de6805a8b08cd04aed5e0d74f82ff5/vlmeval/dataset/image\_mcq.py\#L164}}
and the prompt format entry is accordingly marked as fully available. The sampling
configuration is partially available: we set the temperature to zero following the official
InternVL3 evaluation
script,\footnote{\url{https://github.com/OpenGVLab/InternVL/blob/main/internvl\_chat/eval/vqa/evaluate\_vqa.py\#L498}}
while other generation hyperparameters are left at their defaults.

\paragraph{Llama-3.2-11B-Vision (11B).}
For Llama-3.2-11B-Vision~\citep{meta_llama3_2_2024,grattafiori2024llama3herdmodels}, no
system prompt is documented in either the official paper or the HuggingFace model
repository; we therefore do not apply any system prompt, and that entry is marked as not
applicable in Table~\ref{tab:prompt_summary}. The prompt format is marked as partially
available: the official vision prompt format
documentation\footnote{\url{https://github.com/meta-llama/llama-models/blob/main/models/llama3\_2/vision\_prompt\_format.md}}
specifies that the \texttt{<|image|>} tag must appear within the user message body
immediately after the role header, with the text query following the image tag, as images
only attend to subsequent text tokens; however, no dedicated closed-ended evaluation
protocol is provided, and the answer-choice formatting is therefore inferred from this
structural constraint rather than from an official benchmark setup. No sampling
configuration is documented; we accordingly leave all generation hyperparameters at their
defaults, and the sampling entry is marked as not applicable in
Table~\ref{tab:prompt_summary}.

\paragraph{Qwen3-VL (2B, 4B, 8B, and 30B).}
For the Qwen3-VL family~\citep{bai2025qwen3}, no system prompt is documented in either the
official paper or the HuggingFace model repositories; we therefore do not apply any system
prompt, and that entry is marked as not applicable in Table~\ref{tab:prompt_summary}. The
prompt format is marked as fully available: the official Qwen3-VL paper documents the
multiple-choice prompt format~\cite[page 35]{bai2025qwen3} used for general-domain VQA benchmarks such as
MMBench~\citep{liu2024mmbenchmultimodalmodelallaround} and
MMStar~\citep{chen2024rightwayevaluatinglarge},
and we adopt this format directly. The sampling configuration is likewise fully available:
following the official Qwen3-VL paper~\cite[page 20]{bai2025qwen3}
and the corresponding evaluation
script,\footnote{\url{https://github.com/QwenLM/Qwen3-VL/blob/main/evaluation/mmmu/run\_mmmu.py\#L393}}
we set the temperature to $0.7$, which controls the randomness of the output distribution by scaling the logits before sampling, and the nucleus sampling threshold to $0.8$, which restricts sampling to the smallest set of tokens whose cumulative probability mass meets or exceeds that value; all other generation hyperparameters are left at their defaults.

\paragraph{Qwen2.5-VL (3B, 7B, and 32B).} For the Qwen2.5-VL family~\citep{bai2025qwen25vl}, no system prompt is documented in either the official paper or the HuggingFace model repositories; as with Qwen3-VL above, we do not apply any system prompt. No official closed-ended evaluation protocol or dedicated multiple-choice prompt format is provided; accordingly, the prompt format entry is marked as not applicable in Table~\ref{tab:prompt_summary}. Since Qwen2.5-VL shares the same underlying chat format as Qwen3-VL, we apply the same multiple-choice prompt format documented in the Qwen3-VL paper for consistency across the Qwen model family. Similarly, as no official sampling configuration is provided, we apply the same generation settings used for Qwen3-VL; the sampling entry is accordingly marked as not applicable. This approach is consistent with recent work that evaluates Qwen2-VL and related Qwen families using a unified evaluation framework~\citep{GILAL2025100455}.

\paragraph{Qwen2-VL (2B).} For Qwen2-VL~\citep{wang2024qwen2}, no system prompt is documented in either the official paper or the HuggingFace model repository, and no dedicated closed-ended evaluation protocol is provided; both entries are therefore marked as not applicable in Table~\ref{tab:prompt_summary}. As with Qwen2.5-VL above, Qwen2-VL shares the same underlying chat format across the Qwen family, and we therefore apply the same multiple-choice prompt format and generation settings following the Qwen3-VL paper for consistency.

\paragraph{MedVLThinker-RL (3B, 7B, and 32B).}
For the MedVLThinker-RL family~\citep{huang2025medvlthinker}, both the system prompt and
the multiple-choice prompt format are fully documented and available. The system prompt is
taken directly from the official GitHub evaluation
script,\footnote{\url{https://github.com/UCSC-VLAA/MedVLThinker/blob/main/eval/run\_offline\_inference.py\#L327}}
and the prompt format follows the template specified in both the official
paper\footnote{\url{https://arxiv.org/pdf/2508.02669\#page=12}} and the same evaluation
script;\footnote{\url{https://github.com/UCSC-VLAA/MedVLThinker/blob/main/eval/run\_offline\_inference.py\#L330}}
both entries are accordingly marked as fully available in Table~\ref{tab:prompt_summary}.
For sampling, although the repository defines default generation hyperparameter values, the
official demo instructs users to set the temperature to $0.6$, which controls the randomness
of the output distribution by scaling the logits before sampling, and the nucleus sampling
threshold to $0.95$, which restricts sampling to the smallest set of tokens whose cumulative
probability mass meets or exceeds that value;\footnote{\url{https://github.com/UCSC-VLAA/MedVLThinker/tree/main?tab=readme-ov-file\#demo}}
for consistency with the intended usage, we adopt these values accordingly, and the
sampling entry is marked as fully available.

\paragraph{MedMO-4B-Next and MedMO-8B-Next.}
For both MedMO variants (4B-Next and 8B-Next)~\citep{deria2026medmo}, we adopt the prompt
template specified directly in the official GitHub evaluation
script.\footnote{\url{https://github.com/genmilab/MedMO/blob/main/examples/scripts/medevalkit\_loader.py\#L35}}
No system prompt is documented in either the official paper or the corresponding repository;
we therefore do not apply any system prompt, and that entry is marked as not applicable in
Table~\ref{tab:prompt_summary}. The prompt format is taken directly from the official
evaluation script and is accordingly marked as fully available. No official sampling
configuration or generation hyperparameters are documented for closed-ended evaluation;
we therefore leave all generation hyperparameters at their defaults, and the sampling entry
is marked as not applicable in Table~\ref{tab:prompt_summary}.

\paragraph{MediX-R1 (2B and 30B).}
For both MediX-R1 variants (2B and 30B)~\citep{mullappilly2026medix}, we adopt the prompt
template, system prompt, and generation configuration specified directly in the official
GitHub evaluation
script.\footnote{\url{https://github.com/mbzuai-oryx/MediX-R1/blob/main/eval/utils.py\#L83}}
The system prompt is fully documented in both the official paper
appendix\footnote{\url{https://arxiv.org/pdf/2602.23363\#page=17}} and the corresponding
training prompt
template;\footnote{\url{https://github.com/mbzuai-oryx/MediX-R1/blob/main/training/examples/format\_prompt/medical\_format.jinja}}
both entries are accordingly marked as fully available in
Table~\ref{tab:prompt_summary}. MediX-R1 is the second reasoning-oriented model family
in our evaluation, alongside MedVLThinker-RL; it uses a dedicated system prompt that
instructs the model to first identify the imaging modality, then reason through the
problem within structured tags, before producing a final answer. Following the official
evaluation
script,\footnote{\url{https://github.com/mbzuai-oryx/MediX-R1/blob/main/eval/utils.py\#L144}}
the system prompt and the question body are combined within a single user turn in
chat-style formatting. For sampling, we adopt the generation hyperparameters
specified in the official evaluation
script:\footnote{\url{https://github.com/mbzuai-oryx/MediX-R1/blob/main/eval/utils.py\#L178}}
specifically, the temperature is set to zero, meaning the model always selects the most
probable next token at each step rather than sampling stochastically, which ensures fully
deterministic and reproducible outputs; the nucleus sampling threshold is set to one,
which places no restriction on the cumulative token probability mass and is therefore
effectively inactive under greedy decoding. All other generation hyperparameters are left
at their defaults, and the sampling entry is marked as fully available in
Table~\ref{tab:prompt_summary}.

\subsection{Default Prompt Formats}
\label{subsec:formats}

In this section, we describe the default prompt formats used for each model family, using a
representative multiple-choice question as a running example. We use the following color
conventions throughout: the question text is shown in \textcolor{violet}{\textbf{violet}},
the answer options in \textcolor{orange!80!black}{\textbf{orange}}, and the image token in
\textcolor{red!70!black}{\textbf{red}}. Importantly, the placement of the image token
relative to the text block is not uniform across model families. For the Gemma~3 family
(including MedGemma), the text block is placed \emph{before} the image token within the user
turn, following the ordering convention specified in the official model cards. We follow each model's recommended convention strictly, as incorrect
ordering can affect how the model attends to the visual input.

\paragraph{Gemma 3 (4B and 27B).}
Both Gemma~3 variants are prompted using the standard two-role chat template prescribed by the model architecture, which supports only a \texttt{user} turn and a \texttt{model} turn. Since the Gemma~3 instruction-tuned models do not natively support a system role, any system-level text would be silently folded into the first user turn without a dedicated marker, we omit a system prompt entirely and keep the input as minimal as possible \citep{gemma2025gemma3}. Each prompt consists of a single user turn containing the question text followed by the answer options, formatted as a multiple-choice question consistent with the structure demonstrated in the official MedGemma fine-tuning notebook;\footnote{\url{https://colab.research.google.com/github/google-health/medgemma/blob/main/notebooks/fine_tune_with_hugging_face.ipynb}} we adopt this same format for prompting. A single image is appended after the text within the same user turn, following the ordering convention specified in the model card; note that this model family does not support multi-image inputs \citep{gemma2025gemma3}. The resulting prompt format is illustrated in Box~\ref{box:gemma_format}.

\refstepcounter{promptbox}
\label{box:gemma_format}
\begin{tcolorbox}[
  colback=white,
  colframe=black,
  coltitle=black,
  colbacktitle=gray!10,
  title={\large Box~\thepromptbox: Gemma 3 (4B \& 27B)},
  fonttitle=\large,
  boxrule=0.5pt,
  arc=2pt,
  left=4pt, right=4pt, top=4pt, bottom=4pt
]
\normalsize
\texttt{<start\_of\_turn>user}\\[2pt]
\textcolor{violet}{\texttt{\{question\}}}\\
\textcolor{orange!80!black}{\texttt{A: \{option\_A\}}}\\
\textcolor{orange!80!black}{\texttt{B: \{option\_B\}}}\\
\textcolor{orange!80!black}{\texttt{C: \{option\_C\}}}\\
\textcolor{orange!80!black}{\texttt{D: \{option\_D\}}}\\
\textcolor{red!70!black}{\texttt{<image>}}\\[2pt]
\texttt{<end\_of\_turn>}\\
\texttt{<start\_of\_turn>model}
\end{tcolorbox}
\paragraph{MedGemma (4B and 27B).}
MedGemma shares the same underlying Gemma~3 architecture and tokenizer, and therefore uses an identical chat template structure \citep{sellergren2025medgemma}. As with Gemma~3, the instruction-tuned models support only a user turn and a model turn, and do not natively support a system role; we therefore omit a system prompt entirely and keep the input minimal. Each prompt consists of a single user turn containing the question text followed by the answer options, formatted as a multiple-choice question consistent with the structure demonstrated in the official MedGemma fine-tuning notebook. A single image is appended after the text within the same user turn, following the ordering convention specified in the model card. Although MedGemma has been domain-adapted on medical imaging data \citep{sellergren2025medgemma}, its prompt format is marked as partially available because no dedicated closed-ended evaluation protocol has been publicly released; we therefore rely on the format demonstrated in the fine-tuning notebook. The resulting prompt format is illustrated in Box~\ref{box:medgemma_format}.

\refstepcounter{promptbox}
\label{box:medgemma_format}
\begin{tcolorbox}[
  colback=white,
  colframe=black,
  coltitle=black,
  colbacktitle=gray!10,
  title={\large Box~\thepromptbox: MedGemma (4B \& 27B)},
  fonttitle=\large,
  boxrule=0.5pt,
  arc=2pt,
  left=4pt, right=4pt, top=4pt, bottom=4pt
]
\normalsize
\texttt{<start\_of\_turn>user}\\[2pt]
\textcolor{violet}{\texttt{\{question\}}}\\
\textcolor{orange!80!black}{\texttt{A: \{option\_A\}}}\\
\textcolor{orange!80!black}{\texttt{B: \{option\_B\}}}\\
\textcolor{orange!80!black}{\texttt{C: \{option\_C\}}}\\
\textcolor{orange!80!black}{\texttt{D: \{option\_D\}}}\\
\textcolor{red!70!black}{\texttt{<image>}}\\[2pt]
\texttt{<end\_of\_turn>}\\
\texttt{<start\_of\_turn>model}
\end{tcolorbox}

\paragraph{LLaVA-V0 (7B).}
LLaVA-V0 uses the \texttt{simple\_conv} template, which renders a flat dialogue string
with no dedicated special token for the system role; the general-domain system preamble is
prepended directly to the conversation. The conversation body follows the VICUNA-V0 format,
with each turn prefixed by \texttt{\#\#\# Human:} or \texttt{\#\#\# Assistant:}, and
generation is triggered by a trailing \texttt{\#\#\# Assistant:} with no closing text.
Answer options are appended inline within the human turn. The image is not represented by a
placeholder \texttt{<image>} token at inference time; instead, the runtime loader injects
the model's internal image-patch token block directly into the prompt, expanding it as
\texttt{<im\_start><im\_patch>...<im\_patch><im\_end>} using the checkpoint's
\texttt{mm\_use\_im\_start\_end} configuration\footnote{\url{https://github.com/taekb/eval-medical-dapt/blob/main/src/dataset.py\#L1163}} and an image token length of 256\footnote{\url{https://github.com/taekb/eval-medical-dapt/blob/main/src/vlm/llava_models/infer_llava.py\#L370}}. The
resulting prompt format is illustrated in Box~\ref{box:llavav0_format}.

\refstepcounter{promptbox}
\label{box:llavav0_format}
\begin{tcolorbox}[
  colback=white,
  colframe=black,
  coltitle=black,
  colbacktitle=gray!10,
  title={\large Box~\thepromptbox: LLaVA-V0 (7B)},
  fonttitle=\large,
  boxrule=0.5pt,
  arc=2pt,
  left=4pt, right=4pt, top=4pt, bottom=4pt
]
\normalsize
\texttt{A chat between a curious human and an artificial intelligence assistant. The assistant gives helpful, detailed, and polite answers to the human's questions.}\\[4pt]
\texttt{\#\#\# Human:}
\textcolor{violet}{\texttt{\{question\}}}
\texttt{Please choose from the following options:}
\textcolor{orange!80!black}{\texttt{[\{option\_1\}, \{option\_2\}, \{option\_3\}, \{option\_4\}]}.} \textcolor{red!70!black}{\texttt{<image>}}\\[2pt]
\texttt{\#\#\# Assistant:}
\end{tcolorbox}

\paragraph{LLaVA-Med (7B).}
LLaVA-Med shares the same flat dialogue rendering and VICUNA-V0 conversational structure as
LLaVA-V0, differing only in its system prompt. The \texttt{simple\_conv\_med} template
replaces the general-domain preamble with a medically-adapted one specific to LLaVA-Med.
Image token injection, answer-option placement, and the trailing \texttt{\#\#\# Assistant:}
generation trigger follow the same conventions described for LLaVA-V0 above. The resulting
prompt format is illustrated in Box~\ref{box:llavamed_format}.

\refstepcounter{promptbox}
\label{box:llavamed_format}
\begin{tcolorbox}[
  colback=white,
  colframe=black,
  coltitle=black,
  colbacktitle=gray!10,
  title={\large Box~\thepromptbox: LLaVA-Med (7B)},
  fonttitle=\large,
  boxrule=0.5pt,
  arc=2pt,
  left=4pt, right=4pt, top=4pt, bottom=4pt
]
\normalsize
\texttt{You are LLaVA-Med, a large language and vision assistant trained by a group of researchers at Microsoft, based on the general domain LLaVA architecture. You are able to understand the visual content that the user provides, and assist the user with a variety of medical and clinical tasks using natural language. Follow the instructions carefully and explain your answers in detail.}\\[4pt]
\texttt{\#\#\# Human:}
\textcolor{violet}{\texttt{\{question\}}}
\texttt{Please choose from the following options:}
\textcolor{orange!80!black}{\texttt{ [\{option\_1\}, \{option\_2\}, \{option\_3\}, \{option\_4\}]}.} \textcolor{red!70!black}{\texttt{<image>}}\\[2pt]
\texttt{\#\#\# Assistant:}
\end{tcolorbox}

\paragraph{OpenFlamingo (9B).}
OpenFlamingo uses no system prompt. Each prompt opens with a brief task instruction
prepended directly to the conversation body, followed by the image token and the question
text within the same turn. Answer options are formatted with lettered parenthetical labels
on separate lines, and generation is triggered by a trailing \texttt{Answer:} stub.
Notably, the image token is placed \emph{before} the question text within the user turn,
which is the opposite ordering convention from the Gemma~3 family described above.
The resulting prompt format is illustrated in Box~\ref{box:openflamingo_format}.

\refstepcounter{promptbox}
\label{box:openflamingo_format}
\begin{tcolorbox}[
  colback=white,
  colframe=black,
  coltitle=black,
  colbacktitle=gray!10,
  title={\large Box~\thepromptbox: OpenFlamingo (9B)},
  fonttitle=\large,
  boxrule=0.5pt,
  arc=2pt,
  left=4pt, right=4pt, top=4pt, bottom=4pt
]
\normalsize
\texttt{The following is a multiple-choice visual question requiring medical knowledge. Answer the question by choosing one of the provided answer options.}\\[4pt]
\textcolor{red!70!black}{\texttt{<image>}}
\textcolor{violet}{\texttt{\{question\}}}\\
\textcolor{orange!80!black}{\texttt{(A) \{option\_A\}}}\\
\textcolor{orange!80!black}{\texttt{(B) \{option\_B\}}}\\
\textcolor{orange!80!black}{\texttt{(C) \{option\_C\}}}\\
\textcolor{orange!80!black}{\texttt{(D) \{option\_D\}}}\\
\texttt{Answer:}
\end{tcolorbox}

\paragraph{Med-Flamingo (9B).}
Med-Flamingo shares the same overall prompt structure as OpenFlamingo above, differing
primarily in its system preamble. Rather than a generic task instruction, the preamble is
replaced by the medically-adapted prompt from the official demo, which instructs the model
to act as a helpful medical assistant and to follow provided examples before answering.
The image token, question text, lettered answer options, and trailing \texttt{Answer:} stub
follow the same ordering and formatting conventions described for OpenFlamingo. The
resulting prompt format is illustrated in Box~\ref{box:medflamingo_format}.

\refstepcounter{promptbox}
\label{box:medflamingo_format}
\begin{tcolorbox}[
  colback=white,
  colframe=black,
  coltitle=black,
  colbacktitle=gray!10,
  title={\large Box~\thepromptbox: Med-Flamingo (9B)},
  fonttitle=\large,
  boxrule=0.5pt,
  arc=2pt,
  left=4pt, right=4pt, top=4pt, bottom=4pt
]
\normalsize
\texttt{You are a helpful medical assistant. You are being provided with images, a question about the image and an answer. Follow the examples and answer the last question.}\\[4pt]
\textcolor{red!70!black}{\texttt{<image>}}
\textcolor{violet}{\texttt{\{question\}}}\\
\textcolor{orange!80!black}{\texttt{(A) \{option\_A\}}}\\
\textcolor{orange!80!black}{\texttt{(B) \{option\_B\}}}\\
\textcolor{orange!80!black}{\texttt{(C) \{option\_C\}}}\\
\textcolor{orange!80!black}{\texttt{(D) \{option\_D\}}}\\
\texttt{Answer:}
\end{tcolorbox}

\paragraph{BioMed-Llama-3.2-11B-Vision (11B).}
BioMed-Llama-3.2-11B-Vision uses no system prompt. Each prompt begins with the image
token on its own line, followed by the question text prefixed by \texttt{Question:}, and
then the answer options formatted as a Python-style list on a single line prefixed by
\texttt{The choices are:}. The resulting
prompt format is illustrated in Box~\ref{box:biomedllama_format}.

\refstepcounter{promptbox}
\label{box:biomedllama_format}
\begin{tcolorbox}[
  colback=white,
  colframe=black,
  coltitle=black,
  colbacktitle=gray!10,
  title={\large Box~\thepromptbox: BioMed-Llama-3.2-11B-Vision (11B)},
  fonttitle=\large,
  boxrule=0.5pt,
  arc=2pt,
  left=4pt, right=4pt, top=4pt, bottom=4pt
]
\normalsize
\textcolor{red!70!black}{\texttt{<image>}}\\
\texttt{Question:} \textcolor{violet}{\texttt{\{question\}}}\\
\textcolor{orange!80!black}{\texttt{The choices are: [\{option\_A\}, \{option\_B\}, \{option\_C\}, \{option\_D\}]}}
\end{tcolorbox}

\paragraph{BioMed-Qwen2-VL-2B (2B).}
BioMed-Qwen2-VL-2B shares an identical prompt structure to BioMed-Llama-3.2-11B-Vision
above, with the image token placed before the question text, the same \texttt{Question:}
and \texttt{The choices are:} prefixes, and options rendered as a Python-style list. The
only difference lies in the underlying backbone architecture; the prompt format itself is
unchanged across the three BioMed models released by \citet{cheng-etal-2025-domain}. The
resulting prompt format is illustrated in Box~\ref{box:biomedqwen_format}.

\refstepcounter{promptbox}
\label{box:biomedqwen_format}
\begin{tcolorbox}[
  colback=white,
  colframe=black,
  coltitle=black,
  colbacktitle=gray!10,
  title={\large Box~\thepromptbox: BioMed-Qwen2-VL-2B (2B)},
  fonttitle=\large,
  boxrule=0.5pt,
  arc=2pt,
  left=4pt, right=4pt, top=4pt, bottom=4pt
]
\normalsize
\textcolor{red!70!black}{\texttt{<image>}}\\
\texttt{Question:} \textcolor{violet}{\texttt{\{question\}}}\\
\textcolor{orange!80!black}{\texttt{The choices are: [\{option\_A\}, \{option\_B\}, \{option\_C\}, \{option\_D\}]}}
\end{tcolorbox}

\paragraph{BioMed-InternVL3-1B (1B).}
BioMed-InternVL3-1B follows the same prompt format as BioMed-Llama-3.2-11B-Vision and
BioMed-Qwen2-VL-2B, with no system prompt, the image token prepended before the question
text, and answer options presented as a Python-style list under the \texttt{The choices
are:} prefix. The resulting prompt format is illustrated in
Box~\ref{box:biomedinternvl_format}.

\refstepcounter{promptbox}
\label{box:biomedinternvl_format}
\begin{tcolorbox}[
  colback=white,
  colframe=black,
  coltitle=black,
  colbacktitle=gray!10,
  title={\large Box~\thepromptbox: BioMed-InternVL3-1B (1B)},
  fonttitle=\large,
  boxrule=0.5pt,
  arc=2pt,
  left=4pt, right=4pt, top=4pt, bottom=4pt
]
\normalsize
\textcolor{red!70!black}{\texttt{<image>}}\\
\texttt{Question:} \textcolor{violet}{\texttt{\{question\}}}\\
\textcolor{orange!80!black}{\texttt{The choices are: [\{option\_A\}, \{option\_B\}, \{option\_C\}, \{option\_D\}]}}
\end{tcolorbox}

\paragraph{InternVL3-1B (1B).}
InternVL3-1B uses no system prompt. Each prompt begins with the image token on its own
line, followed by the question text prefixed by \texttt{Question:}, and then the answer
options listed on separate lines under an \texttt{Options:} header, each prefixed by its
corresponding letter label. A closing instruction line, \texttt{Please select the correct
answer from the options above.}, is appended after the options. The image token is placed
\emph{before} the question text, consistent with the ordering convention used by the BioMed-InternVL3-1B described above. The resulting prompt format is illustrated in
Box~\ref{box:internvl3_format}.

\refstepcounter{promptbox}
\label{box:internvl3_format}
\begin{tcolorbox}[
  colback=white,
  colframe=black,
  coltitle=black,
  colbacktitle=gray!10,
  title={\large Box~\thepromptbox: InternVL3-1B (1B)},
  fonttitle=\large,
  boxrule=0.5pt,
  arc=2pt,
  left=4pt, right=4pt, top=4pt, bottom=4pt
]
\normalsize
\textcolor{red!70!black}{\texttt{<image>}}\\
\texttt{Question:} \textcolor{violet}{\texttt{\{question\}}}\\
\texttt{Options:}\\
\textcolor{orange!80!black}{\texttt{A. \{option\_A\}}}\\
\textcolor{orange!80!black}{\texttt{B. \{option\_B\}}}\\
\textcolor{orange!80!black}{\texttt{C. \{option\_C\}}}\\
\textcolor{orange!80!black}{\texttt{D. \{option\_D\}}}\\
\texttt{Please select the correct answer from the options above.}
\end{tcolorbox}

\paragraph{Llama-3.2-11B-Vision (11B).}
Llama-3.2-11B-Vision uses no system prompt and follows the chat template prescribed by the
official vision prompt format documentation. Each prompt is wrapped in the model's native
special tokens, opening with \texttt{<|begin\_of\_text|>} and a
\texttt{<|start\_header\_id|>user<|end\_header\_id|>} role header. The \texttt{<|image|>}
tag is placed immediately after the role header and before the question text, as required
by the architecture: images only attend to subsequent text tokens, so the question must
follow rather than precede the image tag. As no official closed-ended evaluation protocol
is provided, the formatting of answer options and the closing instruction
\texttt{Answer with the letter only.} were designed by us; answer options are listed on
separate lines with parenthetical letter labels, and the closing instruction is appended
after the final option. Generation is triggered by a trailing
\texttt{assistant<|end\_header\_id|>} header. The resulting prompt
format is illustrated in Box~\ref{box:llama32vision_format}.

\refstepcounter{promptbox}
\label{box:llama32vision_format}
\begin{tcolorbox}[
  colback=white,
  colframe=black,
  coltitle=black,
  colbacktitle=gray!10,
  title={\large Box~\thepromptbox: Llama-3.2-11B-Vision (11B)},
  fonttitle=\large,
  boxrule=0.5pt,
  arc=2pt,
  left=4pt, right=4pt, top=4pt, bottom=4pt
]
\normalsize
\texttt{<|begin\_of\_text|><|start\_header\_id|> \\ user<|end\_header\_id|>}\\
\textcolor{red!70!black}{\texttt{<|image|>}}\textcolor{violet}{\texttt{\{question\}}}\\
\texttt{Options:}\\
\textcolor{orange!80!black}{\texttt{A) \{option\_A\}}}\\
\textcolor{orange!80!black}{\texttt{B) \{option\_B\}}}\\
\textcolor{orange!80!black}{\texttt{C) \{option\_C\}}}\\
\textcolor{orange!80!black}{\texttt{D) \{option\_D\}}}\\
\texttt{Answer with the letter only.<|eot\_id|>}\\
\texttt{<|start\_header\_id|>assistant \\<|end\_header\_id|>}
\end{tcolorbox}

\paragraph{Qwen3-VL (2B, 4B, 8B, and 30B).}
Qwen3-VL uses no system prompt. Each prompt begins with the image token on its own line,
followed by the question text prefixed by \texttt{Question:}, and the answer options listed
on separate lines under an \texttt{Options:} header, each prefixed by its corresponding
letter label followed by a period. A closing instruction line, \texttt{Please select the
correct answer from the options above.}, is appended after the final option. This format
follows the multiple-choice convention documented in the official Qwen3-VL paper for
general-domain VQA benchmarks, and is identical in structure to the VLMEvalKit MCQ format
adopted for InternVL3-1B described above. The image token is placed before the question text, following the official Qwen3-VL evaluation script.\footnote{\url{https://github.com/QwenLM/Qwen3-VL/blob/main/evaluation/mmmu/run_mmmu.py\#L27}} The resulting prompt format is illustrated in
Box~\ref{box:qwen3vl_format}.

\refstepcounter{promptbox}
\label{box:qwen3vl_format}
\begin{tcolorbox}[
  colback=white,
  colframe=black,
  coltitle=black,
  colbacktitle=gray!10,
  title={\large Box~\thepromptbox: Qwen3-VL (2B, 4B, 8B, \& 30B)},
  fonttitle=\large,
  boxrule=0.5pt,
  arc=2pt,
  left=4pt, right=4pt, top=4pt, bottom=4pt
]
\normalsize
\textcolor{red!70!black}{\texttt{<image>}}\\
\texttt{Question:} \textcolor{violet}{\texttt{\{question\}}}\\
\texttt{Options:}\\
\textcolor{orange!80!black}{\texttt{A. \{option\_A\}}}\\
\textcolor{orange!80!black}{\texttt{B. \{option\_B\}}}\\
\textcolor{orange!80!black}{\texttt{C. \{option\_C\}}}\\
\textcolor{orange!80!black}{\texttt{D. \{option\_D\}}}\\
\texttt{Please select the correct answer from the options above.}
\end{tcolorbox}

\paragraph{Qwen2.5-VL (3B, 7B, and 32B).}
Qwen2.5-VL uses no system prompt and shares the same prompt structure as Qwen3-VL above,
owing to the common underlying chat format across the Qwen family. The image token, the
\texttt{Question:} prefix, the lettered \texttt{Options:} block, and the closing
instruction line follow the identical conventions described for Qwen3-VL. The resulting
prompt format is illustrated in Box~\ref{box:qwen25vl_format}.

\refstepcounter{promptbox}
\label{box:qwen25vl_format}
\begin{tcolorbox}[
  colback=white,
  colframe=black,
  coltitle=black,
  colbacktitle=gray!10,
  title={\large Box~\thepromptbox: Qwen2.5-VL (3B, 7B, \& 32B)},
  fonttitle=\large,
  boxrule=0.5pt,
  arc=2pt,
  left=4pt, right=4pt, top=4pt, bottom=4pt
]
\normalsize
\textcolor{red!70!black}{\texttt{<image>}}\\
\texttt{Question:} \textcolor{violet}{\texttt{\{question\}}}\\
\texttt{Options:}\\
\textcolor{orange!80!black}{\texttt{A. \{option\_A\}}}\\
\textcolor{orange!80!black}{\texttt{B. \{option\_B\}}}\\
\textcolor{orange!80!black}{\texttt{C. \{option\_C\}}}\\
\textcolor{orange!80!black}{\texttt{D. \{option\_D\}}}\\
\texttt{Please select the correct answer from the options above.}
\end{tcolorbox}

\paragraph{Qwen2-VL (2B).}
Qwen2-VL likewise uses no system prompt and follows the same prompt structure as Qwen3-VL
and Qwen2.5-VL, for the same reasons of shared chat format across the Qwen family. All
formatting conventions, image token placement, \texttt{Question:} prefix, lettered
\texttt{Options:} block, and closing instruction, are identical to those described above.
The resulting prompt format is illustrated in Box~\ref{box:qwen2vl_format}.

\refstepcounter{promptbox}
\label{box:qwen2vl_format}
\begin{tcolorbox}[
  colback=white,
  colframe=black,
  coltitle=black,
  colbacktitle=gray!10,
  title={\large Box~\thepromptbox: Qwen2-VL (2B)},
  fonttitle=\large,
  boxrule=0.5pt,
  arc=2pt,
  left=4pt, right=4pt, top=4pt, bottom=4pt
]
\normalsize
\textcolor{red!70!black}{\texttt{<image>}}\\
\texttt{Question:} \textcolor{violet}{\texttt{\{question\}}}\\
\texttt{Options:}\\
\textcolor{orange!80!black}{\texttt{A. \{option\_A\}}}\\
\textcolor{orange!80!black}{\texttt{B. \{option\_B\}}}\\
\textcolor{orange!80!black}{\texttt{C. \{option\_C\}}}\\
\textcolor{orange!80!black}{\texttt{D. \{option\_D\}}}\\
\texttt{Please select the correct answer from the options above.}
\end{tcolorbox}

\paragraph{MedVLThinker-RL (3B, 7B, and 32B).}
MedVLThinker-RL is the first model family in our evaluation to use a reasoning-oriented
prompt structure. A dedicated system prompt instructs the model to reason through the
problem before answering, and the prompt body follows this convention by requiring the
model to place its reasoning within \texttt{<think>} \texttt{</think>} tags and its final
answer within \texttt{<answer>} \texttt{</answer>} tags. The question text is prefixed by
\texttt{Question:} and the answer options are listed on separate lines under an
\texttt{Options:} header, each prefixed by its corresponding letter label followed by a
period. MedVLThinker places the \texttt{<image>} token at the beginning of the
user turn, before the question text, to indicate where the visual input is injected. The resulting prompt format is illustrated in
Box~\ref{box:medvlthinker_format}.

\refstepcounter{promptbox}
\label{box:medvlthinker_format}
\begin{tcolorbox}[
  colback=white,
  colframe=black,
  coltitle=black,
  colbacktitle=gray!10,
  title={\large Box~\thepromptbox: MedVLThinker-RL (3B, 7B, \& 32B)},
  fonttitle=\large,
  boxrule=0.5pt,
  arc=2pt,
  left=4pt, right=4pt, top=4pt, bottom=4pt
]
\normalsize
\textbf{[System]:} \texttt{You will solve a problem/request. You should provide your thoughts within <think> </think> tags before providing the answer. Write your final answer within <answer> </answer> tags.}\\[4pt]
\textbf{[User]:}\\
\textcolor{red!70!black}{\texttt{<image>}}\\[2pt]
\texttt{Question:} \textcolor{violet}{\texttt{\{question\}}}\\\\[2pt]
\texttt{Options:}\\\\[2pt]
\textcolor{orange!80!black}{\texttt{A. \{option\_A\}}}\\\\[2pt]
\textcolor{orange!80!black}{\texttt{B. \{option\_B\}}}\\\\[2pt]
\textcolor{orange!80!black}{\texttt{C. \{option\_C\}}}\\\\[2pt]
\textcolor{orange!80!black}{\texttt{D. \{option\_D\}}}
\end{tcolorbox}

\paragraph{MedMO-4B-Next and MedMO-8B-Next.}
Both MedMO variants use no system prompt. Each prompt begins with the image token on its
own line, followed by the question text, and then the answer options listed on separate
lines under an \texttt{Options:} header, each prefixed by its corresponding letter label
enclosed in parentheses. No closing instruction line is appended after the final option.
The image token is placed \emph{before} the question text, consistent with the ordering
convention adopted in the official evaluation
script.\footnote{\url{https://github.com/genmilab/MedMO/blob/main/examples/scripts/medevalkit_sft_loader.py\#L198}}
The resulting prompt format is illustrated in Box~\ref{box:medmo_format}.

\refstepcounter{promptbox}
\label{box:medmo_format}
\begin{tcolorbox}[
  colback=white,
  colframe=black,
  coltitle=black,
  colbacktitle=gray!10,
  title={\large Box~\thepromptbox: MedMO-4B-Next \& MedMO-8B-Next},
  fonttitle=\large,
  boxrule=0.5pt,
  arc=2pt,
  left=4pt, right=4pt, top=4pt, bottom=4pt
]
\normalsize
\textcolor{red!70!black}{\texttt{<image>}}\\
\textcolor{violet}{\texttt{\{question\}}}\\
\texttt{Options:}\\
\textcolor{orange!80!black}{\texttt{(A) \{option\_A\}}}\\
\textcolor{orange!80!black}{\texttt{(B) \{option\_B\}}}\\
\textcolor{orange!80!black}{\texttt{(C) \{option\_C\}}}\\
\textcolor{orange!80!black}{\texttt{(D) \{option\_D\}}}
\end{tcolorbox}

\paragraph{MediX-R1 (2B and 30B).}
MediX-R1 uses a reasoning-oriented prompt structure and is the only model family in our
evaluation to require explicit image modality identification as part of its output format.
The system prompt instructs the model to act as a medical AI assistant with advanced
reasoning capabilities, and prescribes a strict three-step output structure: first, the
model must output a single imaging modality tag from a predefined set (e.g.,
\texttt{<X\_RAY>}, \texttt{<MRI\_SCAN>}, \texttt{<CT\_SCAN>}); second, it must place its
reasoning within \texttt{<thinking>}\texttt{</thinking>} tags; and third, it must provide
its final answer within \texttt{<answer>}\texttt{</answer>} tags, with no additional text
permitted outside these tags. Following the official evaluation
script,\footnote{\url{https://github.com/mbzuai-oryx/MediX-R1/blob/main/eval/utils.py\#L144}}
the system prompt is prepended directly within the user turn rather than occupying a
dedicated system role, so that the full input is rendered as a single combined user
message. Within the question body, the \texttt{<image>} token is placed after the
\texttt{Question:} label and before the question text, which is the opposite ordering
convention from most other model families in our evaluation that place the image token
before the question prefix. Answer options are listed on separate lines under a
\texttt{Choices:} header, each prefixed by its corresponding letter label followed by a
period. The resulting prompt format is illustrated in Box~\ref{box:medixr1_format}.

\refstepcounter{promptbox}
\label{box:medixr1_format}
\begin{tcolorbox}[
  colback=white,
  colframe=black,
  coltitle=black,
  colbacktitle=gray!10,
  title={\large Box~\thepromptbox: MediX-R1 (2B \& 30B)},
  fonttitle=\large,
  boxrule=0.5pt,
  arc=2pt,
  left=4pt, right=4pt, top=4pt, bottom=4pt
]
\normalsize
\texttt{You are a Medical AI Assistant with advanced reasoning
capabilities}\\[4pt]
\texttt{Your task:}\\
\texttt{1. First output the image modality tag from this set:}\\
\texttt{\quad <X\_RAY>, <MICROSCOPY>, <CLINICAL\_PHOTOGRAPHY>, <CT\_SCAN>, <GRAPHICS>,}\\
\texttt{\quad <ANGIOGRAPHY>, <PET\_SCAN>, <ULTRASOUND>, <MRI\_SCAN>, <FUNDUS\_PHOTOGRAPHY>,}\\
\texttt{\quad <OCT\_SCAN>, <ENDOSCOPY>, <MAMMOGRAPHY>, <FLUOROSCOPY>, <OTHER>, <SPECT>}\\
\texttt{\quad (Only output the tag, nothing else.)}\\
\texttt{2. Then output the thinking and medical reasoning process in <thinking>...</thinking> tags.}\\
\texttt{3. Finally, provide the correct answer inside <answer>...</answer> tags.}\\
\texttt{4. Do not include any extra information or text outside of these tags.}\\[4pt]
\texttt{Question:}\\
\textcolor{red!70!black}{\texttt{<image>}}\\
\textcolor{violet}{\texttt{\{question\}}}\\\\[4pt]
\texttt{Choices:}\\
\textcolor{orange!80!black}{\texttt{A. \{option\_A\}}}\\
\textcolor{orange!80!black}{\texttt{B. \{option\_B\}}}\\
\textcolor{orange!80!black}{\texttt{C. \{option\_C\}}}\\
\textcolor{orange!80!black}{\texttt{D. \{option\_D\}}}
\end{tcolorbox}

\section{Few-Shot Probing Results}
\label{sec:fewshot}

All probing results reported in the main paper correspond to the zero-shot setting, in which the prompt contains no in-context examples from the target benchmark.
Table~\ref{tab:fewshot} compares zero-shot and 3-shot probing accuracy across a representative set of models and benchmarks.
In the 3-shot setting, the prompt is augmented with three labeled examples from the training split of the \emph{same} benchmark, following standard in-context learning practice.

\begin{table}[ht]
\centering
\caption{Probing accuracy (\%) under 0-shot and 3-shot settings. Bold indicates the better result per model--benchmark pair. The shaded column highlights the average.}
\label{tab:fewshot}
\resizebox{\columnwidth}{!}{%
\begin{tabular}{llccc>{\columncolor{blue!10}}c}
\toprule
\textbf{Model} & \textbf{Shot} & \textbf{PATH} & \textbf{SLAKE} & \textbf{RAD} & \textbf{Avg.} \\
\midrule
\multirow{2}{*}{InternVL3-1B}
  & 3 & 58.20          & 66.99          & \textbf{68.82} & 64.67 \\
  & 0 & \textbf{62.80} & \textbf{75.66} & 65.02          & \textbf{67.83} \\
\midrule
\multirow{2}{*}{BioMed-Qwen2-VL-2B}
  & 3 & 37.20          & 41.93          & 12.17          & 30.43 \\
  & 0 & \textbf{56.20} & \textbf{63.62} & \textbf{65.02} & \textbf{61.61} \\
\midrule
\multirow{2}{*}{Qwen3-VL-2B}
  & 3 & \textbf{57.20} & 74.22          & \textbf{65.02} & 65.48 \\
  & 0 & 56.80          & \textbf{77.83} & 64.64          & \textbf{66.42} \\
\midrule
\multirow{2}{*}{Qwen2.5-VL-3B}
  & 3 & \textbf{60.00} & 62.89          & 68.06          & 63.65 \\
  & 0 & 57.60          & \textbf{74.22} & \textbf{75.29} & \textbf{69.03} \\
\midrule
\multirow{2}{*}{LLaVA-V0-7B}
  & 3 & 54.20          & 42.41          & 44.87          & 47.16 \\
  & 0 & \textbf{56.00} & \textbf{50.36} & \textbf{49.43} & \textbf{51.93} \\
\midrule
\multirow{2}{*}{Qwen3-VL-8B}
  & 3 & \textbf{59.60} & 74.70          & \textbf{73.38} & \textbf{69.23} \\
  & 0 & 58.60          & \textbf{76.63} & 69.58          & 68.27 \\
\midrule
\multirow{2}{*}{Open-Flamingo-9B}
  & 3 & 49.80          & 43.86          & 47.53          & 47.06 \\
  & 0 & \textbf{53.00} & \textbf{48.92} & \textbf{48.67} & \textbf{50.20} \\
\midrule
\multirow{2}{*}{Qwen3-VL-30B}
  & 3 & \textbf{67.20} & 79.76          & 76.81          & \textbf{74.59} \\
  & 0 & 61.20          & \textbf{80.96} & \textbf{77.95} & 73.37 \\
\midrule
\multirow{2}{*}{Qwen2.5-VL-32B}
  & 3 & \textbf{66.60} & \textbf{80.72} & \textbf{87.45} & \textbf{78.26} \\
  & 0 & 57.20          & 77.59          & 82.51          & 72.43 \\
\bottomrule
\end{tabular}%
}
\end{table}

The zero-shot setting is competitive with or superior to 3-shot for most models.
Zero-shot probing outperforms 3-shot on average for 7 of 9 models.
The primary exceptions are Qwen3-VL-30B and Qwen2.5-VL-32B, where 3-shot prompting provides additional signal that improves VQA-RAD and PATH-VQA performance at the largest scale.
For models such as BioMed-Qwen2-VL-2B, 3-shot probing substantially \emph{degrades} performance (30.43\% vs.\ 61.61\% average), suggesting that in-context examples from different domains can disrupt the hidden-state structure on which the probe relies.
These results validate our choice of zero-shot as the primary evaluation setting, as it is both simpler and generally more effective for linear probing.

\section{Option-Likelihood Prompting and Statistical Significance}
\label{sec:option_likelihood_and_significance}

A core methodological concern is whether the prompting baseline underperforms because of free-text generation artifacts, refusals, verbose paraphrases, or formatting failures, rather than genuine knowledge limits.
To address this, we use option-likelihood scoring. We score each candidate answer by its conditional log-probability under the frozen VLM and pick whichever option scores highest.
This protocol sidesteps generation artifacts entirely and provides an upper bound on what prompting can extract without decoding.
Table~\ref{tab:option_likelihood} compares free-text prompting, option-likelihood scoring, and MedProb for three representative models.

Option-likelihood improves over free-text prompting by an average of 4.6 points across nine model--dataset pairs (59.8\%$\,\to\,$64.4\%), confirming that generation does suppress some correct signal.
However, probing outperforms option-likelihood by a further 14.2 points (64.4\%$\,\to\,$78.6\%), a gap far too large to attribute to generation artifacts.
The probing advantage therefore reflects richer access to the model's internal representations, not merely an elicitation-format effect.

\begin{table}[h]
\centering
\scriptsize
\setlength{\tabcolsep}{3pt}
\renewcommand{\arraystretch}{1.1}

\resizebox{\columnwidth}{!}{%
\begin{tabular}{llcccc}
\toprule
\textbf{Model} & \textbf{Method} & \textbf{PVQ} & \textbf{SLK} & \textbf{RAD} & \textbf{Avg.} \\
\midrule
\multirow{3}{*}{LLaVA-V0-7B}
  & Free-text    & 49.6 & 45.5 & 42.6 & 45.9 \\
  & Opt.-likeli. & 45.8 & 53.7 & 56.2 & 51.9 \\
  & Probing      & 84.6 & 68.0 & 66.9 & \textbf{73.2} \\
\midrule
\multirow{3}{*}{Llama-3.2-11B}
  & Free-text    & 59.8 & 69.9 & 66.5 & 65.4 \\
  & Opt.-likeli. & 62.2 & 72.8 & 72.1 & 69.0 \\
  & Probing      & 86.2 & 82.4 & 75.4 & \textbf{81.3} \\
\midrule
\multirow{3}{*}{Qwen3-VL-8B}
  & Free-text    & 59.0 & 76.6 & 68.8 & 68.1 \\
  & Opt.-likeli. & 67.2 & 77.3 & 72.4 & 72.3 \\
  & Probing      & 86.2 & 84.1 & 73.9 & \textbf{81.4} \\
\bottomrule
\end{tabular}%
}

\caption{Accuracy (\%) for free-text prompting, option-likelihood scoring, and MedProb on three representative VLMs. PVQ\,=\,PATH-VQA; SLK\,=\,SLAKE; RAD\,=\,VQA-RAD. Llama\,=\,Llama-3.2-11B-Vision.}
\label{tab:option_likelihood}
\end{table}

To confirm that the probing advantage is statistically reliable, we ran paired significance tests on the same model--dataset pairs, comparing probing against free-text prompting on exactly aligned predictions.
Table~\ref{tab:paired_significance} reports the McNemar exact $p$-value and a 95\% bootstrap confidence interval for the probing-minus-prompting accuracy delta.
Eight of nine comparisons are significant at $p < 0.05$.
The sole exception is Qwen3-VL-8B on VQA-RAD ($\Delta = +5.1$, CI $[-0.4, 11.0]$, $p = 0.098$), where the smaller test set (263 examples) limits statistical power.
On PATH-VQA, all three models clear $p < 10^{-21}$, consistent with the observation that the probing advantage is most pronounced on the pathology benchmark.

\begin{table}[h]
\centering
\small
\setlength{\tabcolsep}{4pt}
\renewcommand{\arraystretch}{1.15}

\resizebox{\columnwidth}{!}{%
\begin{tabular}{llccc}
\toprule
\textbf{Model} & \textbf{Dataset} & $\boldsymbol{\Delta}$ & \textbf{95\% CI} & $\boldsymbol{p}$ \\
\midrule
\multirow{3}{*}{LLaVA-V0-7B}
  & PATH-VQA & $+35.0$ & $[29.6,\;40.4]$ & $6.5\!\times\!10^{-32}$ \\
  & SLAKE    & $+22.4$ & $[15.7,\;29.2]$ & $4.1\!\times\!10^{-10}$ \\
  & VQA-RAD  & $+24.3$ & $[16.2,\;32.7]$ & $5.6\!\times\!10^{-8}$ \\
\midrule
\multirow{3}{*}{Llama-3.2-11B}
  & PATH-VQA & $+26.4$ & $[21.2,\;31.6]$ & $2.4\!\times\!10^{-21}$ \\
  & SLAKE    & $+12.5$ & $[7.7,\;17.3]$  & $1.2\!\times\!10^{-6}$ \\
  & VQA-RAD  & $+8.8$  & $[3.3,\;14.3]$  & $2.7\!\times\!10^{-3}$ \\
\midrule
\multirow{3}{*}{Qwen3-VL-8B}
  & PATH-VQA & $+27.2$ & $[22.6,\;31.8]$ & $1.0\!\times\!10^{-27}$ \\
  & SLAKE    & $+7.5$  & $[3.1,\;11.8]$  & $1.3\!\times\!10^{-3}$ \\
  & VQA-RAD  & $+5.1$  & $[-0.4,\;11.0]$ & $0.098$ \\
\bottomrule
\end{tabular}%
}

\caption{Paired significance analysis for probing vs.\ free-text prompting. $\Delta$ = probing$-$prompting accuracy (percentage points). Model abbreviations as in Table~\ref{tab:option_likelihood}.}
\label{tab:paired_significance}
\end{table}

\section{Cross-Dataset Generalization}
\label{sec:cross_dataset}

Table~\ref{tab:probing_results} reports probing accuracy for three models across all combinations of training and test datasets, including a generalist condition using OmniMedVQA training data.

\begin{table}[t]
\centering
\scriptsize
\setlength{\tabcolsep}{4pt}
\resizebox{\columnwidth}{!}{%
\begin{tabular}{llccc}
\toprule
\textbf{Model} & \textbf{Train Dataset} & \textbf{PATH-VQA} & \textbf{SLAKE} & \textbf{VQA-RAD} \\
\midrule
\multirow{4}{*}{InternVL3-1B}
  & OmniMedVQA & 55.40 & 72.29 & 61.60 \\
  & PATH-VQA   & \textbf{82.20} & 57.35 & 57.41 \\
  & SLAKE      & 63.00 & \textbf{83.13} & 58.94 \\
  & VQA-RAD    & 61.40 & 75.18 & \textbf{72.62} \\
\midrule
\multirow{4}{*}{LLaVA-V0-7B}
  & OmniMedVQA & 53.60 & 51.08 & 53.61 \\
  & PATH-VQA   & \textbf{84.60} & 53.49 & 55.13 \\
  & SLAKE      & 58.60 & \textbf{67.95} & 50.57 \\
  & VQA-RAD    & 62.00 & 54.70 & \textbf{66.92} \\
\midrule
\multirow{4}{*}{Qwen3-VL-2B}
  & OmniMedVQA & 54.20 & 67.71 & 55.89 \\
  & PATH-VQA   & \textbf{83.80} & 56.63 & 52.85 \\
  & SLAKE      & 58.60 & \textbf{84.10} & 60.84 \\
  & VQA-RAD    & 58.60 & 71.33 & \textbf{73.00} \\
\bottomrule
\end{tabular}%
}
\caption{Probing accuracy (\%) across models and training datasets. Bold indicates in-domain (train and test from same benchmark). PATH-VQA-trained probes show severe domain-locking, collapsing on radiology benchmarks. VQA-RAD-trained probes transfer well to SLAKE due to shared radiology domain.}
\label{tab:probing_results}
\end{table}

In-domain training consistently yields the strongest probing accuracy, confirming that domain-matched representations are most linearly separable.
However, a striking asymmetry emerges: models trained on PATH-VQA suffer dramatic performance collapses on SLAKE and VQA-RAD, with InternVL3-1B dropping to 57.35\% on SLAKE and 57.41\% on VQA-RAD, figures barely above random and well below even the generalist OmniMedVQA baseline of 72.29\% and 61.60\%.
This \emph{domain-locked} behavior is consistent across all three models and indicates that pathology-specific visual patterns in PATH-VQA induce highly specialized representations that do not transfer to radiology benchmarks.

In contrast, models trained on VQA-RAD exhibit substantial cross-domain transfer to SLAKE: 75.18\% for InternVL3-1B and 71.33\% for Qwen3-VL-2B.
This asymmetry likely reflects the shared radiology domain between VQA-RAD and SLAKE, where overlapping visual concepts support generalization.
The generalist OmniMedVQA condition produces consistently moderate results but never approaches in-domain performance, indicating that broad medical coverage alone is insufficient to match domain-specific representation learning.

\section{Binary vs.\ Multiclass Breakdown}
\label{app:binary_multiclass}

Section~\ref{sec:binary_multiclass} reports a summary comparison between binary yes/no items and true multiclass items. Table~\ref{tab:binary_breakdown} reports representative binary results, and Table~\ref{tab:multiclass_breakdown} reports true multiclass results on OmniMedVQA (400 examples, all 4-way; chance 25\%). Uniform random states the expected accuracy under uniform random predictions.

\begin{table}[t]
\centering
\resizebox{\columnwidth}{!}{%
\begin{tabular}{llccc}
\toprule
\textbf{Model} & \textbf{Method} & \textbf{PATH-VQA} & \makecell{\textbf{SLAKE}\\\textbf{Binary}} & \makecell{\textbf{VQA-RAD}\\\textbf{Binary}} \\
\midrule
Uniform random & --- & 50.00 & 50.00 & 50.00 \\
\midrule
\multirow{2}{*}{InternVL3-1B}
 & Prompting & 62.80 & 78.31 & 62.55 \\
 & MedProb & 82.20 & 85.35 & 70.52 \\
\midrule
\multirow{2}{*}{Qwen3-VL-8B}
 & Prompting & 59.00 & 78.31 & 69.72 \\
 & MedProb & 86.20 & 86.48 & 75.30 \\
\midrule
\multirow{2}{*}{MedVLThinker-3B}
 & Prompting & 59.60 & 65.35 & 69.72 \\
 & MedProb & 82.80 & 81.13 & 73.31 \\
\bottomrule
\end{tabular}%
}
\caption{Binary (yes/no) accuracy (\%) for representative models.}
\label{tab:binary_breakdown}
\end{table}

\begin{table}[t]
\centering
\resizebox{\columnwidth}{!}{%
\begin{tabular}{lccc}
\toprule
\textbf{Model} & \textbf{Uniform} & \textbf{Prompting} & \textbf{MedProb} \\
\midrule
InternVL3-1B & 25.00 & 67.75 & 71.75 \\
Qwen3-VL-8B & 25.00 & 61.75 & 80.75 \\
MedVLThinker-3B & 25.00 & 63.50 & 67.50 \\
\bottomrule
\end{tabular}%
}
\caption{True multiclass accuracy (\%) on OmniMedVQA (4-way, 400 examples; chance 25\%).}
\label{tab:multiclass_breakdown}
\end{table}

Probing improves substantially over prompting in both the binary and true multiclass settings. Binary gains remain large on PATH-VQA, SLAKE, and VQA-RAD, and probing outperforms prompting overall on the true 4-way multiclass items from OmniMedVQA.

\section{Rejection-Sampling with the Probe on Open-Ended Generations}
\label{app:rejection_sampling}

The main paper restricts evaluation to explicit multiple-choice Med-VQA. As a first step toward open-ended generation, we test whether the probe can score open-ended candidate answers in a rejection-sampling framework: a VLM generates many candidate answers to a question, and the probe scores each candidate, selecting the highest-scoring one.

We restrict this evaluation to the true multiclass items from OmniMedVQA, with 500 training and 100 test examples. The probe is trained to predict correct or incorrect given a single answer candidate (the question, image, and one open-ended candidate answer as input). At inference time, the VLM generates multiple open-ended outputs per question; the probe scores each sampled candidate, and we select the highest-scoring one. Table~\ref{tab:rejection_sampling} reports accuracy for the rejection-sampling probe, a no-probe baseline (a single sampled generation with no reranking), and an oracle that always selects the correct candidate when one exists among the samples.

\begin{table}[t]
\centering
\resizebox{\columnwidth}{!}{%
\begin{tabular}{lccc}
\toprule
\textbf{Model} & \makecell{\textbf{Rejection-}\\\textbf{sampling}\\\textbf{probe (ours)}} & \makecell{\textbf{Without}\\\textbf{probe}\\\textbf{(baseline)}} & \makecell{\textbf{Oracle}\\\textbf{(select correct}\\\textbf{candidate)}} \\
\midrule
Qwen2.5-VL-7B-Instruct & 0.13 & 0.09 & 0.19 \\
Llama-3.2-11B-Vision-Instruct & 0.19 & 0.15 & 0.27 \\
\bottomrule
\end{tabular}%
}
\caption{Accuracy of rejection sampling with the probe as a reranker over open-ended VLM generations, compared to a single-sample baseline (no probe) and an oracle upper bound.}
\label{tab:rejection_sampling}
\end{table}

The rejection-sampling probe improves over the no-probe baseline for both models, and narrows the gap to the oracle, which reflects the highest accuracy achievable given the sampled candidates. While preliminary, this experiment illustrates a practical route for applying MedProb beyond explicit multiple-choice options: rather than predicting an answer directly, the probe can rerank open-ended generations produced by a VLM. We view this as an initial demonstration rather than a complete open-ended Med-VQA solution, consistent with the limitation that most of our results target the multiple-choice/multiclass setting (see Limitations).

\section{Additional Layer-wise Probing Results}

This appendix collects the layer-wise probing accuracy plots for the remaining general-medical model pairs. Across families, clinical information is typically decodable from very early layers, but the effect of medical specialization is not uniform: some medical models improve the earliest and middle representations, whereas several general-purpose backbones recover or surpass them in the deeper layers. The pairwise discussions below summarize the dominant trend in each plot and point to the corresponding figure.

\paragraph{Stability of the Final Layers.} Figure~\ref{fig:layerwise} and the pairwise plots below show that probing accuracy is very stable across the later layers. Table~\ref{tab:last10_layers} summarizes the mean, standard deviation, and range of accuracy across the last 10 layers for two representative models, and compares against zero-shot prompting; all of the last 10 layers outperform prompting in every case.

\begin{table}[t]
\centering
\resizebox{\columnwidth}{!}{%
\begin{tabular}{llccc}
\toprule
\textbf{Model} & \textbf{Dataset} & \makecell{\textbf{Last-10}\\\textbf{(mean$\pm$std)}} & \makecell{\textbf{Last-10}\\\textbf{range}} & \textbf{Prompting} \\
\midrule
\multirow{3}{*}{Qwen2.5-VL-7B}
 & PATH-VQA & 82.06$\pm$0.70 & 80.40--82.80 & 62.40 \\
 & SLAKE    & 80.96$\pm$1.06 & 79.28--82.89 & 73.01 \\
 & VQA-RAD  & 75.82$\pm$1.79 & 72.24--77.57 & 71.10 \\
\midrule
\multirow{3}{*}{MedVLThinker-7B}
 & PATH-VQA & 82.78$\pm$1.10 & 81.00--84.40 & 64.60 \\
 & SLAKE    & 77.06$\pm$1.27 & 75.66--79.76 & 65.78 \\
 & VQA-RAD  & 73.95$\pm$1.18 & 71.86--75.67 & 65.78 \\
\bottomrule
\end{tabular}%
}
\caption{Accuracy (\%) across the final 10 layers (mean $\pm$ std, and range) compared to zero-shot prompting. All 10 final layers outperform promptingin every model/dataset combination.}
\label{tab:last10_layers}
\end{table}

\subsection{Gemma-4B vs. MedGemma-4B}
\begin{figure}[h]
    \centering
    \includegraphics[width=\linewidth]{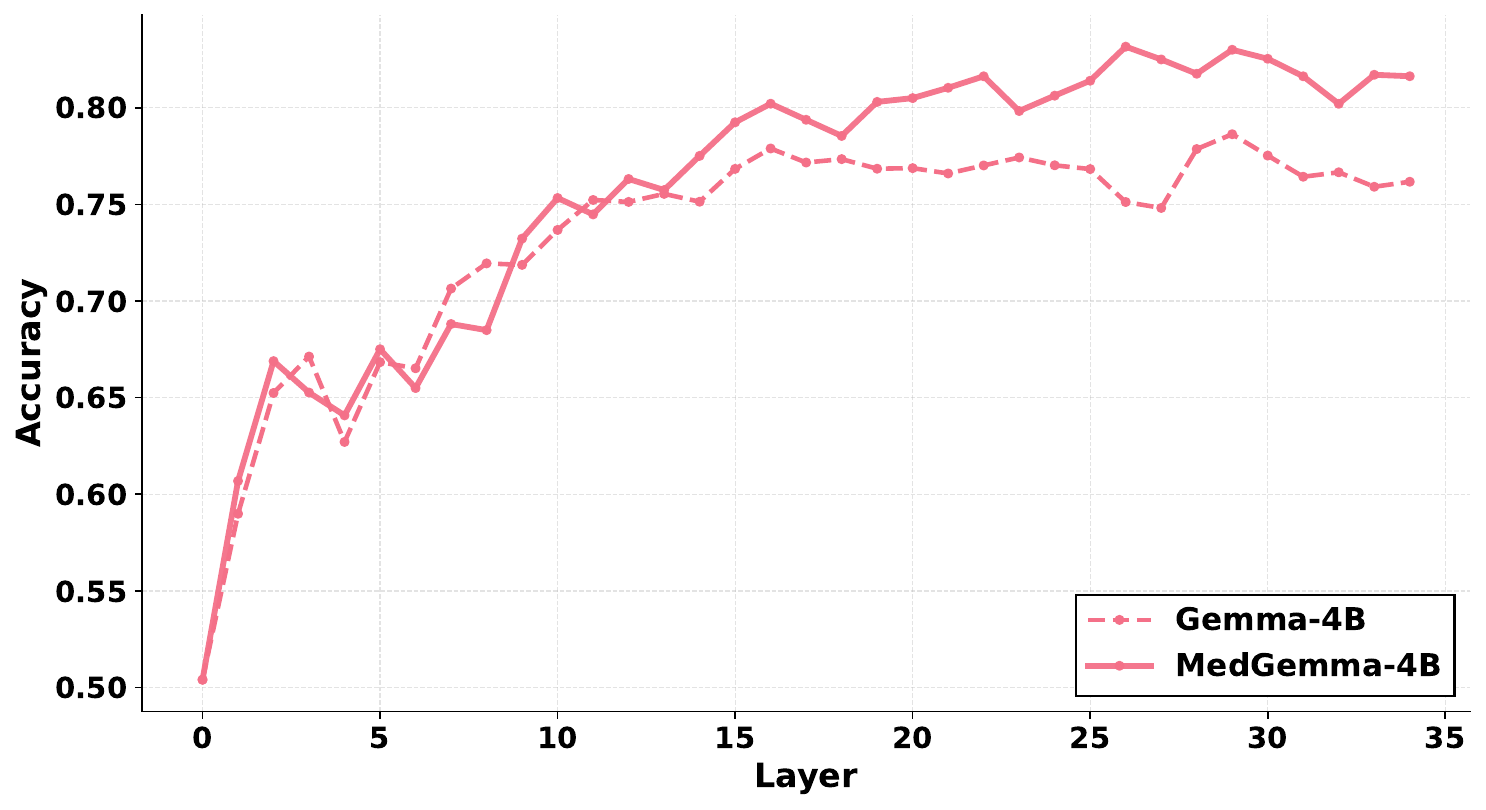}
    \caption{Layer-wise probing accuracy for Gemma-4B and MedGemma-4B. The medical model is competitive from the first layers and establishes a clearer advantage in the deeper part of the network.}
    \label{fig:layerwise-gemma-4b}
\end{figure}

Figure~\ref{fig:layerwise-gemma-4b} shows that MedGemma-4B is already slightly ahead in the early layers and then opens a clearer margin through the middle and late parts of the network. The separation becomes most visible in the final third of the model, suggesting that medical specialization improves how clinically relevant information is preserved and amplified in deeper representations.

\subsection{Gemma-27B vs. MedGemma-27B}
\begin{figure}[h]
    \centering
    \includegraphics[width=\linewidth]{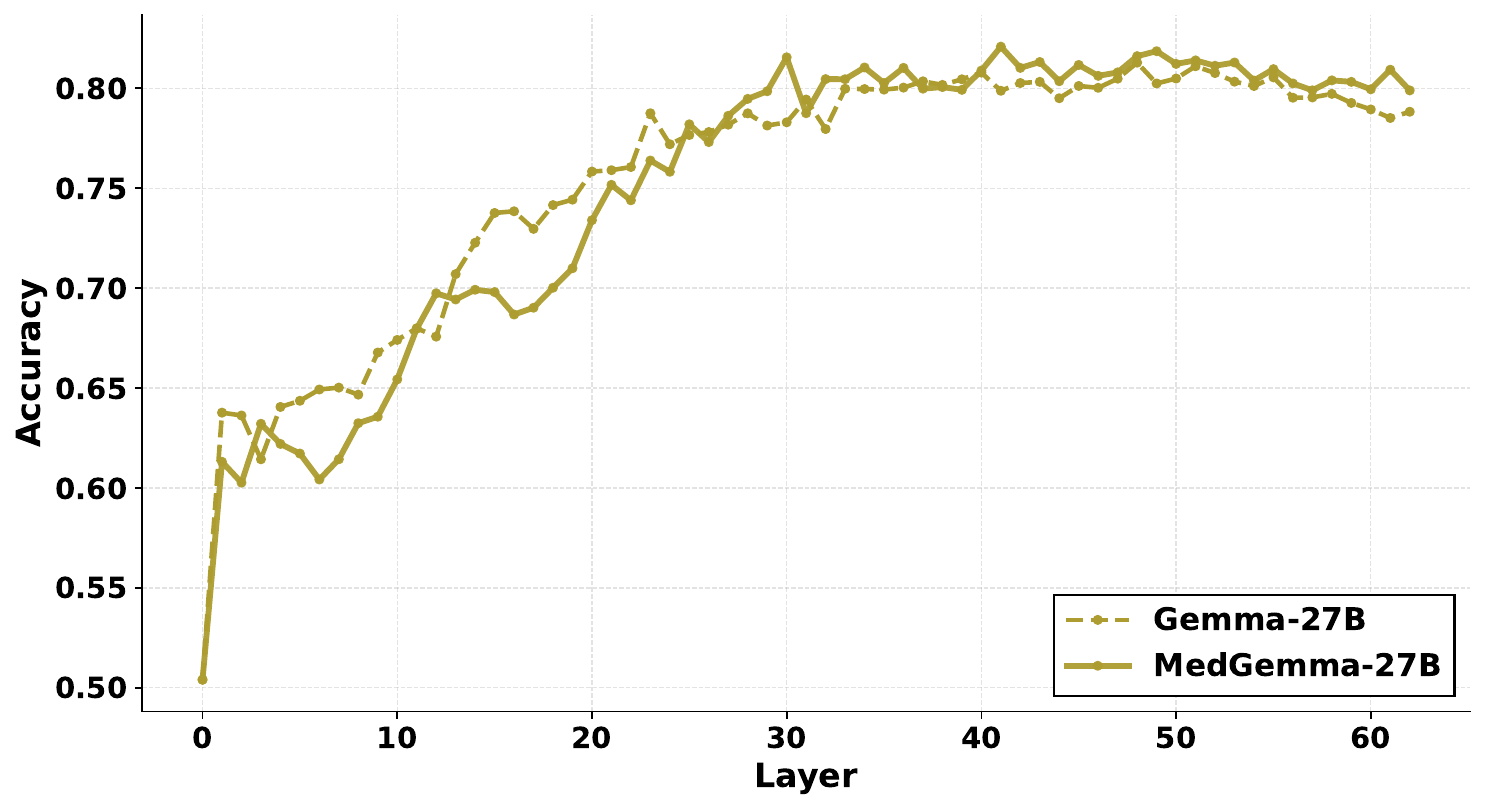}
    \caption{Layer-wise probing accuracy for Gemma-27B and MedGemma-27B. The two models remain close overall, with an early advantage for the general model and a more competitive late-layer profile for the medical variant.}
    \label{fig:layerwise-gemma-27b}
\end{figure}

As shown in Figure~\ref{fig:layerwise-gemma-27b}, the larger Gemma pair is much more balanced. Gemma-27B leads more often in the early layers, but the two curves converge in the middle of the model and MedGemma-27B becomes slightly stronger in part of the late-layer region, indicating that scaling reduces the gap between the general and medically adapted variants.

\subsection{Open-Flamingo-9B vs. Med-Flamingo-9B}
Figure~\ref{fig:layerwise-open-flamingo-9b} presents one of the cleanest specialization gains in the appendix. Med-Flamingo-9B stays above Open-Flamingo-9B for almost the entire depth of the model, and the gap remains fairly stable from shallow to deep layers, implying that the medical adaptation improves clinical separability throughout the representation stack rather than only at the end.

\begin{figure}[h]
    \centering
    \includegraphics[width=\linewidth]{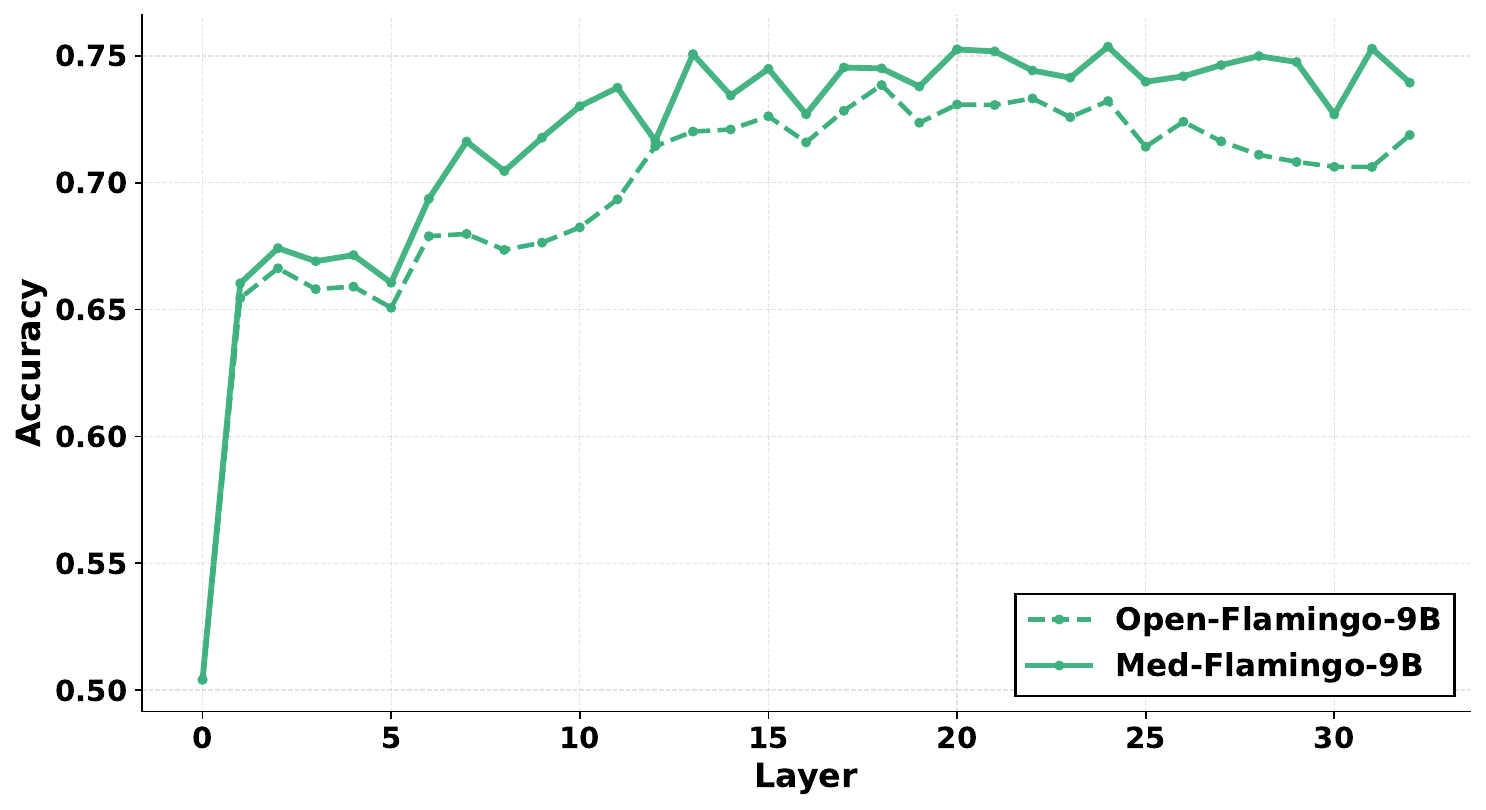}
    \caption{Layer-wise probing accuracy for Open-Flamingo-9B and Med-Flamingo-9B. The medical model maintains a consistent advantage across nearly all layers.}
    \label{fig:layerwise-open-flamingo-9b}
\end{figure}

\subsection{InternVL3-1B vs. BioMed-InternVL3-1B}
\begin{figure}[h]
    \centering
    \includegraphics[width=\linewidth]{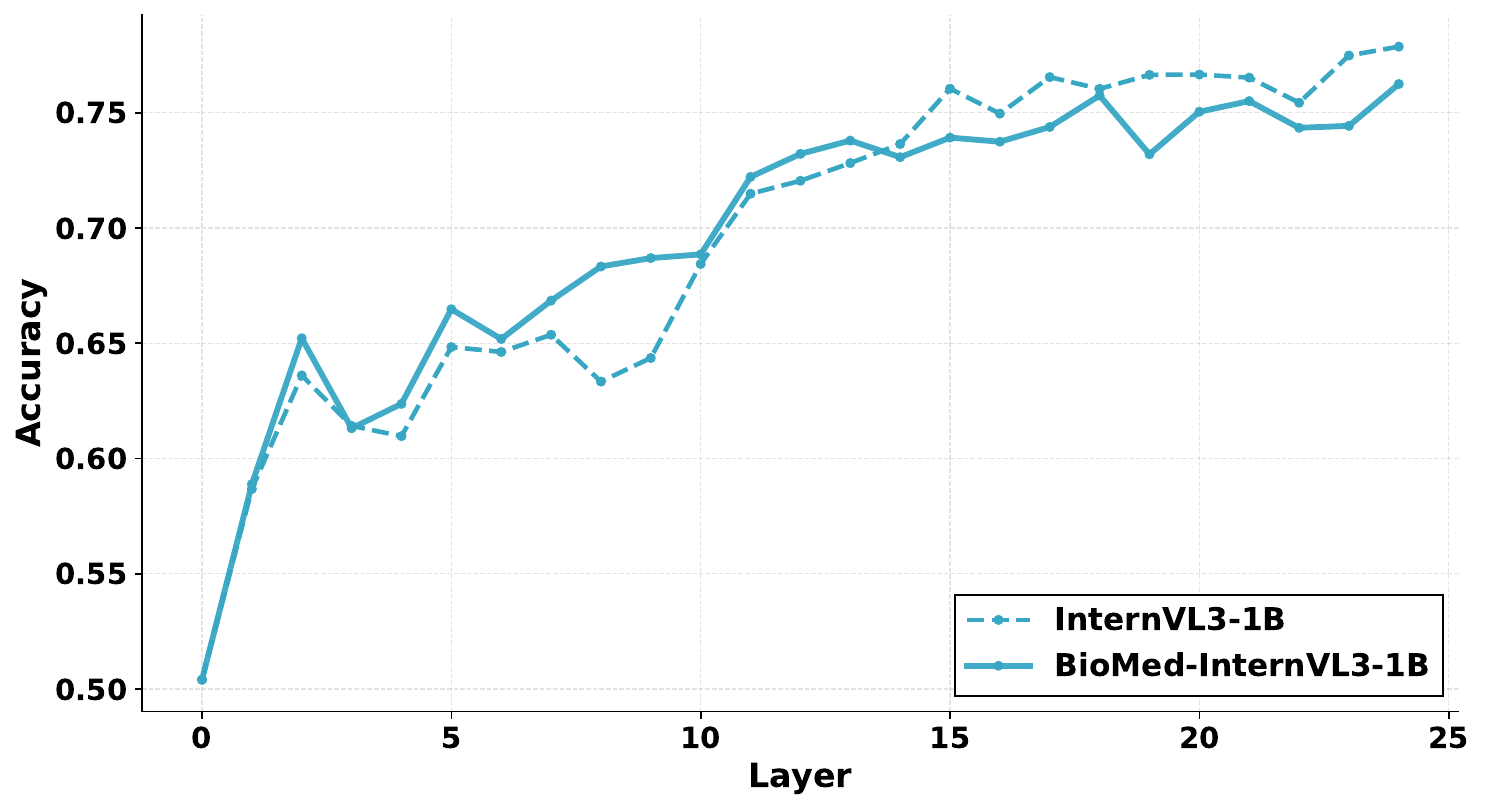}
    \caption{Layer-wise probing accuracy for InternVL3-1B and BioMed-InternVL3-1B. The medical model is stronger early, while the general model overtakes it in the deeper layers.}
    \label{fig:layerwise-internvl3-1b}
\end{figure}

The pattern in Figure~\ref{fig:layerwise-internvl3-1b} is stage-dependent. BioMed-InternVL3-1B has a small advantage across much of the early and middle layers, but the general InternVL3-1B catches up in the final third and finishes with the stronger deep-layer accuracy. This suggests that medical tuning helps earlier clinical extraction, while the base model retains a stronger late-layer refinement stage.

\subsection{LLaVA-7B vs. LLaVA-Med-7B}
\begin{figure}[h]
    \centering
    \includegraphics[width=\linewidth]{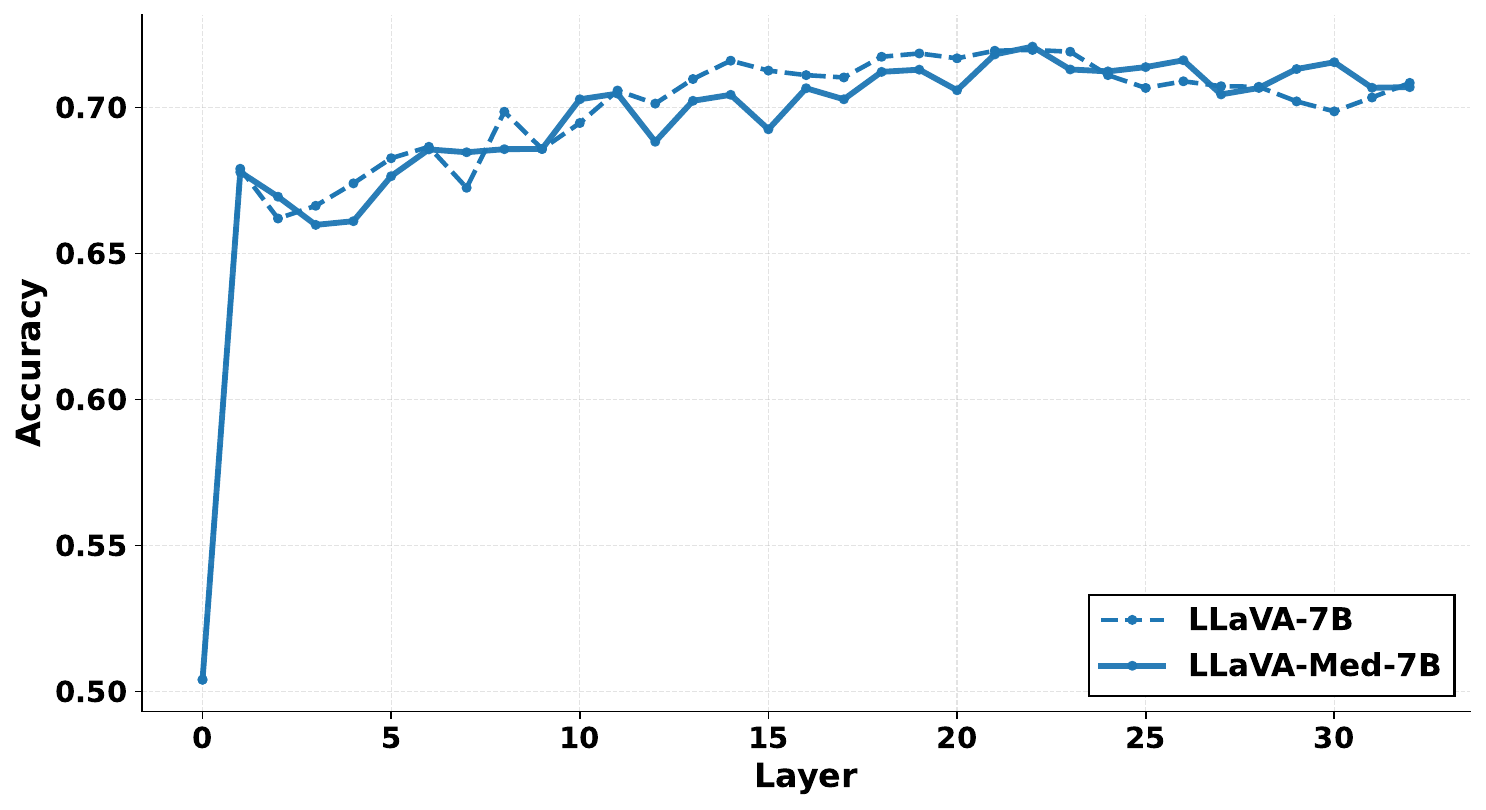}
    \caption{Layer-wise probing accuracy for LLaVA-7B and LLaVA-Med-7B. The two curves are tightly matched, with only small and localized differences across depth.}
    \label{fig:layerwise-llava-7b}
\end{figure}

Figure~\ref{fig:layerwise-llava-7b} shows a near-overlap between LLaVA-7B and LLaVA-Med-7B. The curves cross several times, the mid-layer region slightly favors the general model, and both models peak at almost the same depth and accuracy. For this pair, medical adaptation appears to change the distribution of information only marginally.

\subsection{LLaMA3.2-11B-Vision vs. BioMed-LLaMA3.2-11B}
\begin{figure}[h]
    \centering
    \includegraphics[width=\linewidth]{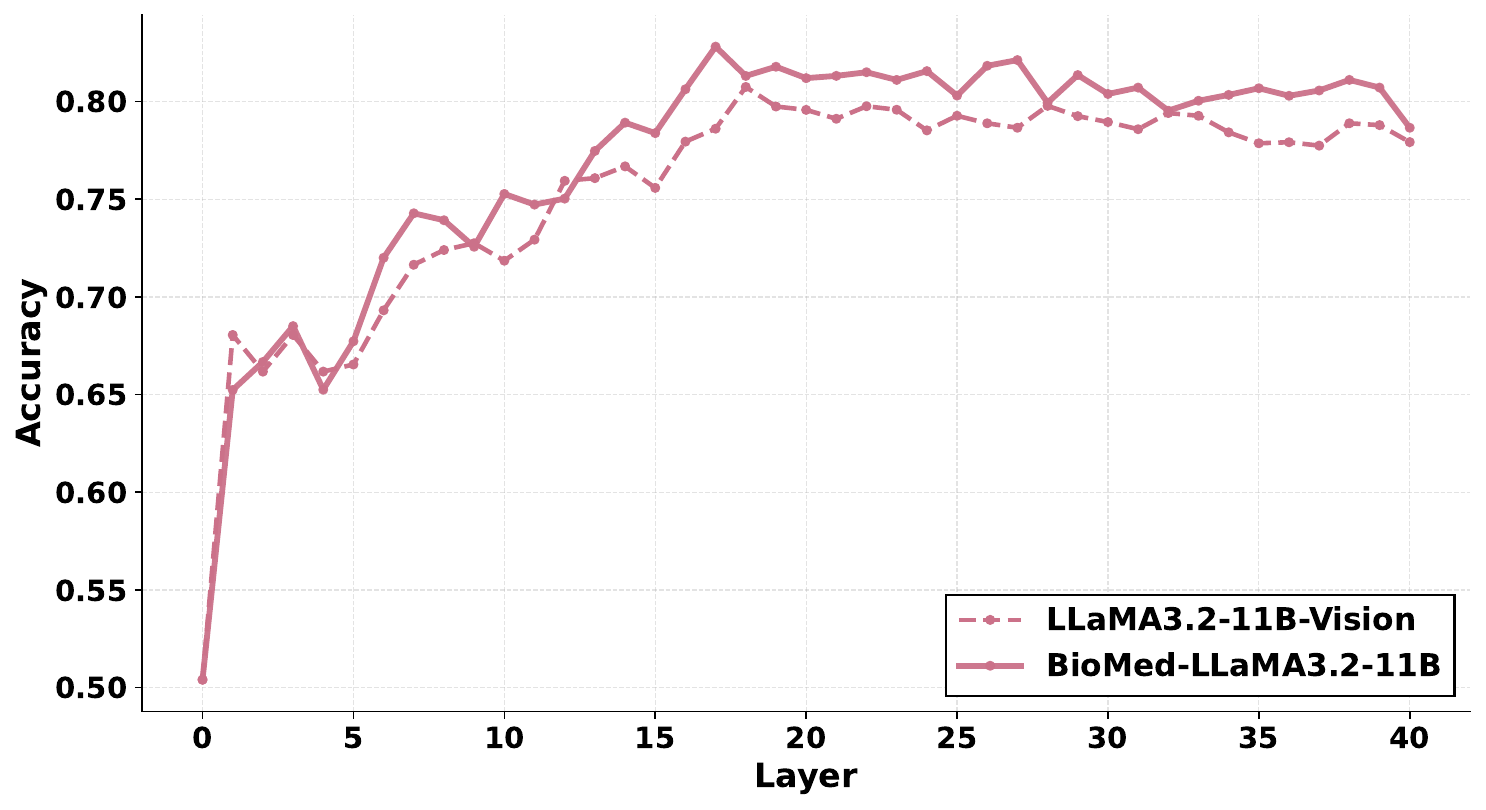}
    \caption{Layer-wise probing accuracy for LLaMA3.2-11B-Vision and BioMed-LLaMA3.2-11B. The medical model holds a modest but persistent advantage across most of the depth.}
    \label{fig:layerwise-llama32-11b-vision}
\end{figure}

In Figure~\ref{fig:layerwise-llama32-11b-vision}, BioMed-LLaMA3.2-11B is generally ahead after the first few layers and keeps that margin through most of the network. The difference is not dramatic, but it is persistent, which is consistent with a medical adaptation that improves clinical linear separability without radically changing the overall shape of the representation trajectory.

\subsection{Qwen2-VL-2B vs. BioMed-Qwen2-VL-2B}
\begin{figure}[h]
    \centering
    \includegraphics[width=\linewidth]{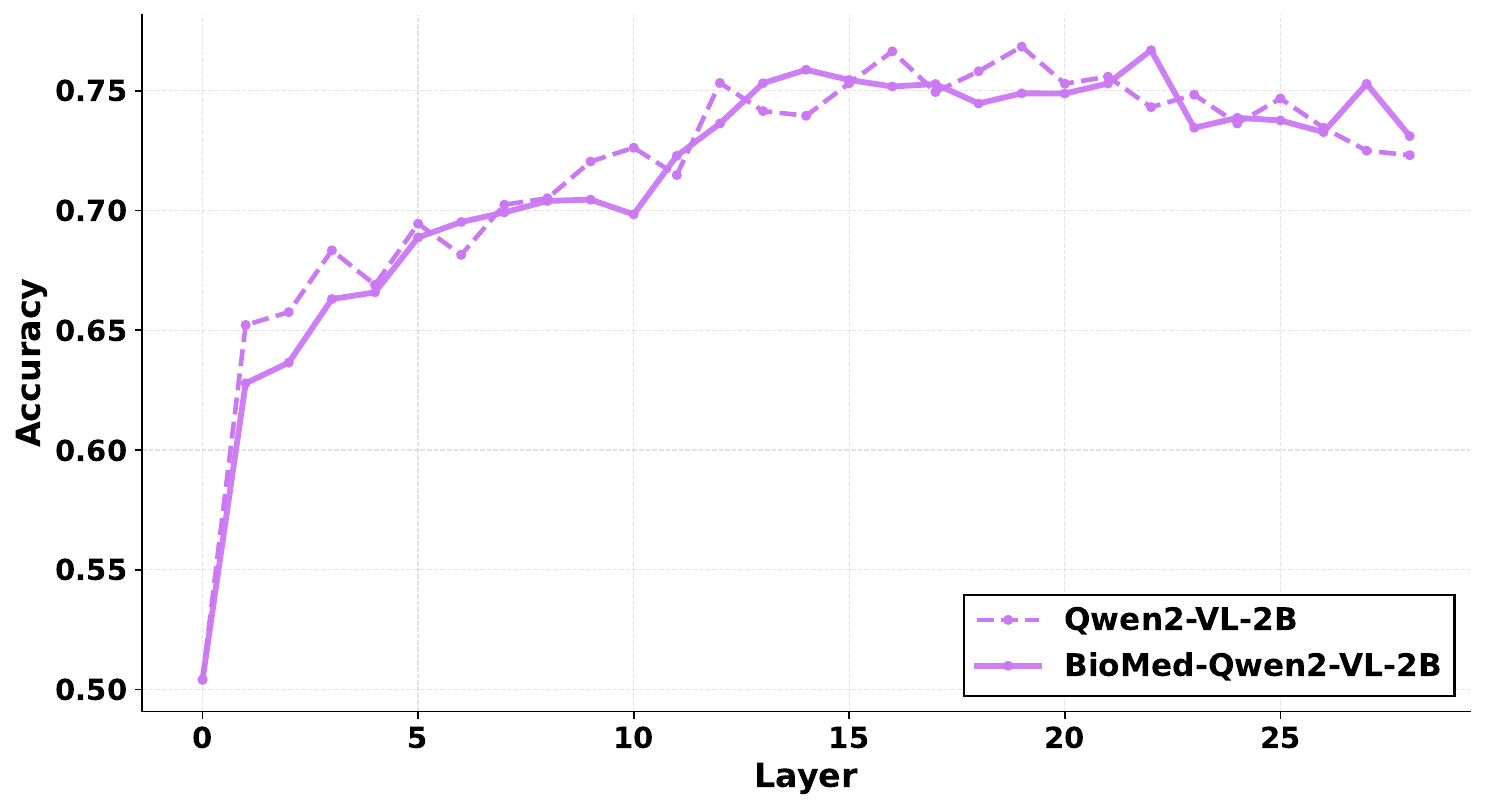}
    \caption{Layer-wise probing accuracy for Qwen2-VL-2B and BioMed-Qwen2-VL-2B. The general model has a small early-to-middle advantage and reaches a slightly higher late-layer maximum.}
    \label{fig:layerwise-qwen2-vl-2b}
\end{figure}

Figure~\ref{fig:layerwise-qwen2-vl-2b} indicates that Qwen2-VL-2B is slightly stronger through the early and middle layers, while the medical variant stays close and reduces the gap toward the end. The general model nevertheless achieves the higher late-layer peak, suggesting that domain adaptation does not consistently dominate once the backbone reaches its deepest representations.

\subsection{Qwen2.5-VL-3B vs. MedVLThinker-RL-3B}
\begin{figure}[h]
    \centering
    \includegraphics[width=\linewidth]{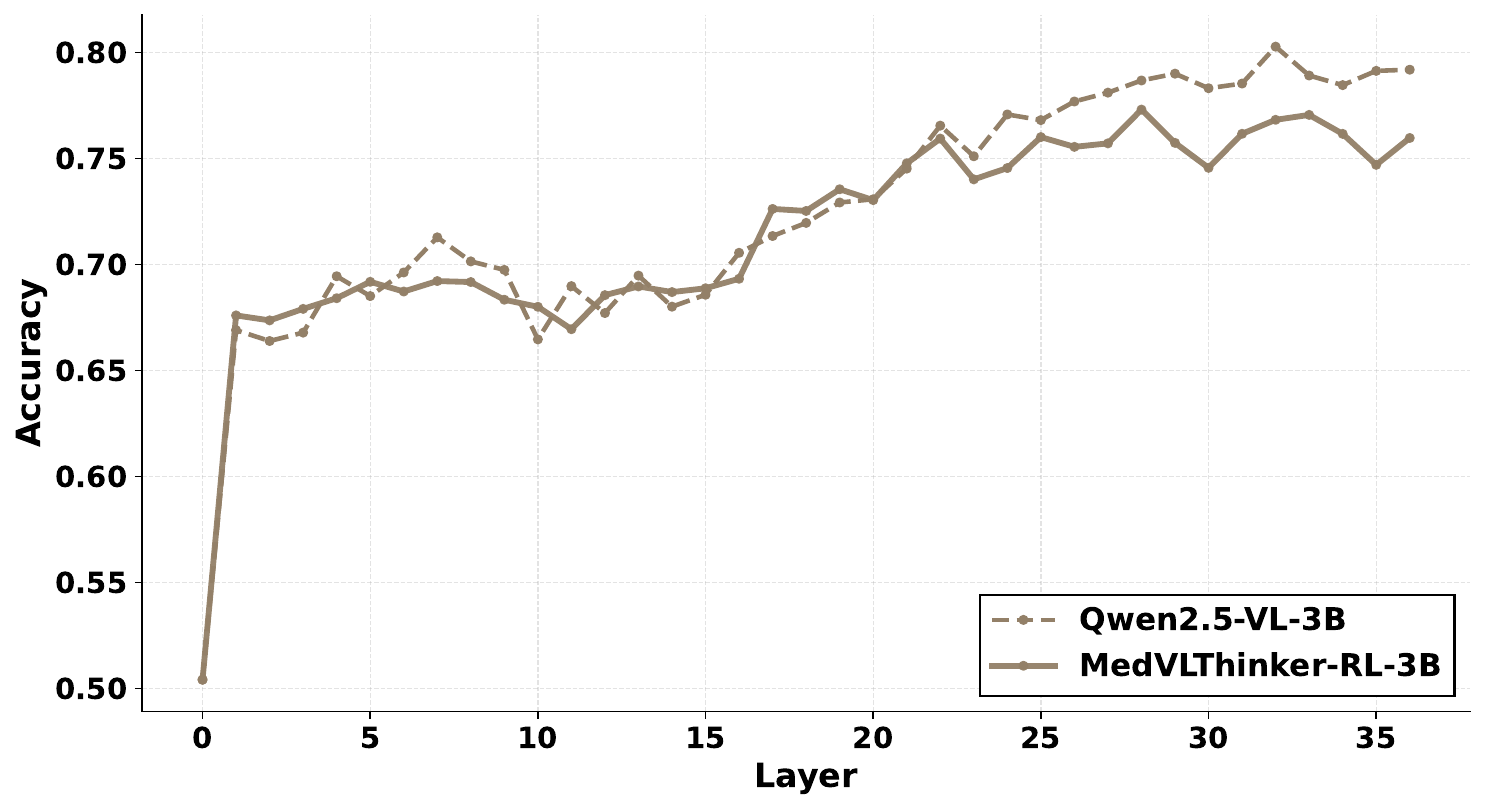}
    \caption{Layer-wise probing accuracy for Qwen2.5-VL-3B and MedVLThinker-RL-3B. The curves stay close early on, but the general model pulls ahead in the deepest layers.}
    \label{fig:layerwise-qwen25-vl-3b}
\end{figure}

As shown in Figure~\ref{fig:layerwise-qwen25-vl-3b}, Qwen2.5-VL-3B and MedVLThinker-RL-3B remain close for much of the network, but the general model separates more clearly in the final third and attains the stronger peak accuracy. This pattern suggests that the base Qwen2.5-VL-3B keeps a deeper-layer advantage even when the medical model is competitive in the earlier stages.

\subsection{Qwen2.5-VL-32B vs. MedVLThinker-RL-32B}
\begin{figure}[h]
    \centering
    \includegraphics[width=\linewidth]{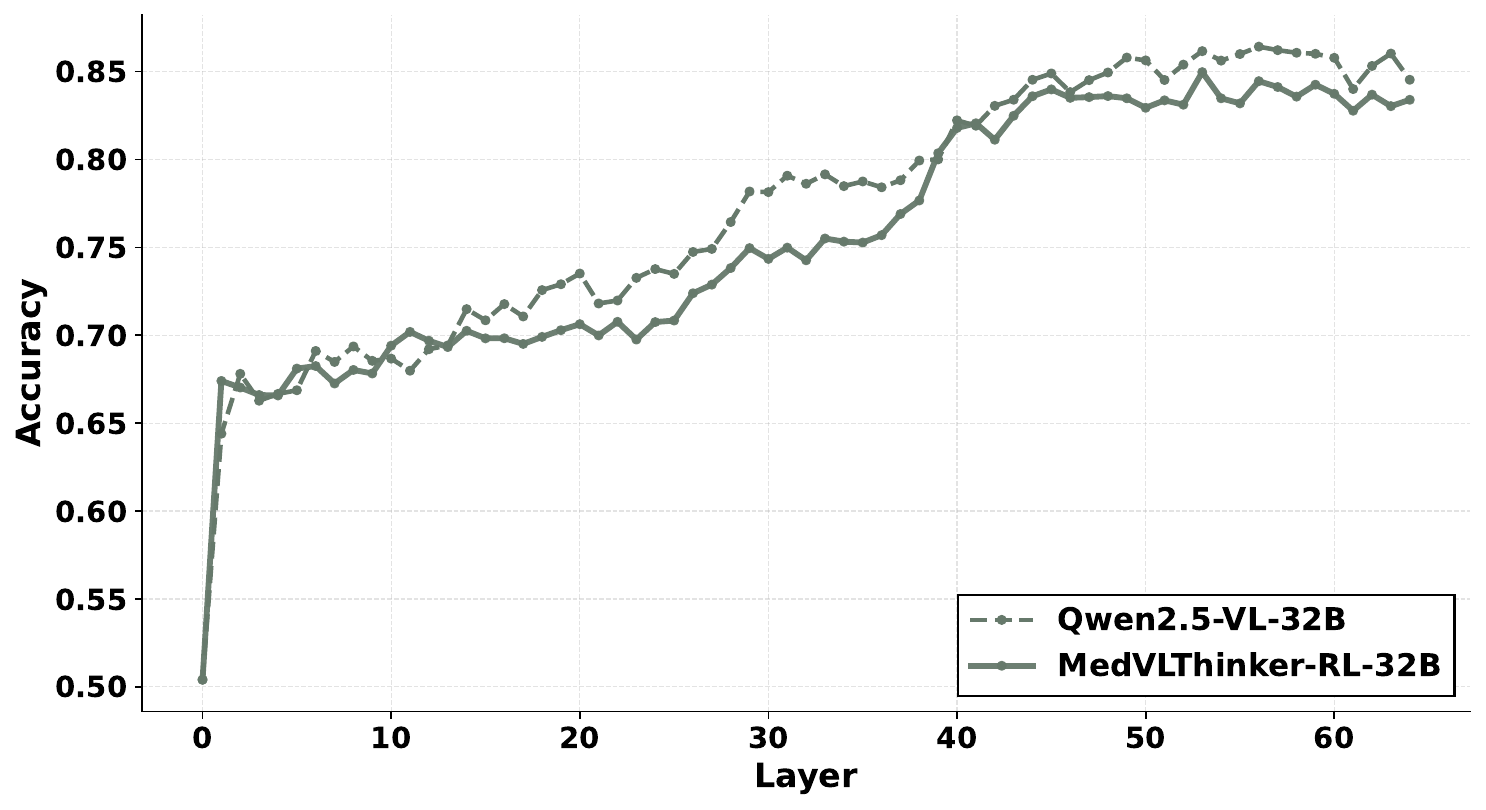}
    \caption{Layer-wise probing accuracy for Qwen2.5-VL-32B and MedVLThinker-RL-32B. The general model leads through most middle and late layers and achieves the higher peak accuracy.}
    \label{fig:layerwise-qwen25-vl-32b}
\end{figure}

Figure~\ref{fig:layerwise-qwen25-vl-32b} shows a broad advantage for Qwen2.5-VL-32B. After a few minor early exchanges, the general model is ahead for most middle and late layers and reaches the higher overall maximum. The medical variant remains competitive, but at this scale the general backbone appears to retain stronger clinically decodable information in deep representations.

\subsection{Qwen3-VL-2B vs. MediX-R1-2B}
\begin{figure}[h]
    \centering
    \includegraphics[width=\linewidth]{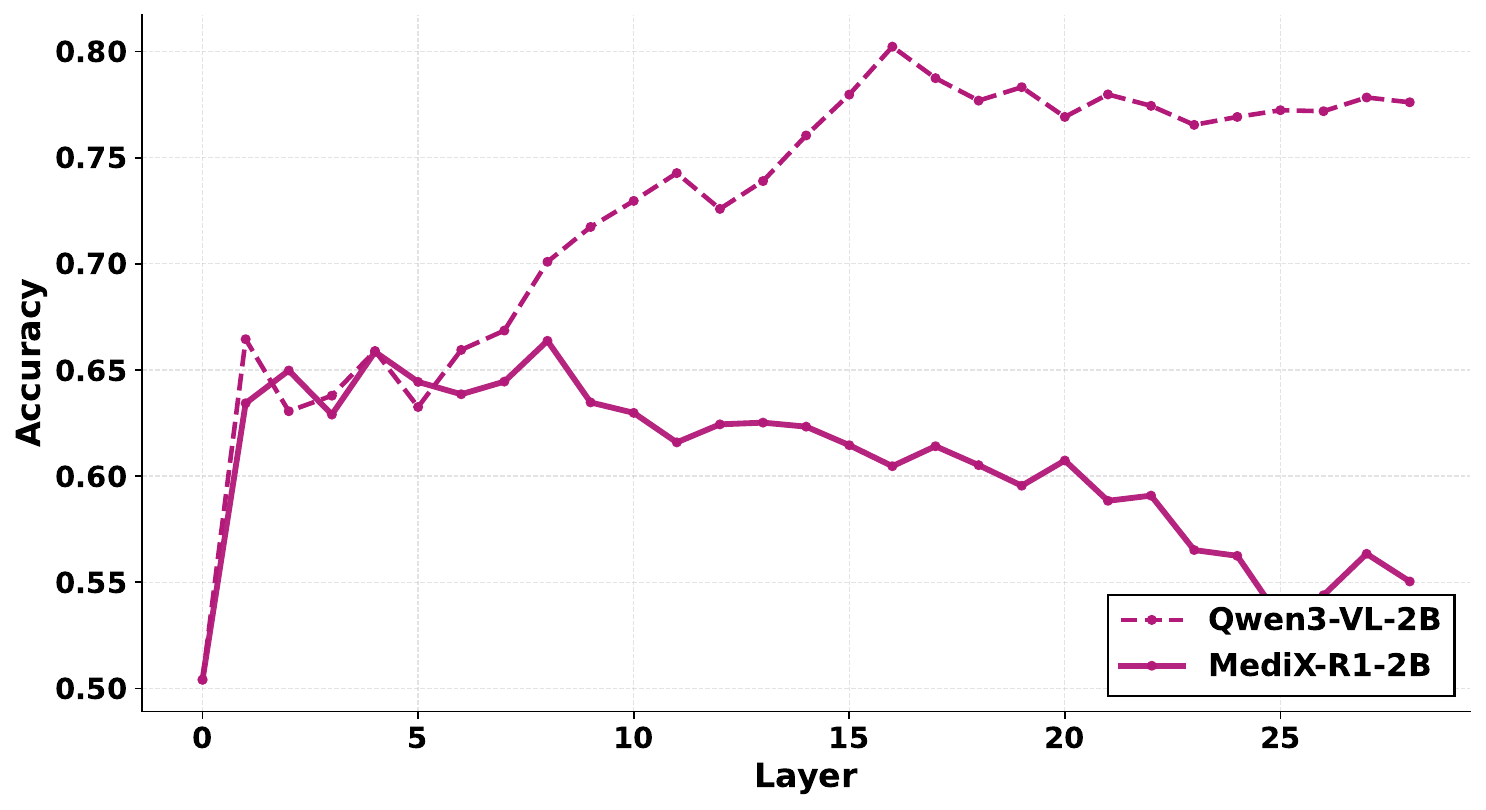}
    \caption{Layer-wise probing accuracy for Qwen3-VL-2B and MediX-R1-2B. The general model quickly establishes a large and growing advantage across the deeper layers.}
    \label{fig:layerwise-qwen3-vl-2b}
\end{figure}

Figure~\ref{fig:layerwise-qwen3-vl-2b} exhibits the largest gap among all appendix pairs. After a brief and noisy start, Qwen3-VL-2B moves decisively ahead and the margin widens sharply through the middle and final layers, while MediX-R1-2B peaks much earlier and at a substantially lower level. This suggests that the medical variant loses a considerable amount of deep-layer clinical separability relative to the base model.

\subsection{Qwen3-VL-4B vs. MedMo-4B}
\begin{figure}[h]
    \centering
    \includegraphics[width=\linewidth]{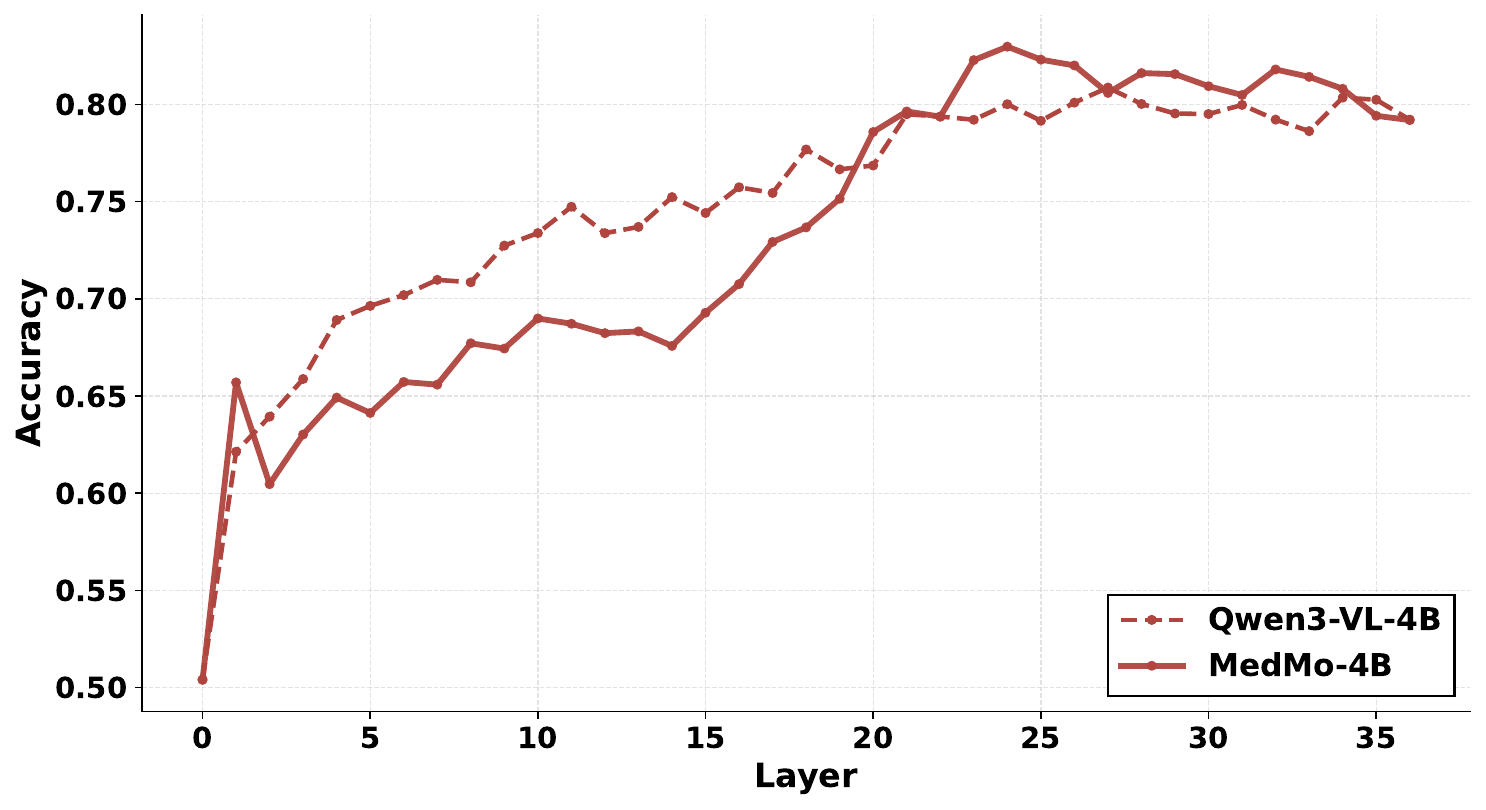}
    \caption{Layer-wise probing accuracy for Qwen3-VL-4B and MedMo-4B. The general model leads across much of the network, but the medical model rebounds late and reaches a slightly higher maximum.}
    \label{fig:layerwise-qwen3-vl-4b}
\end{figure}

The curves in Figure~\ref{fig:layerwise-qwen3-vl-4b} show a split trend. Qwen3-VL-4B leads for much of the first two thirds of the model, but MedMo-4B recovers in the later layers and eventually reaches the higher absolute peak. The result points to a medical model that is weaker in the intermediate regime yet competitive again at the deepest layers.

\subsection{Qwen3-VL-8B vs. MedMo-8B}
\begin{figure}[h]
    \centering
    \includegraphics[width=\linewidth]{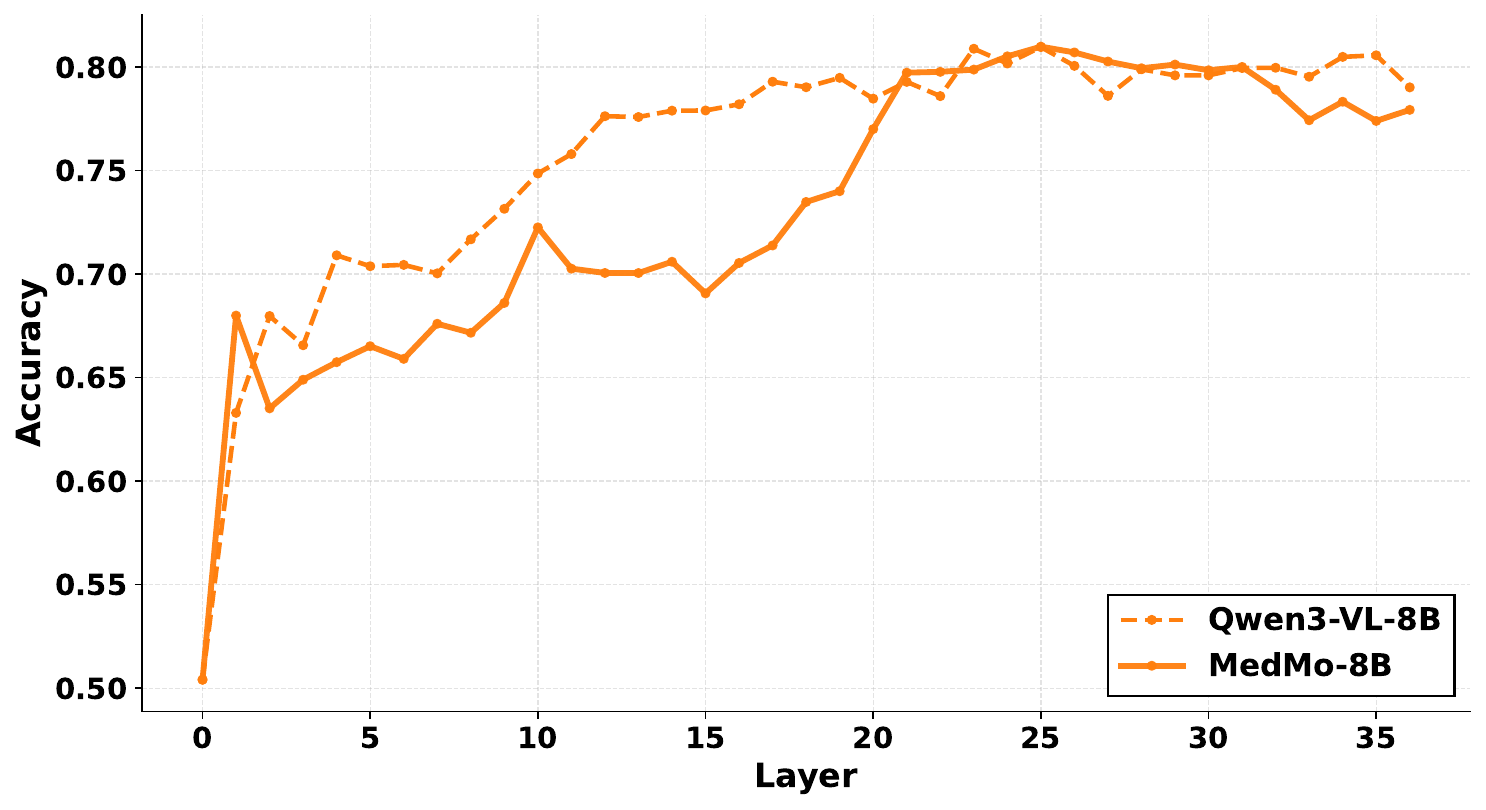}
    \caption{Layer-wise probing accuracy for Qwen3-VL-8B and MedMo-8B. The general model leads through most early and middle layers, while the medical model catches up in the deeper region.}
    \label{fig:layerwise-qwen3-vl-8b}
\end{figure}

Figure~\ref{fig:layerwise-qwen3-vl-8b} follows a related but more balanced pattern. Qwen3-VL-8B is stronger across most early and middle layers, whereas MedMo-8B narrows the gap later and nearly matches the general model at the best-performing depth. The medical adaptation therefore improves late-layer competitiveness, but it does not overturn the general model's broader lead across the stack.

\subsection{Qwen3-VL-30B vs. MediX-R1-30B}
\begin{figure}[h]
    \centering
    \includegraphics[width=\linewidth]{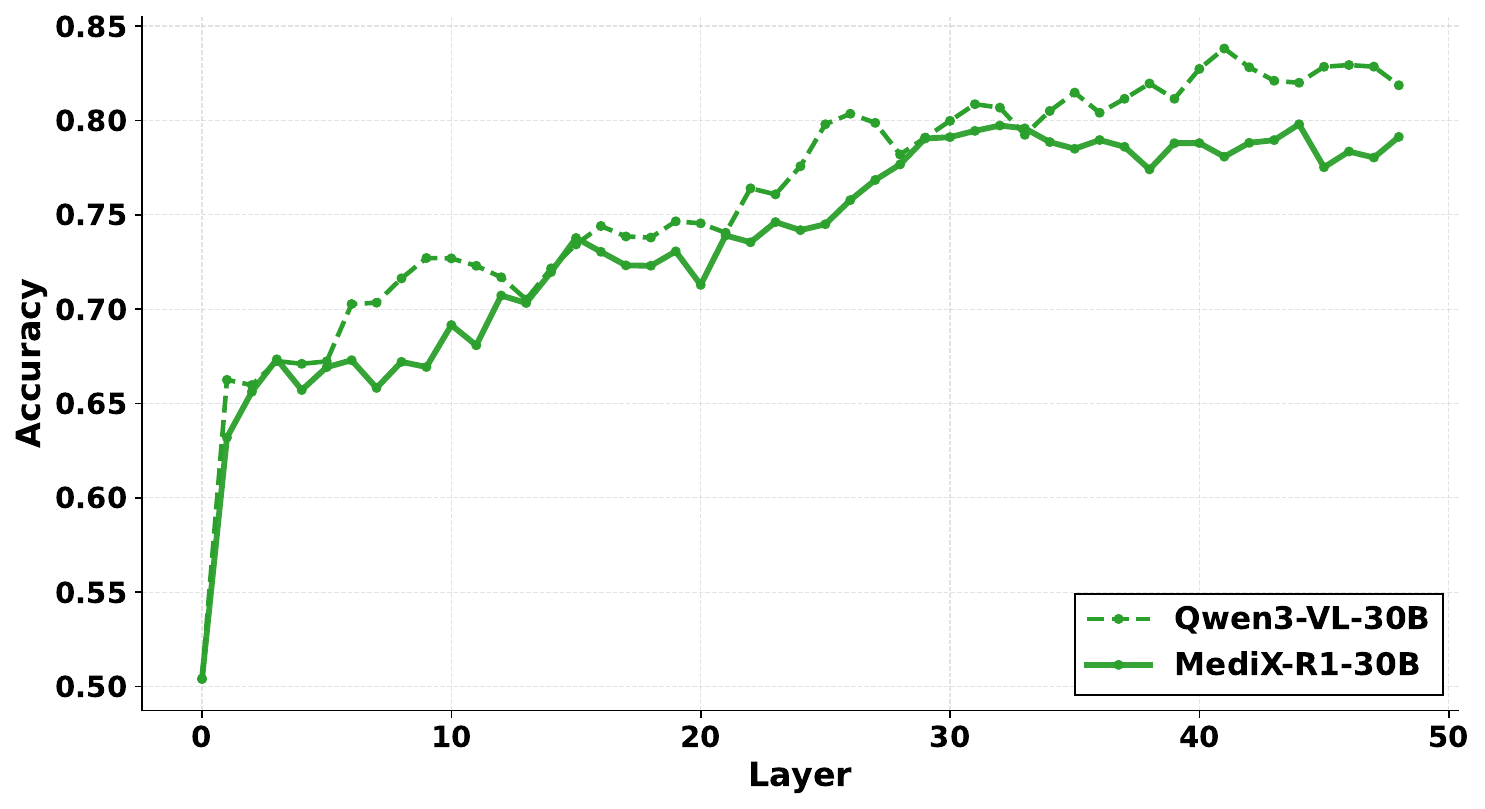}
    \caption{Layer-wise probing accuracy for Qwen3-VL-30B and MediX-R1-30B. The general model stays ahead across almost the entire network and widens the gap again in the deepest layers.}
    \label{fig:layerwise-qwen3-vl-30b}
\end{figure}

Finally, Figure~\ref{fig:layerwise-qwen3-vl-30b} shows a strong and stable advantage for Qwen3-VL-30B. Aside from a few isolated layer crossings, the general model remains ahead almost everywhere, and the gap grows again in the final layers where the best deep representations emerge. This indicates that, for the 30B pair, the base Qwen3-VL model preserves the more linearly accessible clinical information across the full depth of the network.

\newtcolorbox[]{modelbox}[2][]{
  colback   = gray!7,
  colframe  = gray!45,
  fonttitle = \bfseries\small,
  title     = {#2},
  left      = 6pt, right = 6pt, top = 4pt, bottom = 4pt,
  arc       = 3pt,
  breakable,
  #1
}

\section{Extended Error Analysis}
\label{app:error_analysis}

This appendix provides representative case studies that support the qualitative error analysis
presented in Section~\ref{sec:error_analysis}.
For each of the three behavioral categories, we first present a summary table of the selected
examples, then give a per-example breakdown that includes a verbatim excerpt from the model's
generated response, the prediction made by the linear probe, the ground-truth label, and an
interpretation of why each method succeeded or failed.
Examples are drawn from the 300-response analysis described in Section~\ref{sec:error_analysis}
(10 samples $\times$ 3 datasets $\times$ 10 models).
We denote the ground truth label as \textbf{GT}, the prompt-based prediction as
\textbf{Prompt}, and the linear probe prediction as \textbf{Probe}.
Correct predictions are marked \textcolor{teal}{\checkmark} and incorrect ones \textcolor{red}{$\times$}.

\subsection{Where MedProb Improves Over Prompting}
\label{app:probe_fixes}

These are cases in which the probe recovers the correct answer despite the prompted model
reasoning itself into a wrong conclusion.
The common thread is that prompting introduces hallucinated or misapplied medical language that
overrides a correct visual signal already present in the model's latent representation.

\begin{table}[h]
\centering
\scriptsize
\setlength{\tabcolsep}{3pt}
\renewcommand{\arraystretch}{1.15}

\begin{tabularx}{\columnwidth}{c X c c c X}
\toprule
\textbf{Ex.} & \textbf{Modality / Task} & \textbf{GT} & \textbf{Prompt} & \textbf{Probe} & \textbf{Failure Pattern} \\
\midrule
E-1 & Histopathology (soft-tissue mass classification)
& B (No)
& A \textcolor{red}{$\times$}
& B \textcolor{teal}{\checkmark}
& Hallucinated pathology narrative \\

E-2 & Brain MRI (imaging modality identification)
& B (No)
& A \textcolor{red}{$\times$}
& B \textcolor{teal}{\checkmark}
& Correct rule, misapplied context \\
\bottomrule
\end{tabularx}
\caption{Examples where MedProb succeeds and prompting fails
(\textbf{Probe} $=$ \textcolor{teal}{\checkmark},
\textbf{Prompt} $=$ \textcolor{red}{$\times$}).}
\label{tab:probe_fixes}
\end{table}

\bigskip

\noindent\textbf{Example E-1}

\begin{figure}[h]
\centering
\begin{tcolorbox}[
    title=Prompt model output,
    colback=gray!5,
    colframe=blue!50!black,
    boxrule=0.6pt,
    arc=2mm
]
\small

``Based on the image, the answer is \textbf{A: yes}.
The image shows a lobulated mass with a bluish cartilaginous hue infiltrating
the soft tissues.
This is a characteristic appearance of a cartilaginous tumor.''

\par\smallskip

\textit{
Prompt prediction:
\textcolor{red}{\textbf{A (Yes)} $\times$}
\quad
Probe prediction:
\textcolor{teal!70!black}{\textbf{B (No)} \checkmark}
\quad
GT:
\textcolor{teal!70!black}{\textbf{B (No)}}
}
\end{tcolorbox}
\caption{Semantic hallucination failure: the model accurately describes the visual features but constructs a false pathology narrative and answers incorrectly; the probe at layer~16 is unaffected by this chain-of-thought error.}
\label{fig:e1}
\end{figure}

As shown in Figure~\ref{fig:e1}, the prompted model correctly identifies two real visual features, the lobulated shape and
the bluish cartilaginous hue, but then constructs a pathology narrative from them
(``characteristic appearance of a cartilaginous tumor'') and uses that narrative to justify
answering \emph{yes}, the wrong answer.
This is a prototypical \textbf{semantic hallucination} failure:
the model describes what it sees accurately but draws a false clinical conclusion.
The linear probe, reading directly from layer-16 activations without generating any text,
is not susceptible to this chain-of-thought error and correctly outputs the \emph{no} label.
This pattern, where prompting over-interprets a real visual cue into a wrong binary answer, is
the single most common failure mode corrected by the probe.

\bigskip

\noindent\textbf{Example E-2}

\begin{figure}[h]
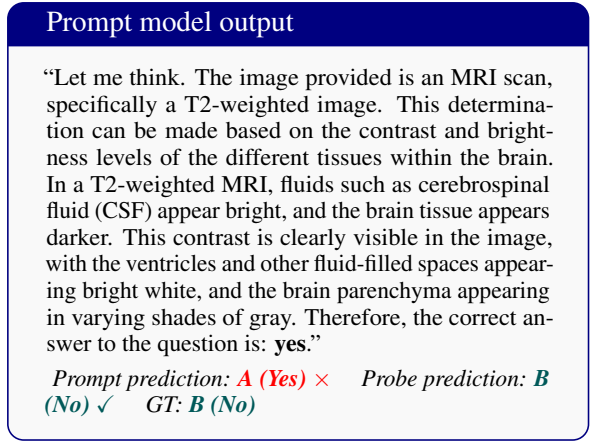

\centering
\begin{tcolorbox}[
    title=Prompt model output,
    colback=gray!5,
    colframe=blue!50!black,
    boxrule=0.6pt,
    arc=2mm
]
\small

``Let me think.
The image provided is an MRI scan, specifically a T2-weighted image.
This determination can be made based on the contrast and brightness levels of the different tissues
within the brain.
In a T2-weighted MRI, fluids such as cerebrospinal fluid (CSF) appear bright,
and the brain tissue appears darker.
This contrast is clearly visible in the image, with the ventricles and other fluid-filled spaces
appearing bright white, and the brain parenchyma appearing in varying shades of gray.
Therefore, the correct answer to the question is: \textbf{yes}.''

\par\smallskip

\textit{
Prompt prediction:
\textcolor{red}{\textbf{A (Yes)} $\times$}
\quad
Probe prediction:
\textcolor{teal!70!black}{\textbf{B (No)} \checkmark}
\quad
GT:
\textcolor{teal!70!black}{\textbf{B (No)}}
}
\end{tcolorbox}
\caption{Rule-application failure: the model's radiological reasoning is internally correct but misattributed to the wrong conclusion; the probe at layer~22 holds the correct answer in its activation pattern regardless of the verbalized reasoning chain.}
\label{fig:e2}
\end{figure}

Figure~\ref{fig:e2} demonstrates a subtler failure: the model's radiological reasoning is
internally correct.
It accurately defines T2 MRI contrast (CSF bright, parenchyma dark) and even observes those
features in the image.
Yet the final answer is wrong.
The model applied a textbook rule correctly but misattributed it to the wrong conclusion given the
actual question being asked.
This is a \textbf{rule-application failure}, not a knowledge failure.
The probe at layer 22 is unaffected by the verbalized reasoning chain and holds the correct answer
in its activation pattern.
The key insight is that a latent representation can encode the correct binary signal even when the
language decoder reasons convincingly but incorrectly from the same visual input.

\subsection{Where MedProb Fails to Improve}
\label{app:both_fail}

MedProb fails to resolve errors on questions where the underlying visual distinction is
too fine-grained, structurally ambiguous, or beyond the discriminative capacity of the
model's representations at any single layer.
In these cases, both prompting and probing fail simultaneously, indicating that the deficit lies
in the depth of perceptual encoding rather than in the elicitation mechanism.
These tasks tend to involve subtle radiological patterns (e.g., distinguishing ring-enhancing from
non-ring-enhancing lesions), complex histological tissue architecture, or multi-feature pathological
descriptions requiring simultaneous integration of several visual cues.
When representations do not encode a reliable decision boundary,
a linear probe has nothing to extract, and these questions expose the true perceptual ceiling
of the underlying vision-language model.
No elicitation mechanism can recover signal that is absent from the representation.

\begin{table}[h]
\centering
\scriptsize
\setlength{\tabcolsep}{3pt}
\renewcommand{\arraystretch}{1.15}

\begin{tabularx}{\columnwidth}{c X c c c X}
\toprule
\textbf{Ex.} & \textbf{Modality / Task} & \textbf{GT} & \textbf{Prompt} & \textbf{Probe} & \textbf{Failure Pattern} \\
\midrule
E-3 & Brain MRI (ring-enhancement pattern detection)
& A (Yes)
& B \textcolor{red}{$\times$}
& B \textcolor{red}{$\times$}
& Subtle spatial pattern beyond model's resolution \\

E-4 & Gross pathology (bowel wall skip-lesion detection)
& B (No)
& A \textcolor{red}{$\times$}
& A \textcolor{red}{$\times$}
& Fine-grained structural discrimination failure \\
\bottomrule
\end{tabularx}
\caption{Examples where both MedProb and prompting fail
(\textbf{Probe} $=$ \textcolor{red}{$\times$},
\textbf{Prompt} $=$ \textcolor{red}{$\times$}).}
\label{tab:both_fail}
\end{table}

\bigskip

\noindent\textbf{Example E-3}

Figure~\ref{fig:e3} shows responses from \textbf{two different evaluated models} on the same
question.
Each model generates its own prompt response and has its own probe applied at a different hidden
layer.
Their independent failure on the same question strengthens the conclusion: this is a task-level
difficulty, not an artifact of any single model.

\begin{figure}[h]
\centering
\begin{tcolorbox}[
    title=Model~1 --- prompt response,
    colback=gray!5,
    colframe=blue!50!black,
    boxrule=0.6pt,
    arc=2mm
]
\small

``Ring-enhancing lesions are typically seen in conditions like abscesses or some tumors,
where there is a ring of enhancement around a central area.
Looking at the image, the lesions here, do they show ring enhancement?
The lesions don't show a ring pattern.
So the answer would be B. no.''

\par\smallskip

\textit{
Prompt:
\textcolor{red}{\textbf{B (No)} $\times$}
\quad
Probe:
\textcolor{red}{\textbf{B (No)} $\times$}
\quad
GT:
\textcolor{green!50!black}{\textbf{A (Yes)}}
}
\end{tcolorbox}

\vspace{0.5em}

\begin{tcolorbox}[
    title=Model~2 --- prompt response,
    colback=gray!5,
    colframe=blue!50!black,
    boxrule=0.6pt,
    arc=2mm
]
\small

``The lesions have a more uniform appearance with a hypointense core and a surrounding area of
hyperintensity, which is more consistent with the appearance of multiple sclerosis plaques on
T2-weighted MRI images.
These plaques are often referred to as `Dawson's fingers' and are a common finding in patients
with multiple sclerosis.
Therefore, based on the MRI sequence presented, the lesions are not ring-enhancing.''

\par\smallskip

\textit{
Prompt:
\textcolor{red}{\textbf{B (No)} $\times$}
\quad
Probe:
\textcolor{red}{\textbf{B (No)} $\times$}
\quad
GT:
\textcolor{green!50!black}{\textbf{A (Yes)}}
}
\end{tcolorbox}
\caption{Perceptual ceiling failure: two independently trained models both fail on ring-enhancement detection across different reasoning strategies, with both prompt and probe predicting incorrectly, indicating a task-level perceptual limitation rather than a model-specific artifact.}
\label{fig:e3}
\end{figure}

Both models demonstrate genuine radiological expertise: Model~1 correctly defines ring enhancement,
and Model~2 appropriately considers multiple sclerosis as a differential diagnosis, even invoking
the clinical term ``Dawson's fingers.''
Yet both arrive at the wrong answer, and both probes also fail.
The fact that two independently trained models, using different reasoning strategies, produce the
same wrong conclusion, with both their latent representations and their language outputs
failing, indicates this is a \textbf{perceptual encoding failure} rather than a reasoning failure.
The fine-grained spatial distinction between ring-shaped and nodular enhancement is not encoded
reliably in any probed layer of either model.

\bigskip

\noindent\textbf{Example E-4}

\begin{figure}[h]
\centering
\begin{tcolorbox}[
    title=Prompt model output,
    colback=gray!5,
    colframe=blue!50!black,
    boxrule=0.6pt,
    arc=2mm
]
\small

``The image shows a segment of the bowel with a thickened wall and a narrowed lumen,
which is highlighted in pink.
The surrounding areas of the bowel appear to be normal or `skipped,' meaning they are not
involved in the thickening or narrowing.
This pattern is characteristic of certain conditions where segments of the bowel are affected,
while intervening areas remain normal.
Given the question asks if the corresponding area shows a segment of thickened wall with a narrow
lumen [...] and the intervening areas are uninvolved or skipped, the image supports this
description.''

\par\smallskip

\textit{
Prompt prediction:
\textcolor{red}{\textbf{A (Yes)} $\times$}
\quad
Probe prediction:
\textcolor{red}{\textbf{A (Yes)} $\times$}
\quad
GT:
\textcolor{green!50!black}{\textbf{B (No)}}
}
\end{tcolorbox}
\caption{Confident bilateral failure: the model identifies all key pathological features and concludes they are present, but the ground truth is negative; both prompt and probe agree on the wrong answer, indicating representational incapacity for this class of fine-grained bowel pathology.}
\label{fig:e4}
\end{figure}

As shown in Figure~\ref{fig:e4}, the prompted model explicitly identifies the key pathological
features
(thickened wall, narrowed lumen, skip areas) and concludes they are present, but the ground
truth says they are not.
The probe also predicts incorrectly.
Both methods agree on the wrong answer, suggesting that either the visual features being described
are genuinely ambiguous in this image, or the model is generating a plausible narrative without
reliable visual grounding.
This is a \textbf{confident bilateral failure}: neither elicitation mechanism provides a corrective
signal, because the underlying visual discrimination required is beyond the model's current
representational capacity for this class of fine-grained bowel pathology.

\subsection{Where MedProb Underperforms Prompting}
\label{app:prompt_wins}

MedProb underperforms prompting on questions that benefit from explicit semantic reasoning,
medical world knowledge, or direct visual labeling naturally expressible in language.
Prompting succeeds on these cases by leveraging the model's broad medical knowledge base, composing
pathological features into a clinical judgment, naming anatomy from contextual cues,
or reasoning step-by-step from visible findings to a diagnosis.
A linear probe is a shallow decoder capable only of recovering information that is
\emph{linearly separable} in a single layer's activation space.
Compositional reasoning, clinical inference chains, and world-knowledge-driven interpretation
are fundamentally nonlinear and distributed, precisely the features a linear probe is
least equipped to capture.
As a result, the probe performs worse on questions where the correct answer requires integrating
multiple visible features into a high-level clinical concept.

\begin{table}[h]
\centering
\caption{Examples where prompting succeeds but MedProb fails
(\textbf{Probe} $=$ \textcolor{red}{$\times$},
\textbf{Prompt} $=$ \textcolor{teal}{\checkmark}).}
\label{tab:prompt_wins}
\scriptsize
\setlength{\tabcolsep}{3pt}
\renewcommand{\arraystretch}{1.15}

\begin{tabularx}{\columnwidth}{c X c c c X}
\toprule
\textbf{Ex.} & \textbf{Modality / Task} & \textbf{GT} & \textbf{Prompt} & \textbf{Probe} & \textbf{Failure Pattern} \\
\midrule
E-5 & Lung histology (coarse tissue category recognition)
& A (Yes)
& A \textcolor{teal}{\checkmark}
& B \textcolor{red}{$\times$}
& Direct semantic label, probe misses coarse class \\

E-6 & Gross pathology (cardiac CHF classification)
& A (Yes)
& A \textcolor{teal}{\checkmark}
& B \textcolor{red}{$\times$}
& Compositional clinical reasoning required \\
\bottomrule
\end{tabularx}
\end{table}

\bigskip

\noindent\textbf{Example E-5}

\begin{figure}[h]
\centering
\begin{tcolorbox}[
    title=Prompt model output,
    colback=gray!5,
    colframe=blue!50!black,
    boxrule=0.6pt,
    arc=2mm
]
\small

``The image clearly shows lung tissue. So the answer is A.''

\par\smallskip

\textit{
Prompt prediction:
\textcolor{teal!70!black}{\textbf{A (Yes)} \checkmark}
\quad
Probe prediction:
\textcolor{red}{\textbf{B (No)} $\times$}
\quad
GT:
\textcolor{green!50!black}{\textbf{A (Yes)}}
}
\end{tcolorbox}
\caption{Coarse semantic label failure: prompting trivially names the tissue in one sentence and is correct; the probe fails on the same question, suggesting coarse semantic categories may not be linearly separable in intermediate representation layers.}
\label{fig:e5}
\end{figure}

Figure~\ref{fig:e5} is the starkest example in the dataset.
The prompt requires a single grounded visual label, produces it in one sentence, and is correct.
Yet the probe fails on this same question.
The brevity of the correct output is the key observation: when the answer reduces to
``name the tissue you see,'' language generation has a direct pathway to the correct answer
that the linear probe, a single affine transformation over one layer's activations, does not
reliably replicate.
This suggests that \textbf{coarse semantic category labels} (lung, liver, kidney) may not be
linearly separable in the intermediate representation layers probed, even when the language
model can verbalize them trivially.
The probe is decoding at the wrong abstraction level for this class of question.

\bigskip

\noindent\textbf{Example E-6}

\begin{figure}[h]
\centering
\begin{tcolorbox}[
    title=Prompt model output,
    colback=gray!5,
    colframe=blue!50!black,
    boxrule=0.6pt,
    arc=2mm
]
\small

``Let's think.
The image depicts a heart with significant pathological changes,
including thickening of the ventricular walls and the presence of fibrosis.
These features are consistent with the pathology of congestive heart failure (CHF),
where the heart muscle becomes enlarged and less efficient in pumping blood.
The dilated chambers and the presence of fibrotic tissue are indicative of the chronic changes
that occur in the heart due to the inability to pump blood effectively,
leading to a build-up of fluid in the body.
Therefore, based on the visible characteristics in the image,
it can be concluded that the heart shown is indeed affected by congestive heart failure.
The answer is: [yes]''

\par\smallskip

\textit{
Prompt prediction:
\textcolor{teal!70!black}{\textbf{A (Yes)} \checkmark}
\quad
Probe prediction:
\textcolor{red}{\textbf{B (No)} $\times$}
\quad
GT:
\textcolor{green!50!black}{\textbf{A (Yes)}}
}
\end{tcolorbox}
\caption{Compositional reasoning failure: the prompted model correctly chains wall thickening $\rightarrow$ fibrosis $\rightarrow$ dilated chambers $\rightarrow$ CHF; the probe cannot replicate this nonlinear multi-step clinical judgment from a single hidden layer.}
\label{fig:e6}
\end{figure}

Figure~\ref{fig:e6} illustrates the maximum expressiveness gap between prompting and probing.
The prompted model performs a multi-step compositional inference chain:
ventricular wall thickening $\rightarrow$ fibrosis $\rightarrow$ dilated chambers
$\rightarrow$ CHF diagnosis.
Each step draws on medical world knowledge to connect a visual feature to a pathophysiological
mechanism, and the chain is correct.
A linear probe operating on a single hidden layer cannot integrate these features into a CHF
judgment in the same way.
Doing so would require the concept ``CHF as a clinical gestalt'' to be linearly separable in
that layer's activation space, which it evidently is not for this model.
This case exemplifies the fundamental limitation of probing for questions that require
\textbf{compositional clinical reasoning}:
the information needed to answer correctly is distributed nonlinearly across the model's
computation, making it inaccessible to a shallow linear decoder.

\subsection{Summary}

Table~\ref{tab:error_summary} provides a consolidated view of all six examples across the three
behavioral categories.
Taken together, these cases suggest that the choice between probing and prompting is not simply
a question of which approach is generally superior, but depends critically on
\emph{what type of inference is required}:
probing outperforms prompting when language generation introduces spurious reasoning,
while prompting outperforms probing when the correct answer requires compositional knowledge
or grounded semantic labeling.
Questions where both fail point to genuine perceptual limitations of the underlying
vision-language model that neither elicitation strategy can overcome.

\begin{table}[h]
\centering
\scriptsize
\setlength{\tabcolsep}{2.5pt}
\renewcommand{\arraystretch}{1.1}

\begin{tabularx}{\columnwidth}{l c c c c X}
\toprule
\textbf{Cat.} & \textbf{Ex.} & \textbf{GT} & \textbf{Prompt} & \textbf{Probe} & \textbf{Key Failure Mode} \\
\midrule

\multirow{2}{*}{\textbf{P$>$M}}
& E-1 & B
& \textcolor{red}{$\times$} A
& \textcolor{teal}{\checkmark} B
& Hallucinated pathology narrative \\

& E-2 & B
& \textcolor{red}{$\times$} A
& \textcolor{teal}{\checkmark} B
& Textbook rule misapplied \\

\midrule

\multirow{2}{*}{\textbf{Both$\times$}}
& E-3 & A
& \textcolor{red}{$\times$} B
& \textcolor{red}{$\times$} B
& Ring-enhancement too subtle \\

& E-4 & B
& \textcolor{red}{$\times$} A
& \textcolor{red}{$\times$} A
& Bowel skip-lesion ambiguity \\

\midrule

\multirow{2}{*}{\textbf{M$>$P}}
& E-5 & A
& \textcolor{teal}{\checkmark} A
& \textcolor{red}{$\times$} B
& Coarse semantic label inaccessible to probe \\

& E-6 & A
& \textcolor{teal}{\checkmark} A
& \textcolor{red}{$\times$} B
& CHF requires compositional reasoning \\

\bottomrule
\end{tabularx}
\caption{Consolidated summary of all six error analysis examples.
Category abbreviations:
\textbf{P$>$M} = Probe better than prompting;
\textbf{Both$\times$} = Both fail;
\textbf{M$>$P} = Prompting better than probe.}
\label{tab:error_summary}
\end{table}

\section*{H Probing vs.\ Prompting Across Medical Imaging Modalities}

\begin{table*}[!t]
\centering
\scriptsize
\setlength{\tabcolsep}{3pt}
\renewcommand{\arraystretch}{1.05}
\resizebox{\textwidth}{!}{%
\begin{tabular}{l l c c c c c c c c >{\columncolor[RGB]{235,245,250}}c}
\toprule
\multirow{2}{*}{\textbf{Model}}
& \multirow{2}{*}{\textbf{Approach}}
& \multicolumn{9}{c}{\textbf{Modality}} \\
\cmidrule(lr){3-11}
&
& \textbf{CT}
& \textbf{Dermoscopy}
& \makecell{\textbf{Fundus}\\\textbf{Photography}}
& \makecell{\textbf{Microscopy}\\\textbf{Images}}
& \textbf{MRI}
& \textbf{OCT}
& \textbf{Ultrasound}
& \textbf{X-Ray}
& \textbf{Average} \\
\midrule
\multirow{2}{*}{Open-Flamingo-9B}
& Prompting & 18.0 & 26.0 & 36.0 & 30.0 & 26.0 & 26.0 & 24.0 & 40.0 & \textbf{28.0} \\
& Probing   & 62.0 & 58.0 & 76.0 & 56.0 & 60.0 & 66.0 & \textbf{82.0} & 64.0 & \textbf{66.0} \\
\noalign{\vskip 1mm}
\cline{2-11}
\noalign{\vskip 1mm}
\multirow{2}{*}{Med-Flamingo-9B}
& Prompting & 26.0 & 20.0 & 28.0 & 36.0 & 40.0 & 38.0 & 40.0 & 36.0 & 33.0 \\
& Probing   & \textbf{80.0} & 64.0 & \textbf{80.0} & 60.0 & 64.0 & 76.0 & 76.0 & 72.0 & \textbf{72.0} \\
\midrule
\multirow{2}{*}{InternVL3-1B}
& Prompting & 62.0 & 70.0 & 68.0 & 74.0 & 74.0 & 46.0 & 76.0 & 72.0 & 68.0 \\
& Probing   & \textbf{84.0} & 74.0 & 76.0 & 70.0 & 78.0 & 62.0 & 74.0 & 78.0 & \textbf{75.0} \\
\noalign{\vskip 1mm}
\cline{2-11}
\noalign{\vskip 1mm}
\multirow{2}{*}{BioMed-InternVL3-1B}
& Prompting & 48.0 & 30.0 & 32.0 & 24.0 & 30.0 & \textbf{8.0} & 32.0 & 28.0 & 29.0 \\
& Probing   & 62.0 & 50.0 & 60.0 & 50.0 & 50.0 & 44.0 & 50.0 & 56.0 & 53.0 \\
\midrule
\multirow{2}{*}{LLaVA-7B}
& Prompting & 12.0 & 16.0 & 8.0 & \textbf{4.0} & \textbf{2.0} & 10.0 & 12.0 & 12.0 & \textbf{10.0} \\
& Probing   & 52.0 & 38.0 & 64.0 & 50.0 & 50.0 & 50.0 & 56.0 & 46.0 & \textbf{51.0} \\
\noalign{\vskip 1mm}
\cline{2-11}
\noalign{\vskip 1mm}
\multirow{2}{*}{LLaVA-Med-7B}
& Prompting & 20.0 & 20.0 & 16.0 & 14.0 & 10.0 & 24.0 & 10.0 & 8.0 & 15.0 \\
& Probing   & 48.0 & 46.0 & 54.0 & 48.0 & 50.0 & 50.0 & 58.0 & 50.0 & 51.0 \\
\midrule
\multirow{2}{*}{Qwen2-VL-2B}
& Prompting & 64.0 & 62.0 & 66.0 & 64.0 & 66.0 & 64.0 & 58.0 & 76.0 & 65.0 \\
& Probing   & 82.0 & 74.0 & 78.0 & 68.0 & 70.0 & 74.0 & 62.0 & \textbf{82.0} & \textbf{74.0} \\
\noalign{\vskip 1mm}
\cline{2-11}
\noalign{\vskip 1mm}
\multirow{2}{*}{BioMed-Qwen2-VL-2B}
& Prompting & 70.0 & 52.0 & 54.0 & 58.0 & 58.0 & 42.0 & 48.0 & 64.0 & 56.0 \\
& Probing   & 70.0 & 52.0 & 78.0 & 60.0 & 50.0 & 54.0 & 54.0 & 60.0 & 60.0 \\
\midrule
\multirow{2}{*}{Qwen2.5-VL-3B}
& Prompting & 70.0 & 72.0 & 66.0 & 62.0 & 66.0 & 78.0 & 30.0 & 80.0 & 66.0 \\
& Probing   & 80.0 & \textbf{84.0} & \textbf{84.0} & 76.0 & 70.0 & \textbf{88.0} & 58.0 & 82.0 & \textbf{78.0} \\
\noalign{\vskip 1mm}
\cline{2-11}
\noalign{\vskip 1mm}
\multirow{2}{*}{MedVLThinker-RL-3B}
& Prompting & 58.0 & 70.0 & 66.0 & 64.0 & 66.0 & 66.0 & 36.0 & 82.0 & 64.0 \\
& Probing   & 74.0 & 68.0 & 80.0 & 64.0 & 70.0 & 70.0 & 56.0 & 80.0 & 70.0 \\
\midrule
\multirow{2}{*}{Qwen2.5-VL-32B}
& Prompting & \textbf{8.0} & 22.0 & 14.0 & \textbf{4.0} & \textbf{2.0} & \textbf{6.0} & \textbf{0.0} & 24.0 & \textbf{10.0} \\
& Probing   & \textbf{86.0} & 80.0 & \textbf{86.0} & 76.0 & 76.0 & 78.0 & 72.0 & 80.0 & \textbf{79.0} \\
\noalign{\vskip 1mm}
\cline{2-11}
\noalign{\vskip 1mm}
\multirow{2}{*}{MedVLThinker-RL-32B}
& Prompting & 60.0 & 62.0 & 62.0 & 74.0 & 62.0 & 70.0 & 36.0 & 78.0 & 63.0 \\
& Probing   & 68.0 & 68.0 & 82.0 & 64.0 & 70.0 & 70.0 & 72.0 & 76.0 & 71.0 \\
\midrule
\multirow{2}{*}{Qwen3-VL-2B}
& Prompting & 50.0 & 76.0 & 70.0 & 68.0 & 52.0 & 48.0 & 80.0 & 72.0 & 65.0 \\
& Probing   & 82.0 & 78.0 & \textbf{90.0} & \textbf{86.0} & 76.0 & 70.0 & \textbf{90.0} & \textbf{86.0} & \textbf{82.0} \\
\noalign{\vskip 1mm}
\cline{2-11}
\noalign{\vskip 1mm}
\multirow{2}{*}{MediX-R1-2B}
& Prompting & 64.0 & 78.0 & 80.0 & 78.0 & 76.0 & 62.0 & \textbf{88.0} & 72.0 & \textbf{75.0} \\
& Probing   & 70.0 & 72.0 & 84.0 & 76.0 & 58.0 & 72.0 & 84.0 & 76.0 & 74.0 \\
\midrule
\multirow{2}{*}{Qwen3-VL-8B}
& Prompting & 68.0 & 68.0 & 66.0 & 56.0 & 76.0 & 58.0 & 50.0 & 52.0 & 62.0 \\
& Probing   & \textbf{84.0} & \textbf{82.0} & \textbf{92.0} & \textbf{82.0} & \textbf{82.0} & \textbf{84.0} & \textbf{82.0} & \textbf{84.0} & \textbf{84.0} \\
\noalign{\vskip 1mm}
\cline{2-11}
\noalign{\vskip 1mm}
\multirow{2}{*}{MedMo-8B}
& Prompting & \textbf{98.0} & \textbf{98.0} & \textbf{100.0} & \textbf{98.0} & \textbf{100.0} & \textbf{100.0} & \textbf{100.0} & \textbf{100.0} & \textbf{99.0} \\
& Probing   & 92.0 & 98.0 & \textbf{100.0} & 98.0 & 96.0 & \textbf{100.0} & \textbf{100.0} & 94.0 & 97.0 \\
\midrule
\multirow{2}{*}{Qwen3-VL-30B}
& Prompting & \textbf{6.0} & \textbf{6.0} & 12.0 & \textbf{4.0} & \textbf{0.0} & \textbf{2.0} & \textbf{0.0} & 18.0 & \textbf{6.0} \\
& Probing   & \textbf{86.0} & \textbf{90.0} & \textbf{90.0} & \textbf{84.0} & \textbf{88.0} & 82.0 & \textbf{90.0} & \textbf{94.0} & \textbf{88.0} \\
\noalign{\vskip 1mm}
\cline{2-11}
\noalign{\vskip 1mm}
\multirow{2}{*}{MediX-R1-30B}
& Prompting & 72.0 & 64.0 & 82.0 & 74.0 & \textbf{84.0} & 72.0 & 76.0 & 80.0 & \textbf{76.0} \\
& Probing   & 62.0 & 62.0 & 72.0 & 58.0 & 64.0 & 64.0 & 70.0 & 70.0 & 65.0 \\
\bottomrule
\end{tabular}
} 
\caption{Prompting and probing accuracy (\%) across eight medical imaging modalities on OmniMedVQA. Bold values highlight the most striking contrasts between the two paradigms, including near-zero prompting scores for large general-purpose models and the rare cases where prompting matches or exceeds probing.}
\label{tab:modality}
\end{table*}

Table~\ref{tab:modality} reports accuracy for each model under prompting and probing across eight medical imaging modalities on the OmniMedVQA evaluation set.
Probing consistently and substantially outperforms prompting across virtually all modalities and model families, with the advantage being most dramatic for larger general-purpose models.
Qwen2.5-VL-32B is the starkest example: its prompting average collapses to \textbf{10.0\%}, with near-zero scores on MRI (\textbf{2.0\%}) and Ultrasound (\textbf{0.0\%}), whereas probing recovers a strong average of \textbf{79.0\%} on the same frozen model.
A nearly identical pattern holds for Qwen3-VL-30B, whose prompting average of \textbf{6.0\%} contrasts with a probing average of \textbf{88.0\%}, a gap of 82 percentage points, confirming that the clinical knowledge encoded in large-scale general-purpose models is severely suppressed by free-text generation.
Similarly, LLaVA-7B achieves a prompting average of only \textbf{10.0\%} yet reaches \textbf{51.0\%} under probing, and Open-Flamingo-9B improves from \textbf{28.0\%} to \textbf{66.0\%}.
The probing advantage is consistent across modalities: even in domains typically considered visually demanding, such as Dermoscopy and OCT, probing reliably extracts clinically relevant signal that prompting fails to surface.
The principal exception in the table is MedMo-8B, whose prompting average of \textbf{99.0\%} is the highest across all conditions, a result that likely reflects the specific prompt format and dataset overlap with its medical training corpus rather than a general superiority of generation-based decoding.
MediX-R1-30B also reverses the trend, with prompting (\textbf{76.0\%}) exceeding probing (\textbf{65.0\%}), consistent with the observation in Section~5.4 that reinforcement-learning-based adaptation can alter internal representations in ways that reduce linear separability at the layer level.
Taken together, these modality-level results reinforce the central conclusion of the paper: the bottleneck in medical VQA lies in how knowledge is elicited, not in whether the model encodes it, and this failure of prompted generation is pervasive across all eight imaging modalities examined.

\section{Baseline Details}
\label{sec:baseline_details}

This section describes the baseline systems evaluated in our study, covering both multi-agent frameworks (\S\ref{subsec:multiagent_baselines}) and specialized medical vision-language models (\S\ref{subsec:medvlm_baselines}). All our experiments are performed on a server equipped with 2 NVIDIA H200 GPUs (140GB memory each) and an Intel(R) Xeon(R) Gold 6442Y CPU.

\subsection{Multi-Agent Frameworks}
\label{subsec:multiagent_baselines}

We evaluate three multi-agent frameworks that coordinate multiple vision-language agents for medical visual question answering.

\paragraph{MAM~\cite{zhou2025mam}.}
MAM (Modular Multi-Agent Framework) decomposes the medical diagnostic process into specialized roles, General Practitioner, Specialist Team, Radiologist, Medical Assistant, and Director, each embodied by a separate LLM-based agent.
In our experiments, all agents are instantiated using \textbf{HuatuoGPT-Vision-7B}~\cite{chen2024towards}.

\paragraph{MMedAgent~\cite{li2024mmedagent}.}
MMedAgent (Multi-modal Medical Agent) is the first agent framework designed specifically for the medical domain, enabling tool selection across six medical tasks spanning five modalities.
The backbone vision-language model used by MMedAgent is \textbf{LLaVA-Med}~\cite{li2023llava}.

\paragraph{UCAgents~\cite{feng2025ucagents}.}
UCAgents employs a unidirectional convergence mechanism for visual evidence-anchored multi-agent medical decision-making.
In our experiments, we instantiate UCAgents with \textbf{Qwen2.5-VL-72B-Instruct}~\cite{bai2025qwen25vl}, which yielded the best performance among the model variants they evaluated.

\subsection{Medical Vision-Language Models}
\label{subsec:medvlm_baselines}

We evaluate four specialized Med-VLMs: InfiMed-RL-3B~\cite{liu2025infimed}, UniMedVL-14B~\cite{ning2025unimedvl}, HuatuoGPT-Vision-34B~\cite{chen2024towards}, and Aloe-Vision-72B-AR~\cite{guaschaloe}.
Following the official documentation and recommended configurations of each model, we describe the inference hyperparameters used for each below.

\paragraph{InfiMed-RL-3B~\cite{liu2025infimed}.}
InfiMed-RL-3B is built on top of Qwen2.5-VL-3B and runs in bfloat16 precision. We use greedy decoding, meaning the model always selects the most probable next token at each step, and cap generation at 256 tokens. Input image resolution follows the processor defaults recommended in the official documentation, with a minimum of $256 \times 28^{2}$ pixels and a maximum of $1280 \times 28^{2}$ pixels.

\paragraph{UniMedVL-14B~\cite{ning2025unimedvl}.}
UniMedVL-14B is run in bfloat16 precision with greedy decoding and a maximum generation length of 128 tokens. A fixed random seed is set for full reproducibility, following the official inference configuration.

\paragraph{HuatuoGPT-Vision-34B~\cite{chen2024towards}.}
HuatuoGPT-Vision-34B is loaded in float16 precision. Unlike the other Med-VLMs evaluated, it uses stochastic sampling rather than greedy decoding: the temperature is set to 0.2, which slightly sharpens the output distribution, and a repetition penalty of 1.2 is applied to discourage repetitive outputs. Generation is capped at 256 tokens. These settings correspond to the default generation configuration provided in the official model interface.

\paragraph{Aloe-Vision-72B-AR~\cite{guaschaloe}.}
Aloe-Vision-72B-AR is run in bfloat16 precision with greedy decoding and a maximum of 32 new tokens, following the generation configuration recommended in the official model documentation.

\subsection{Fine-Tuned Models}
\label{sec:fine_tuned_models}

We fine-tuned two vision-language models: Qwen2.5-VL-7B-Instruct and Llama-3.2-11B-Vision-Instruct. Each model was fine-tuned in four settings: three single-domain settings using PATH-VQA, SLAKE, and VQA-RAD separately, and one multi-domain setting using the combined training data from all three benchmarks. This setup lets us compare in-domain specialization against broader medical adaptation under a consistent fine-tuning pipeline.

All runs used full fine-tuning rather than probe training or adapter-only updates, and the supervision format matched the repository's standard medical VQA prompting setup. Inputs were formatted as image-plus-question prompts with multiple-choice options, and the model was trained to generate the gold answer. Optimization used fused AdamW with learning rate $2\times10^{-5}$, weight decay $0.01$, warmup ratio $0.03$, bf16 precision, seed 42, and an effective batch size of 8. For Llama-3.2-11B-Vision-Instruct, we used per-device batch size 1 with gradient accumulation 8, while for Qwen2.5-VL-7B-Instruct, we used per-device batch size 8 without gradient accumulation.

\section{Probe Ablation Details}
\label{sec:probe_ablations}

\subsection{Nonlinear Probe (MLP)}

A natural question is whether the linear constraint of logistic regression is the bottleneck.
We trained a small MLP probe on the same frozen embeddings for three representative models.
The MLP does not consistently improve over logistic regression: it raises average accuracy for only one of the two models (Qwen3-VL-8B: 81.4\%$\,\to\,$81.6\%), while Llama-3.2-11B-Vision is unchanged or lower (81.3\%$\,\to\,$80.9\%).
These results indicate that the clinical knowledge relevant to our benchmarks is largely linearly separable within a single transformer layer's activation space, validating logistic regression as the appropriate and parsimonious probe architecture.
Table~\ref{tab:mlp_probe} provides full per-dataset results.

\begin{table}[h]
\centering
\resizebox{\columnwidth}{!}{%
\begin{tabular}{lcccc}
\toprule
\textbf{Model} & \textbf{Log.\ Acc.} & \textbf{MLP Acc.} & \textbf{Log.\ F1} & \textbf{MLP F1} \\
\midrule
Llama-3.2-11B-Vision & 81.3 & 80.9 & 85.9 & 81.1 \\
Qwen3-VL-8B          & 81.4 & 81.6 & 82.3 & 71.3 \\
\bottomrule
\end{tabular}%
}
\caption{Average accuracy and macro-F1 (\%) across PATH-VQA, SLAKE, and VQA-RAD for the linear logistic probe and the nonlinear MLP probe.}
\label{tab:mlp_probe}
\end{table}

\subsection{PATH-VQA Sampling Sensitivity}

PATH-VQA results throughout the paper are based on a stratified 500-example sample from the full 3,389-example test set.
To confirm that our conclusions are not sensitive to a particular random draw, we repeated the evaluation with three independently resampled subsets (seeds 42, 43, 44).
Accuracy standard deviations are small across all models: LLaVA-V0-7B $87.1 \pm 0.9$, Llama-3.2-11B-Vision $89.9 \pm 0.5$, Qwen3-VL-8B $91.7 \pm 0.6$.
Best layers vary modestly across resamples (e.g., layers 16--27 for LLaVA-V0-7B), but the rankings among models remain stable.
Table~\ref{tab:pathvqa_seeds} reports best-layer probing accuracy on three independently resampled subsets, confirming that the stratified 500-example protocol is a stable proxy for the full test distribution.

\begin{table}[h]
\centering
\resizebox{\columnwidth}{!}{%
\begin{tabular}{lccc}
\toprule
\textbf{Model} & \textbf{Acc.} & \textbf{Macro-F1} & \textbf{Best Layers} \\
\midrule
LLaVA-V0-7B          & $87.1 \pm 0.9$ & $87.0 \pm 0.9$ & 18 / 16 / 27 \\
Llama-3.2-11B-Vision & $89.9 \pm 0.5$ & $89.8 \pm 0.5$ & 20 / 13 / 23 \\
Qwen3-VL-8B          & $91.7 \pm 0.6$ & $91.6 \pm 0.6$ & 21 / 25 / 23 \\
\bottomrule
\end{tabular}%
}
\caption{PATH-VQA sampling sensitivity: mean\,$\pm$\,std across seeds 42, 43, and 44. Best Layers lists the optimal layer per seed in the same order.}
\label{tab:pathvqa_seeds}
\end{table}

\section{Probing vs.\ Prompting Across All Models and Benchmarks}
\label{sec:all_models_probing_prompting}

\begin{figure*}[t]
    \centering
    \includegraphics[width=\linewidth]{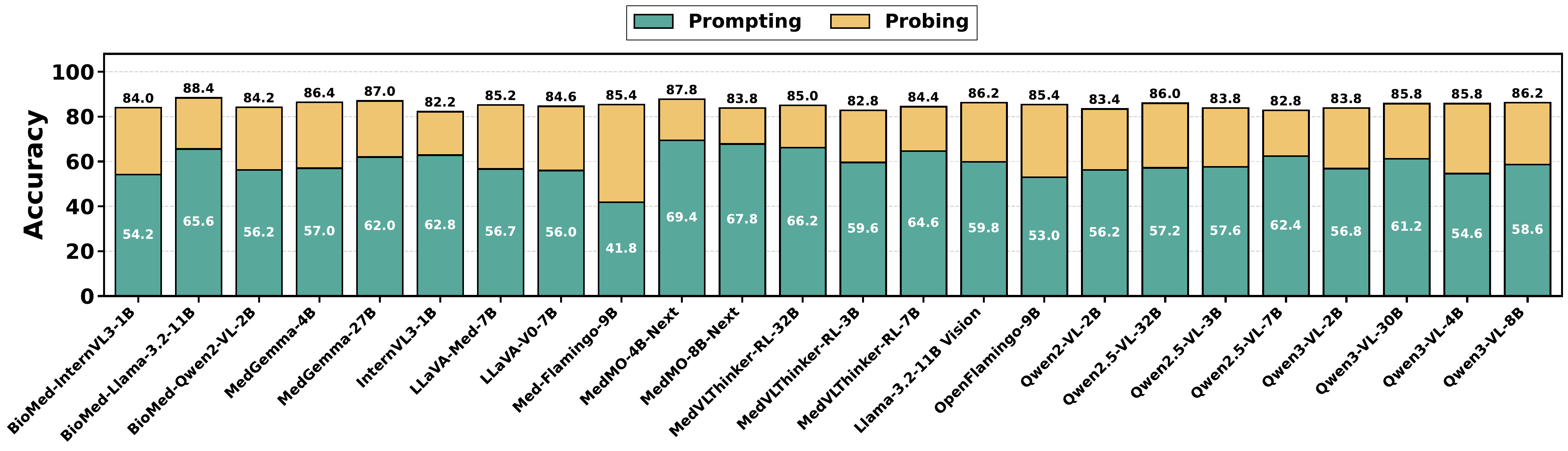}
    \caption{Probing vs.\ prompting accuracy (\%) on PATH-VQA for 24 of the 28 evaluated models, sorted by probing score.}
    \label{fig:all_path_vqa}
\end{figure*}

\begin{figure*}[h]
    \centering
    \includegraphics[width=\linewidth]{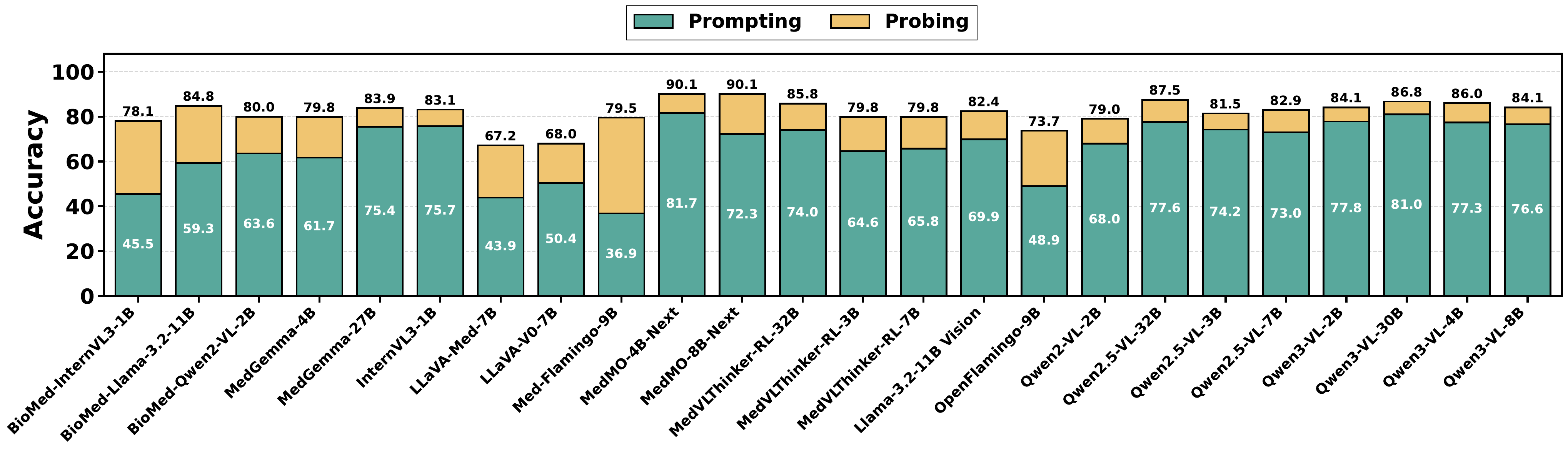}
    \caption{Probing vs.\ prompting accuracy (\%) on SLAKE for 24 of the 28 evaluated models, sorted by probing score.}
    \label{fig:all_slake}
\end{figure*}

\begin{figure*}[h]
    \centering
    \includegraphics[width=\linewidth]{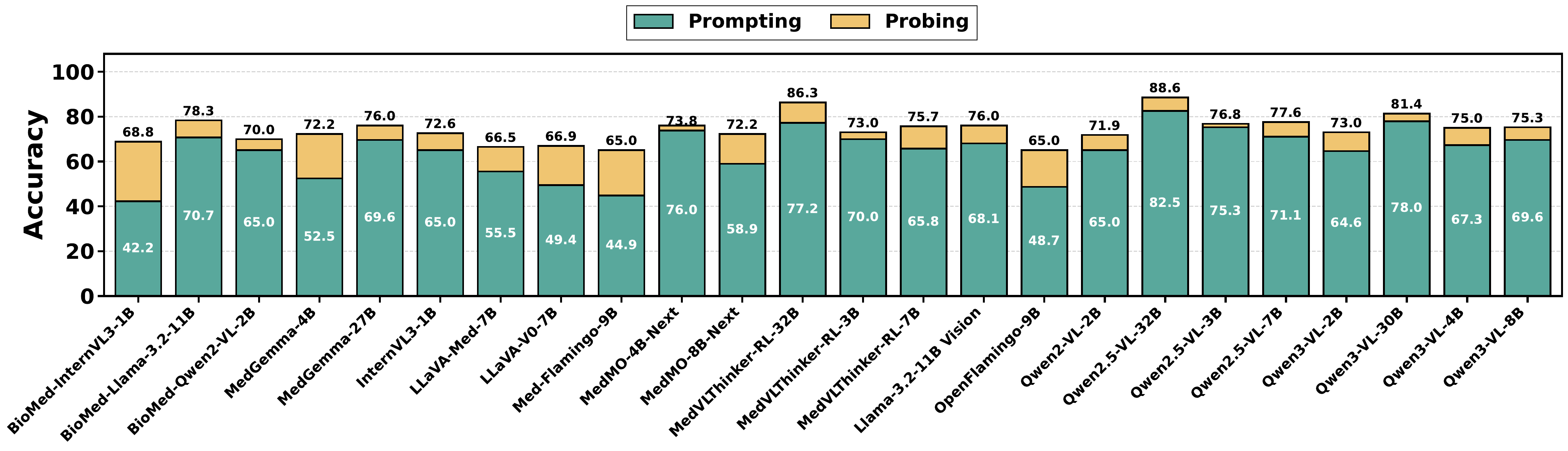}
    \caption{Probing vs.\ prompting accuracy (\%) on VQA-RAD for 24 of the 28 evaluated models, sorted by probing score.}
    \label{fig:all_vqa_rad}
\end{figure*}

Figures~\ref{fig:all_path_vqa}--\ref{fig:all_vqa_rad} report probing and prompting accuracy for 24 models on PATH-VQA, SLAKE, and VQA-RAD. Across all three benchmarks, probing consistently and substantially outperforms prompting, confirming the main finding of the paper at scale: models encode clinically relevant knowledge that their generation-based interface fails to elicit.

On PATH-VQA (Figure~\ref{fig:all_path_vqa}), the probing advantage is universal and often dramatic. Med-Flamingo-9B exhibits the largest gap, with probing reaching 85.4\% against a prompting accuracy of only 41.8\%, a margin of 43.6 points. OpenFlamingo-9B follows a similar pattern (85.4\% vs.\ 53.0\%), confirming that Flamingo-family models are particularly susceptible to elicitation failure despite encoding strong medical signal. BioMed-Llama-3.2-11B achieves the highest probing accuracy at 88.4\%, while MedMO-4B-Next leads under prompting at 69.4\%, the only model that partially closes the gap between the two paradigms on this benchmark.

On SLAKE (Figure~\ref{fig:all_slake}), the probing advantage persists across all models, but the prompting scores are generally higher than on PATH-VQA, suggesting that SLAKE questions are somewhat more amenable to generation-based decoding. Med-Flamingo-9B again shows the largest collapse under prompting (36.87\%), compared to a probing accuracy of 79.52\%. MedMO-4B-Next and MedMO-8B-Next both reach the highest probing scores at 90.12\%, and MedMO-4B-Next also achieves the highest prompting score at 81.69\%, nearly halving the typical gap. Among larger models, MedVLThinker-RL-32B (85.78\% probing, 73.98\% prompting) and Qwen2.5-VL-32B (87.47\% probing, 77.59\% prompting) show that reinforcement-learning-based training and scale can simultaneously improve both the richness of internal representations and their accessibility through prompting.

VQA-RAD (Figure~\ref{fig:all_vqa_rad}) is the benchmark where the prompting gap narrows most for top-performing models. Qwen2.5-VL-32B achieves the highest probing accuracy at 88.59\%, followed closely by MedVLThinker-RL-32B at 86.31\%; the corresponding prompting scores are 82.51\% and 77.19\%, respectively, representing more modest gaps than are observed on the other two benchmarks. Notably, MedMO-4B-Next is the single model where prompting (76.05\%) marginally exceeds probing (73.76\%) on VQA-RAD, a rare reversal that mirrors the MedMo-8B exception observed in the modality analysis (\S\ref{tab:modality}) and again points to dataset-specific prompt format alignment rather than a general superiority of generation-based decoding. Flamingo-family models (Med-Flamingo-9B, OpenFlamingo-9B) and older architectures (LLaVA-V0-7B, LLaVA-Med-7B) continue to show the largest probing-to-prompting gaps, confirming that elicitation failure is most severe in models whose training did not explicitly optimize for structured VQA generation.

Taken together, these per-benchmark figures reinforce the central conclusion at full model scale: probing reliably recovers latent clinical knowledge that prompted generation cannot surface, and this advantage holds across diverse model families, parameter counts, and medical VQA benchmarks.

\section{MedGemma-27B Results (Omitted from Table~\ref{tab:main_results} for Space)}
\label{app:medgemma27b}

MedGemma-27B is omitted from Table~\ref{tab:main_results} due to space constraints in the main-paper table layout; it is included in Figure~\ref{fig:stacked_accuracy} because that figure focuses on the general vs.\ medical comparison within the Gemma family, where MedGemma-27B is the matched medical counterpart to Gemma-27B. Table~\ref{tab:medgemma27b} reports its full results for completeness; the pattern follows the other models in Table~\ref{tab:main_results}.

\begin{table}[t]
\centering
\small
\setlength{\tabcolsep}{4pt}
\begin{tabular}{lcccc}
\toprule
\textbf{Method} & \textbf{PATH-VQA} & \textbf{SLAKE} & \textbf{VQA-RAD} & \textbf{Avg.} \\
\midrule
Prompting & 36.40 & 63.13 & 54.37 & 51.30 \\
MedProb & 85.00 & 83.37 & 81.37 & 83.25 \\
\bottomrule
\end{tabular}
\caption{MedGemma-27B accuracy (\%), omitted from Table~\ref{tab:main_results} for space.}
\label{tab:medgemma27b}
\end{table}

\section{Use of AI Assistants}
AI-based tools were used to assist with language refinement, coding, and clarity. All ideas, analyses, and conclusions presented in this work are solely those of the authors.

\end{document}